# The Transformer Revolution, Part 1: Dynamic Processing through Output-Weight Interconnections

Marco Giunti[1] and Fabrizia Giulia Garavaglia[2]*

[1] ALOPHIS – Applied Logic, Philosophy and History of Science, University of Cagliari, Cagliari, Italy

[2] Department of Pedagogy, Psychology and Philosophy, University of Cagliari, Cagliari, Italy

** Corresponding author*

**Abstract:** We reinterpret Transformer inference by developing a functionally equivalent mechanical-structural description of its functional architecture. Parameterized transformations of token representations are identified with simple neural networks organized through output-input and output-weight interconnections. This reconstruction makes explicit a constitutive feature of ordinary Transformer inference: outputs generated from the current input determine the weights, and hence the transformations, of other networks. We show that the resulting mechanical-structural organization is functionally equivalent to the standard matrix description and gives rise to Sequence-level Interactive Dynamic Parallel Processing (SIDPP). We then quantify the dynamic component of inference. For a fixed Transformer architecture, the number of dynamic parameters generated by a prompt grows linearly with prompt length, with a proportionality constant determined by the model architecture, whereas the number of static parameters remains constant. We call the resulting structural dependence of processing on the prompt strong prompt sensitivity. GPT-4.5's recent Turing test success provides a striking behavioral phenomenon for which this framework offers a viable explanatory route. Finally, we discuss implications for interpretability, control, predictability, and neuromorphic construction. We argue that known biological mechanisms support the in-principle neural constructibility of a form of SIDPP, and on this basis formulate Conjecture T as a conditional research hypothesis concerning cortical language processing.

## 1 Introduction

Since the launch of ChatGPT in November 2022, the Transformer architecture has become the basis of AI systems with striking new capabilities. GPT-4.5, for example, recently passed a standard three-party Turing test with statistically robust results (Jones & Bergen, 2025, 2026a; cf. Giunti, 2026). Independently of its status as a test of intelligence, results of this kind raise a basic explanatory question: what is distinctive about the Transformer's organization and processing that makes such capacities possible?

A widespread view treats inference largely as the application of statistical regularities encoded in trained parameters, an idea captured by the "stochastic parrot" metaphor (Bender et al., 2021). Yet inference also generates matrices and vectors whose values depend on the current input and are then used to transform token representations. Understanding the Transformer therefore requires explaining not only what is fixed by training, but also how prompt-dependent transformations are produced and used.

This is also a problem of scientific explanation. Mechanistic accounts explain systems by identifying components and operations and showing how their organization produces the phenomenon of interest (Bechtel & Abrahamsen, 2005). Likewise, mathematical models acquire explanatory force when their elements can be plausibly related to elements of the relevant mechanism (Kaplan & Craver, 2011). The standard matrix description of the Transformer specifies its computation precisely, but does not make all organizational dependencies equally salient. Our analysis therefore proceeds in two steps: we first reformulate this functional description so as to make the relevant transformations and dependencies explicit; we then develop a functionally equivalent mechanical-structural description that brings into view an organizationally significant aspect of the architecture: during inference, the outputs of some networks determine the weights, and hence the transformations, of others.

Architectures in which one network generates or modulates parameters used by another predate the Transformer, including hypernetworks and fast-weight systems (Ha et al., 2017; Schlag et al., 2021). Our claim is therefore not that input-dependent weights are unique to the Transformer. Rather, we show through an explicit reconstruction of standard Transformer inference that dependencies of this kind are built into its ordinary operation: outputs generated from the current input supply weights for subsequent transformations. This reconstruction makes those dependencies explicit as output-weight interconnections and provides the basis for the analysis of dynamic processing developed below.

At the functional level, we characterize the Transformer as a system that transforms concepts by means of concepts (§§ 2.1–2.2). Token representations are the concepts to be transformed, whereas the transforming concepts are parameterized transformations of token vectors, each of which can be identified with a simple neural network (§ 2.3). Because matrix operations organize the corresponding transformations in parallel (§ 2.4), this identification can be extended to the Transformer's functional architecture as a whole, yielding groups of simple neural networks linked by different types of interconnections (§ 2.5). Among these, output-weight interconnections allow input-dependent outputs of some networks to determine the weights of others (§ 2.6). The resulting mechanical-structural reconstruction is functionally equivalent to the standard description with respect to the Transformer's input-output function (Proposition 1). We then show that this organization gives rise to SIDPP—Sequence-level Interactive Dynamic Parallel Processing (§ 3).

We also quantify the dynamic component of inference. For a fixed Transformer architecture, the number of dynamic parameters generated by a prompt is a linear function of its length, whereas the number of static parameters is constant (Proposition 2). This determines the number of dynamic parameters generated by a prompt of any given length and yields a direct expression for their numerical ratio to the static parameters (§ 4). The resulting structural dependence of processing on the information supplied by the prompt, which we call *strong prompt sensitivity*, provides a natural explanatory framework for a striking behavioral phenomenon in GPT-4.5's recent Turing test results (§ 5).

These results also have implications for AI research. By making prompt-dependent structural dependencies explicit, the analysis may contribute to interpretability, control, and predictability (§ 6.1). The mechanical-structural description also suggests the possibility of constructing physical systems based on a neuromorphic realization of the Transformer's functional architecture, in which the relevant dynamics are produced through simpler local mechanisms (§ 6.2).

We then ask whether the processing principle identified here is neurally constructible. We argue that known biological mechanisms provide the functional components required for output-weight interconnections and support the in-principle neural constructibility of a form of SIDPP (§ 7.1). On this basis, we formulate Conjecture T as a conditional research hypothesis: if SIDPP is the type of processing actually produced by the cortical systems involved in human language processing, we conjecture that it is produced by a neural construction of a functional architecture relevantly similar to that of the Transformer (§ 7.2).

We do not assess Conjecture T in the present paper; rather, we argue that it is a scientifically plausible hypothesis that warrants further investigation. Assessing whether it is supported or challenged by the known morphology and functioning of cortical systems involved in language processing lies beyond the scope of the present paper and will be the subject of a separate study, Part 2, specifically devoted to this question. Finally, § 8 summarizes the results of the present paper and their implications.

## 2 The Transformer as a System Composed of Groups of Simple Neural Networks Linked by Output–Input and Output–Weight Interconnections

We focus here on the Transformer's operation during inference. Its standard mathematical description (Vaswani et al., 2017; Jurafsky & Martin, 2026) characterizes processing in terms of operations on matrices of token representations. Our aim is not to propose an alternative mathematical formulation, but to redescribe the Transformer's functional architecture at a level of organization that makes its prompt-dependent transformations and the relations among them explicit. We begin by isolating the elementary transformations of vector representations of tokens involved in inference and interpreting them as transforming concepts acting on transformee concepts (§§ 2.1–2.2). We then identify these vector transformations with simple neural networks (§ 2.3). This provides the basis for reformulating the matrix description in terms of parallel applications of such transformations (§ 2.4; Supplementary Material (OR1), § S2) and, subsequently, for developing the corresponding mechanical-structural description (§§ 2.5–2.6).

### *2.1 The Transformer Is a System That Transforms Concepts by Means of Concepts*

From this perspective, the Transformer can be characterized as a system that transforms concepts by means of concepts (Figure 1). *Transformee concepts* are the vector representations of tokens, both the initial embedding vectors and those generated during processing. *Transforming concepts* are parameterized transformations of these representations, defined by matrices or vectors of model parameters. They are *static* when their parameters are fixed through training, and *dynamic* when their defining matrices or vectors are generated during inference (OR1, § S1).

Why can token representations be regarded as concepts? Each position in a token vector corresponds to a representational feature, or property, and the number occupying that position gives the value of that feature for the token. The vector therefore represents the token through an ordered set of feature values, in the sense of a distributed representation (Bengio et al., 2003; Hinton, 1986; Bengio et al., 2013). This does not require the individual features to be directly interpretable in human terms (Şenel et al., 2021) or to remain invariant across processing stages (Jawahar et al., 2019).

Let us now consider a transforming concept, $\mathcal{L}[\boldsymbol{M}]$, defined by a matrix $\boldsymbol{M} \in \mathbb{R}^{q\times p}$ of model parameters, and apply it to the vector representation $\boldsymbol{u} \in \mathbb{R}^{q}$ of a token. Each row $r_i(\boldsymbol{M})$ can be regarded as a vector representation of the $i$-th feature, or property, of $\boldsymbol{u}$, that is, as a second-order representation. Each $r_i(\boldsymbol{M})$ is weighted by the corresponding feature value $u_i$, and the resulting weighted representations are summed to produce the new token representation:

$$\mathcal{L}[\boldsymbol{M}](\boldsymbol{u}) \coloneqq \sum_{i=1}^{q} u_i \, r_i(\boldsymbol{M}).$$

The new token representation is therefore a combination of the vector representations of the features of the previous representation, weighted according to the values of those features, so that the representations of higher-valued features contribute more strongly.

A similar mechanism applies to the transforming concepts $+[\boldsymbol{v}]$ and $\circ[\boldsymbol{v}]$, defined by a vector $\boldsymbol{v} \in \mathbb{R}^q$ of model parameters. Applied to a token representation $\boldsymbol{u} \in \mathbb{R}^q$, each element $v_i \in \mathbb{R}$ can be regarded as a *numerical representation*[1] of the $i$-th feature, or property, of $\boldsymbol{u}$, and thus again as a second-order representation. Each second-order representation is then modified by adding the corresponding feature value $u_i$ to it, or by multiplying it by that value. Finally, all the representations thus modified are concatenated in their original order, thereby producing the new vector representations of the token:

$$+[\boldsymbol{v}](\boldsymbol{u}) \coloneqq \boldsymbol{u} + \boldsymbol{v} = \left(u_1 + v_1, \dots, u_q + v_q\right)$$

and

$$\circ[\boldsymbol{v}](\boldsymbol{u}) \coloneqq \boldsymbol{u} \circ \boldsymbol{v} = \left(u_1 v_1, \dots, u_q v_q\right).$$

Thus, as in the case of $\mathcal{L}[\boldsymbol{M}]$, the new token representations are aggregations of representations of the features of the previous representation, in which the representations of the features with the highest values predominate.

### *2.2 Dynamic Transforming Concepts as Adaptations of Static Transforming Concepts Based on an Input Sequence*

An important consequence of this interpretation is that inference cannot be understood as the mere application of what has been learned during training. *Static* transforming concepts embody previously acquired knowledge: their parameters are fixed through training, and they transform token representations accordingly. During inference, however, the Transformer also generates new matrices and vectors on the basis of the current input and uses them to define further transformations. These are the *dynamic* transforming concepts. More specifically, they are the transformations defined, in each attention head module, by the transpose of the key matrix and by the value matrix, and, in each residual connection, by the output vectors of the corresponding attention or feed-forward module (see OR1, § S1).

More precisely, given an input sequence $X$, the Transformer first uses static transforming concepts to generate new matrices or vectors whose values depend on $X$; these then define dynamic transforming concepts that are applied to the token representations. Dynamic transforming concepts can therefore be understood as adaptations of static transforming concepts on the basis of an input sequence. Inference thus involves two successive stages: first, previously acquired knowledge is adapted to the current input; second, the resulting adapted knowledge is applied to transform the representations of that input.

This explanatory picture is also compatible with in-context learning (Brown et al., 2020; Akyürek et al., 2023). If Transformer inference involves not only applying previously acquired transformations but also constructing input-dependent ones, then in-context adaptation may depend, at least in part, on the construction of transformations better suited to the specific problem posed by the current context.

[1] In general, a numerical representation of an object of a certain type identifies a single feature (or property) of objects of that type, whose value for that object is the number itself.

To understand how this dynamic adaptation is mechanically realized, however, we must move from this functional description to the corresponding mechanical-structural description. We therefore begin by showing that each of the three types of transforming concept can be identified with a simple neural network (§ 2.3).

### *2.3 Transforming Concepts as Three Types of Simple Neural Networks*

The transition to a mechanical-structural description begins by noting that transforming concepts—the parameterized transformations of the three types $\mathcal{L}[\boldsymbol{M}]$, $+[\boldsymbol{v}]$, and $\circ[\boldsymbol{v}]$—can be identified with three corresponding types of simple neural networks.

1. Transforming concepts of the type $\mathcal{L}[\boldsymbol{M}]$, where $\boldsymbol{M} \in \mathbb{R}^{q \times p}$, are *dense linear networks* with $q$ input units and $p$ output units. Each input unit is connected to every output unit. The $i$-th row of $\boldsymbol{M}$ gives the weights of the connections outgoing from the $i$-th input unit, while the $j$-th column gives the weights of the connections incoming to the $j$-th output unit. Input units merely route their incoming signals through their outgoing connections; each output unit produces the weighted sum of the signals it receives, thereby implementing the corresponding component of $\mathcal{L}[\boldsymbol{M}]$.
2. Transforming concepts of the type $+[\boldsymbol{v}]$, where $\boldsymbol{v} \in \mathbb{R}^{q}$, are *one-to-one additive networks* with $q$ input and $q$ output units. The $i$-th input unit is connected only to the $i$-th output unit, and the weight of this connection is $v_i$. The input unit merely routes its incoming signal, while the output unit adds that signal to $v_i$, thereby implementing the $i$-th component of $+[\boldsymbol{v}]$.
3. Transforming concepts of the type $\circ[\boldsymbol{v}]$, where $\boldsymbol{v} \in \mathbb{R}^{q}$, are *one-to-one multiplicative networks*. Their morphology is identical to that of one-to-one additive networks; they differ only in the operation performed by the output units, each of which *multiplies* its incoming signal by the corresponding weight $v_i$, thereby implementing the $i$-th component of $\circ[\boldsymbol{v}]$.

This identification provides the first step in a mechanistic decomposition of the Transformer. At the level of decomposition adopted here, the relevant basic components are the three types of simple neural networks just described, rather than the individual units and connections that constitute them. Each transforming concept is thus mapped onto a component capable of realizing the corresponding transformation. By itself, however, this local mapping does not yet provide a mechanical-structural description of the system, since such a description also requires specifying how these components are organized. We therefore next show how matrix operations organize the corresponding transformations in parallel (§ 2.4) and how the resulting groups of simple networks are interconnected (§§ 2.5–2.6).

### *2.4 Sequence-Level Parallel Processing in the Transformer*

Having identified each transforming concept with a simple neural network, we now turn to how the corresponding transformations are organized within the Transformer. The analysis concerns a single-stack Transformer architecture (see Figure 1) whose layers have the functional organization and mathematical operations of the encoder layers introduced by Vaswani et al. (2017). With bidirectional attention heads, the stack preserves the encoder's functional organization. With autoregressive attention heads, it constitutes an autoregressive decoder-only Transformer. Autoregressivity can be obtained either through masking or through the alternative functional organization without masking specified in OR1, §S4. The mechanical-structural reconstruction developed below also extends to the GPT-2/GPT-3 variants, subject to the differences in their functional organization discussed in §2.6.

At the functional level, inference can be divided into a sequence of successive modules. Each module $\mathcal{M}$ receives a given input $\boldsymbol{I}$ and produces a specific output $\boldsymbol{O}$, which is then passed as input to a subsequent module or constitutes the final output. From the module that performs positional encoding to the final one that produces probability distributions, the input and output of each module are matrices whose rows are vector representations of the prompt tokens. The number $n$ of rows therefore cannot exceed $n_{max}$, the size of the Transformer's context window. The dimensions $d_{in}$ and $d_{out}$ of the input and output token representations may differ from one module to another.

*Functional diagram of the inference processing cycle of a single-stack Transformer.*

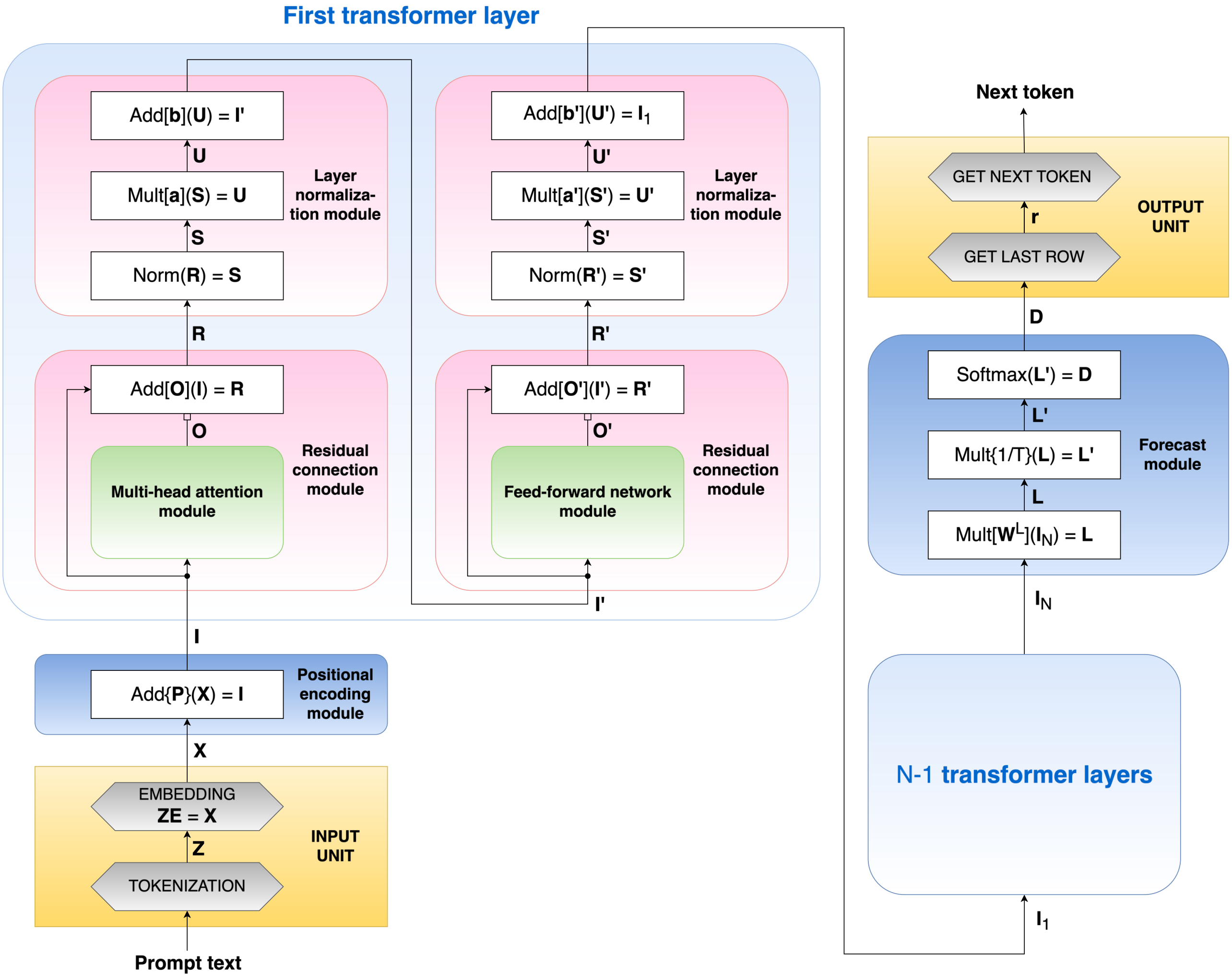

Within each module $\mathcal{M}$, processing may follow a single process from input $\boldsymbol{I}$ to output $\boldsymbol{O}$, as in a feed-forward network module (OR1, Fig. S2) and in the Positional encoding, Layer normalization, and Forecast modules in Figure 1. Alternatively, multiple processes may operate in parallel on the same input $\boldsymbol{I}$. In a multi-head attention module, these processes *merge* into a single process with output $\boldsymbol{O}$ (OR1, Fig. S3); in a residual connection module, two processes are reduced to one, whereas in an attention head module successive *reductions* of two processes to one occur (Figure 1; OR1, Figs. S4–S6).

Each process consists of a finite sequence of matrix operations, each taking the output of the preceding one as input. These operations are *parameterized, structural, or formal*. Parameterized operations are built from the transforming concepts introduced above: they transform token representations by means of matrices or vectors of model parameters, whether their values are fixed through training or dynamically generated during inference. They include vector addition or componentwise multiplication ($Add[\boldsymbol{v}]$, $Mult[\boldsymbol{v}]$) and matrix addition or multiplication ($Add[\boldsymbol{M}]$, $Mult[\boldsymbol{M}]$).[2] Structural operations are fixed by the architecture and include rescaling ($Mult\{c\}$), masking or fixed positional addition ($Add\{\mathfrak{M}\}$), nonlinear activation ($ReLU$), softmax ($Softmax$), and normalization ($Norm$). Parameterized and structural operations act in parallel on the $n$ rows of their input matrices, each row corresponding to one of the $n$ tokens being processed. Formal operations only rearrange their inputs: matrix concatenation ($Concat$) is used in merging multiple processes into one, while matrix transposition ($Transp$) is used in one form of reducing two processes to one. Formal definitions are given in OR1, § S2.

Multiple parallel processes merge into one through a shared $Concat$ operation, which concatenates their outputs along the column axis. In a multi-head attention module, the output matrices of the $h$ attention head modules are thus concatenated into a single matrix, which is then processed by the output projection (OR1, Fig. S3). Any operations following concatenation count as continuations of all $h$ processes, rather than as a further process.

Two parallel processes are reduced to one when the final output of one process, rather than being further processed as an input, supplies the parameters of an operation in the other process. In the functional organization illustrated here, these reductions occur at four classes of architectural loci and correspond to four distinct modes, determined by the terminating and receiving operations. One of these modes involves the formal operation of matrix transposition, *Transp* (OR1, §S3.1). The corresponding classification for GPT-2/GPT-3 is discussed in §2.6.

### *2.5 Matrix Operations as Groups of Simple Neural Networks, or as Output-Input Interconnections—Operational and Routing—Between such Groups*

The parallel organization described in § 2.4 allows us to extend the local identification established in § 2.3 from individual transforming concepts to matrix operations. Every *parameterized operation* applies in parallel a transforming concept of the same type to the *rows* of its input matrix $\boldsymbol{I}$, that is, to the sequence of vector representations of the prompt tokens (see OR1, § S2). *Structural operations* have the same row-wise parallel form, although the function applied is fixed by the architecture rather than defined by model parameters. In particular, $Norm$ normalizes each token row across its features (Ba et al., 2016).

On this basis, every parameterized operation can be identified with a group of simple neural networks. Operations $Add[\boldsymbol{v}]$ and $Mult[\boldsymbol{v}]$ add or multiply each token representation by the same parameter vector, respectively, and thus correspond to parallel copies of a one-to-one additive or multiplicative network. The operation $Add[\boldsymbol{M}]$ corresponds instead to one-to-one additive networks whose weights are the corresponding rows of the parameter matrix, while matrix multiplication, $Mult[\boldsymbol{M}]$, corresponds to parallel copies of a dense linear network. For a prompt of $n$ tokens, any parameterized operation activates $n$ networks. Since $n \leq n_{max}$, however, it can be identified with a fixed group of $n_{max}$ networks of the relevant type, of which only the first $n$ are active (OR1, Figs. S8–S15c).

---

[2] We use square brackets for parameterized operations and curly brackets for structural operations. This notation marks a conceptual distinction. The parameters of a parameterized operation depend on the system's interactions with its environment: their values may be fixed through training or determined dynamically on the basis of the input during inference. By contrast, the constants occurring in structural operations are fixed by the theoretical and architectural specification of the model and do not depend on such interactions. Thus, for example, $Mult[\boldsymbol{v}]$ is parameterized, whereas $Mult\{c\}$ is structural.

An analogous identification holds for the structural operations $Mult\{c\}$ and $Add\{\mathfrak{M}\}$. Rescaling by a constant corresponds to one-to-one multiplicative networks whose weights are all equal to $c$, while adding a fixed matrix corresponds to one-to-one additive networks whose weights are the corresponding rows of $\mathfrak{M}$. Thus, for a prompt of $n$ tokens, each operation activates $n$ networks and can be identified with a fixed group of $n_{max}$ networks, of which only the first $n$ are active (OR1, Figs. S10–S12e, S15–S15c).

The remaining structural operations—$ReLU$, $Softmax$, and $Norm$, formally defined in OR1, § S2—differ from the preceding cases: they are not identified with groups of *simple neural networks*. Instead, each is realized by a group of operation-specific operational blocks inserted between the preceding and following groups of simple neural networks. Each block has two levels: input units that merely route incoming signals, and *functional units* that *modify* those signals and produce the block's output. Crucially, the connections between the two levels are unweighted. $ReLU$, $Softmax$, and $Norm$ operate token-wise: each operational block receives the output signals of one active preceding simple network and passes the transformed signals to the corresponding following network. They differ, however, in how they transform the components of each token representation. $ReLU$ acts componentwise, transforming each value independently (OR1, Figs. S14–S14c). $Softmax$ and $Norm$ instead normalize the components jointly, but in different ways: $Softmax$ produces a probability distribution (OR1, Figs. S10–S12e and S15–S15c), whereas $Norm$ uses their mean and variance (OR1, Figs. S13–S13c). Thus, each of the three matrix operations—$ReLU$, $Softmax$, and $Norm$—can be identified with a specific type of *operational output-input interconnection* between groups of simple neural networks, realized by the corresponding group of operation-specific operational blocks.

However, there are also interconnections that merely *transfer* the outputs of the preceding simple networks to the following ones, without modifying the signals, but regrouping them in a different way. These are *routing output-input interconnections*, which correspond to the formal operation $Concat$ (OR1, § S2; Figs. S9–S9c). More specifically, $h$ preceding groups, each consisting of $n_{max}$ networks of which only $n$ are active, feed a single following group of $n_{max}$ networks, again with only $n$ active. The output of each active network in a preceding group is transferred as one part of the input of the corresponding active network in the following group.

### *2.6 Output-Weight Interconnections Between Networks with Trainable Parameters and Networks with Dynamic Parameters*

The analysis of § 2.5 leaves only the formal transposition operation, $Transp$, to be accounted for in the mechanical-structural description. Its role becomes clear when we consider the reductions of two parallel processes to one introduced in § 2.4.

In each such reduction, one of the two processes *terminates*: its final operation produces an output that is not further processed as an input, but instead supplies the parameters of an operation in the other process, which we call the *receiving operation*. The parameters of the receiving operation are therefore not fixed through training, but are *dynamic parameters* determined by the output of the final operation of the terminating process.

In the functional organization illustrated here, reductions occur at four kinds of loci marked by small squares in OR1, Figs. S2–S6: (i) key transposition supplies the parameter matrix for multiplication of the queries; (ii) value projection supplies that for multiplication of the normalized attentions (both (i) and (ii) within attention heads, Figs. S4–S6); (iii) multi-head output projection supplies the parameters of the first residual addition (Fig. S3); and (iv) the final feed-forward bias addition supplies those of the second residual addition (Fig. S2).

Therefore, the receiving operations are always parameterized operations of the types $Mult[\boldsymbol{M}]$ or $Add[\boldsymbol{M}]$, whereas the final operations are parameterized operations of the types $Mult[\boldsymbol{M}]$ or $Add[\boldsymbol{v}]$, or the formal operation $Transp$ applied to the output of a $Mult[\boldsymbol{M}]$ operation. At the mechanical-structural level, reduction of two processes to one thus gives rise to a new type of interconnection between groups of simple networks: each *output unit* in the final group of the terminating process is linked to those *connections* in the receiving group *whose weights are identical to that unit's output*. We call these links *output-weight interconnections* (Figure 2; OR1, Figs. S8–S8c and S10–S12e).

*Figure 2. A dense linear network with dynamic weights determined by output–weight interconnections originating from output units of other networks.*

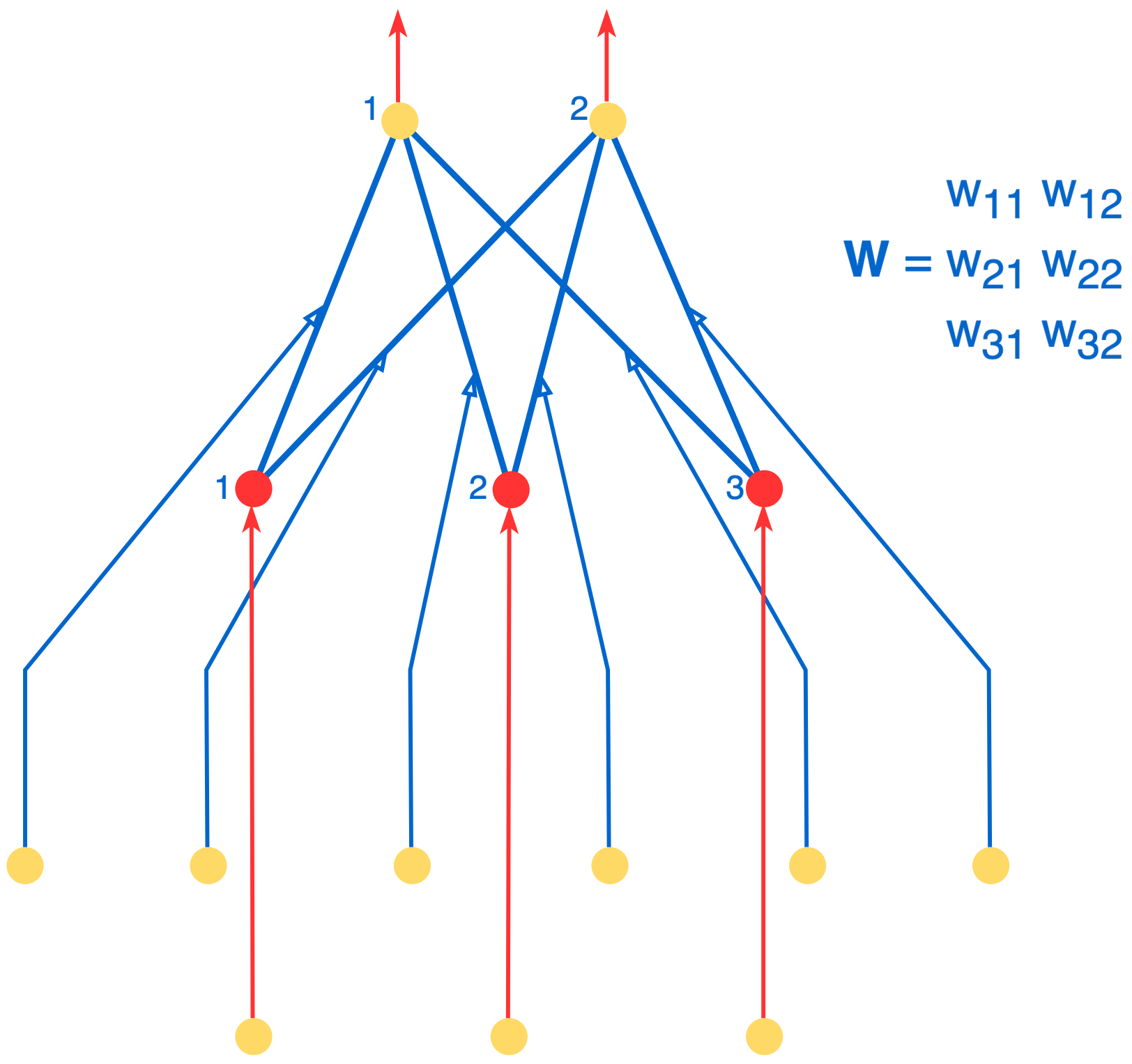

In GPT-2/GPT-3, the four kinds of reduction loci remain unchanged, but the introduction of attention-projection biases reduces the distinct reduction modes, and hence the types of output-weight interconnections, to three: (i) $Add[\boldsymbol{v}]$ followed by receiving $Mult[M]$ for values; (ii) Transp applied after $Add[\boldsymbol{v}]$, followed by receiving $Mult[M]$ for keys; and (iii) $Add[\boldsymbol{v}]$ followed by receiving $Add[M]$ for both residual additions (OR1, §S3). These variants also place layer normalization before each multi-head attention and feed-forward module, rather than after the corresponding residual addition, as in Figure 1. They additionally include a final normalization module and use $GELU$ instead of $ReLU$ (Radford et al., 2019; Brown et al., 2020). The different placement of layer normalization changes the functional organization of the residual branches but not the reduction modes.

The generation or modulation of weights by other parts of a system is not unique to the Transformer and has precedents in hypernetworks and fast-weight architectures (Ha et al., 2017; Schlag et al., 2021). The point established here is more specific: output-weight interconnections can be systematically identified as a constitutive part of the ordinary inference organization of the standard Transformer.

This completes the mechanical-structural reconstruction begun in § 2.3:

The argument developed in §§2.3–2.6 is constructive. It provides a mechanical realization of the functional organization of a single-stack Transformer, with autoregressive and/or bidirectional attention heads. The functional organization considered here is based on the general architectural structure and mathematical operations introduced by Vaswani et al. (2017), adapted to a single-stack configuration and to the variants specified in OR1. The complete functional organization is represented in Figures S1–S6. Its mechanical realization is obtained by associating each functional module with a system of simple neural networks linked by output-input interconnections, including operational interconnections, and/or output-weight interconnections (Figures S7–S15). Each functional module and its corresponding system of neural networks perform the same mathematical operations and functions in the same functional roles and are therefore input-output equivalent. The equivalence of the complete systems follows from the composition of these correspondences.

**Proposition 1 — Mechanical-structural equivalence.** *The standard matrix description of Transformer inference admits a functionally equivalent mechanical-structural representation in terms of groups of simple neural networks connected by output-input and output-weight interconnections.*

The functional and mechanical-structural descriptions make different aspects of the Transformer's organization salient. In particular, the mechanical-structural description exposes a relation among the relevant components that the matrix description does not display as such: the outputs of some simple networks determine the weights of others. In this respect, our reconstruction develops the functional description into a more explicit account of the system's mechanistic organization (Machamer, Darden, & Craver, 2000; Piccinini & Craver, 2011). Output-weight interconnections thus identify an organizational relation among components that is essential to the production of dynamic transforming concepts.

## 3 Sequence-level Interactive Dynamic Parallel Processing in the Transformer

The preceding analysis has yielded two functionally equivalent descriptions of the Transformer that make different aspects of its organization salient (§ 2.6). At the functional level, it is a system of successive processing modules, each consisting of one or more processes, where each process is a succession of operations on matrices whose rows are vector representations of the prompt tokens. At the mechanical-structural level, it is a system composed of groups of $n_{max}$ simple neural networks, where the $i$-th ($1 \leq i \leq n \leq n_{max}$) network of each group performs the processing related to the $i$-th prompt token. As established in § 2.5, in the second description each group of simple networks is identified with either a parameterized operation or a structural operation of the type $Mult\{c\}$ or $Add\{\mathfrak{M}\}$. In addition, groups of simple networks are linked by: (a) standard output-input interconnections, in which the output of a network in one group is the input of the network corresponding to the same token in the following group; (b) two other types of output-input interconnections, *operational* and *routing*: operational interconnections correspond to $ReLU$, $Softmax$, or $Norm$, whereas routing interconnections correspond to the formal operation $Concat$; (c) *output-weight interconnections of four different types in the mathematical configuration illustrated here.*

Although Parallel Distributed Processing (PDP) is preserved within each constituent simple neural network, the notion of parallel processing introduced here concerns a different level of organization. In PDP, many units operate simultaneously and jointly transform an input pattern into an output pattern (Rumelhart, McClelland, & the PDP Research Group, 1986; Rumelhart, Hinton, & Williams, 1986). In the Transformer, by contrast, the relevant components at the level considered here are groups of simple neural networks, each network processing the representation of one token, while the groups are coordinated through standard and non-standard interconnections. Sequence-level parallel processing is therefore an emergent property of the organization and interaction of these groups, rather than simply of parallel activity within each constituent network. Output-weight interconnections are especially important because they make this sequence-level processing dynamic and, when interwoven in attention head modules, interactive.

Dynamic neural networks constitute a broader established research area (Han et al., 2022). The processing organization identified here is more specific, combining sequence-level parallelism, cross-token interaction, and input-dependent transformations within ordinary Transformer inference.

We call this new type of processing SIDPP—*Sequence-level Interactive Dynamic Parallel Processing*—, where: (i) SPP—*Sequence-level Parallel Processing*—means that the input sequence is processed in parallel, producing a corresponding output sequence. The elements of these sequences are vector representations of the prompt's tokens that, from a mathematical point of view, are treated as rows of matrices; (ii) SIPP—*Sequence-level Interactive Parallel Processing*—is an SPP in which at least one vector representation in the output sequence depends on a vector representation of a different token in the input sequence; (iii) SDPP—*Sequence-level Dynamic Parallel Processing*—is an SPP in which the output sequence depends on parameterized transformations whose parameters in turn depend on the input sequence. The parameters of these transformations are called *dynamic*.

*Figure 3. Sequence-level Interactive Dynamic Parallel Processing (SIDPP) in a bidirectional attention head with a three-token input. The token representations are processed in parallel at the sequence level. Processing is dynamic because outputs generated from the current input determine the weights of receiving networks through output-weight interconnections, and interactive because these interconnections are interwoven across token-processing streams, allowing the processing of each token representation to depend on representations of the other tokens. Both the bidirectional attention head and the three-token input are chosen solely for illustrative clarity; the same processing principles apply to autoregressive attention heads and to input sequences of other lengths, subject to the corresponding constraints on cross-token interactions (see OR1, Figs. S10–S12e).*

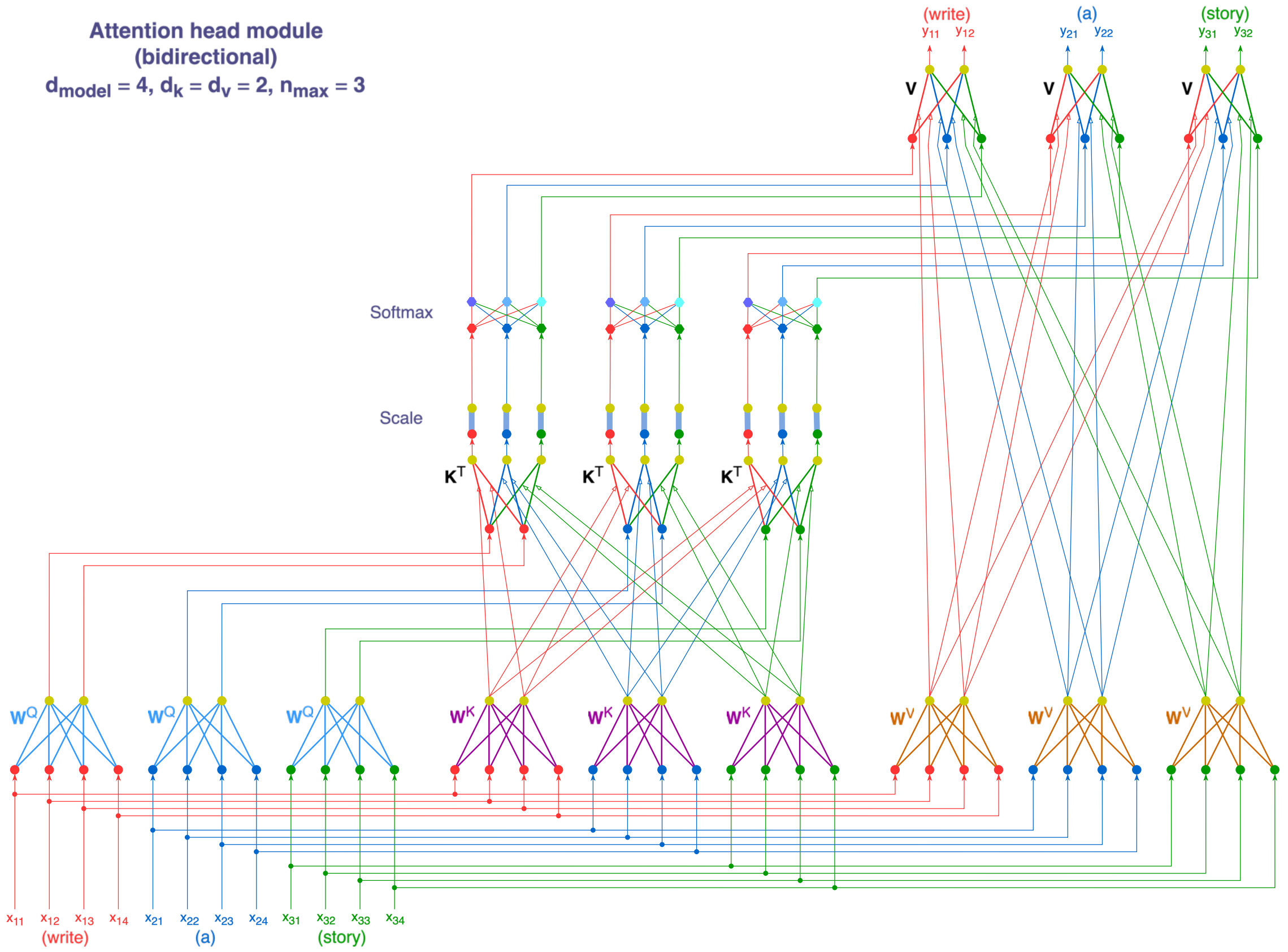

Sequence-level parallel processing (SPP) is the type of processing that takes place in all parameterized and structural operations of the Transformer, from the operation that performs positional encoding to the one that produces the probability distributions by applying the function $softmax$. Normalization itself does not introduce cross-token interaction: it operates independently on the features of each token. In the addition operations of residual connection modules, processing is dynamic (SDPP), but the addition is not locally interactive: its inputs can already contain information about other tokens, which the addition itself does not newly integrate. Finally, in each attention head module, sequence-level parallel processing is both dynamic and interactive (SIDPP).

Considering the Transformer as a system of interconnected groups of simple neural networks, we see that the different types of processing depend on different morphological characteristics of the system. First, sequence-level parallel processing (SPP) is produced (i) by the groups of simple neural networks identified with the Transformer's parameterized operations or with structural operations of the type $Mult\{c\}$ or $Add\{\mathfrak{M}\}$, and (ii) by the groups of operational blocks that realize the operational interconnections corresponding to $ReLU$, $Softmax$, or $Norm$. Second, dynamic processing (SDPP) is made possible by output-weight interconnections, which are present both in the residual connections (OR1, Figs. S8–S8c) and in the attention head modules (OR1, Figs. S10–S12e). Finally, interactive processing (SIPP) arises because the two types of output-weight interconnections in the attention head modules are *interwoven* links between two groups of simple networks: they link networks in one group to networks in the other group that do not correspond to the same token. By contrast, all other interconnections are non-interwoven, because they link networks in one group to networks in the other group corresponding to the same token.

Interwoven interconnections ensure inter-token interaction between two groups of simple neural networks, and they occur in two different ways. Output-weight interconnections of the first type (OR1, § S3.2) make the $j$-th feature ($1 \leq j \leq q$) of the $i$-th token ($1 \leq i \leq n \leq n_{max}$) depend on the $j$-th features of all tokens; they are therefore feature ⇛ feature links (OR1, Figs. S10d, S11d). The output-weight interconnections of the second type (OR1, § S3.2) make the $j$-th feature ($1 \leq j \leq n \leq n_{max}$) of the $i$-th token ($1 \leq i \leq n \leq n_{max}$) depend on all features of the $j$-th token; they are therefore token ⇛ feature links (OR1, Figs. S10e, S11e). Finally, we note that, in attention head modules, it is precisely the succession of these two types of interaction,[3] implemented by the corresponding interwoven output-weight interconnections, that constitutes the *causal mechanism* by which a token attends to other tokens.

## 4 The Relationship Between Dynamic and Static Parameters in the Transformer

Section 3 showed that SDPP depends morphologically on output-weight interconnections, through which dynamic parameters are generated during inference. We now compare the number of these parameters with the number of static parameters fixed through training.

As specified in § 2.2 (see also OR1, § S1), *dynamic* parameters are the elements of the matrices and vectors that define dynamic transforming concepts. Their number $D(n)$ can be calculated for a given Transformer as a function of the prompt length $n$. Let $N$ be the number of transformer layers, $h$ the number of attention head modules per layer, $d_k$ the dimension of the query and key vectors, $d_v$ the dimension of the value vectors, and $d_{model}$ the dimension of the embedding vectors. We then obtain:

$$D(n) = Cn\,,$$

where

$$C = N[h(d_k + d_v) + 2d_{model}]\,.$$

Thus, $C$ is a system-specific constant, whereas $D(n)$ grows linearly with prompt length $n$, up to $n_{max}$.

The *static* parameters are the parameters of the static transforming concepts defined in § 2.1 (see also OR1, § S1). They are trainable parameters defining static transforming concepts, and their number, $S$, is a system-specific constant which, if the model uses sinusoidal encoding,[4] is given by:

$$S = d_{model}d_{voc} + N\big[2h\,d_{model}(d_k + d_v) + 2d_{model}d_f + \big(d_f + 5d_{model}\big)\big]\,,$$

where $d_{voc}$ is the number of tokens in the vocabulary and $d_f$ is the hidden dimension of the feed-forward modules. We also note that, if the system uses weight tying between the input embedding matrix $\boldsymbol{E}$ and the output projection matrix $\boldsymbol{W}^L$ (with $\boldsymbol{W}^L = \boldsymbol{E}^\top$), $S$ is identical to the total number $A$ of trainable parameters. Otherwise,

$$A = S + d_{voc}d_{model}\,,$$

where $d_{voc}d_{model}$ is the number of elements of the embedding matrix $\boldsymbol{E}$.

---

[3] Token ⇛ feature interaction occurs first, followed by feature ⇛ feature interaction.

[4] If, instead, the $n_{max}$ positional encoding vectors are trainable, they too define static transforming concepts, and the $n_{max}d_{model}$ parameters that constitute them must therefore be added to $S$.

The preceding formula describes the simplified sinusoidal-position, post-LN configuration illustrated in the diagrams, without attention-projection biases or a final LayerNorm. For the reconstructed GPT-2/GPT-3 configuration, including learned positional embeddings, attention-projection biases, the final LayerNorm and tied token-embedding/logit weights, the static-parameter count becomes:

$$S_{GPT} = d_{model}(d_{voc} + n_{max} + 2) + N\big[2hd_{model}(d_k + d_v) + 2d_{model}d_f\big]$$
$$+N\big[d_f + 6d_{model} + h(2d_k + d_v)\big].$$

Relative to the simplified count, this adds learned positional embeddings, $Q/K/V$ and output-projection biases, and the final LayerNorm parameters (Radford et al., 2019; Brown et al., 2020).

**Proposition 2 — Dynamic-parameter scaling.** *For a fixed Transformer architecture, the number $D(n)$ of dynamic parameters generated by a prompt of length $n$ is a linear function of $n$, whereas the number $S$ of static parameters is constant.*

Let $n^*$ denote the value of $n$ for which $D(n) = S$. Since $D(n) = Cn$, it follows from the definition of $n^*$ that

$$D(n^*) = S = Cn^*.$$

Hence,

$$n^* = \frac{S}{C}.$$

Thus, $n^*$ depends only on the model-specific constants $S$ and $C$, and is therefore determined once the architecture is fixed. Moreover,

$$D(1) = C = \frac{S}{n^*}.$$

This gives $C$, equivalently $S/n^*$, a direct interpretation: it is the number of dynamic parameters generated by a one-token prompt. Since

$$D(n) = nD(1) = n\frac{S}{n^*},$$

it follows that

$$\frac{D(n)}{S} = \frac{n}{n^*}.$$

The ratio between dynamic and static parameters is therefore equal to the ratio between the actual prompt length $n$ and the prompt length $n^*$ that would make $D(n) = S$.

The number of dynamically generated parameters should be distinguished from the total number of parameters used by the active instances of a transformation, counting repeated uses of the same parameter. In the mechanical realization of dense attention, the number of active copies of a given neural network grows linearly with sequence length, whereas the total number of parameters used by these copies grows quadratically. This is because the same dynamically generated parameters can be used by multiple copies of the network. It is the number of distinct dynamically generated parameters, rather than the total number of their uses, that is compared with the number of static parameters, providing a measure of the relative size of the dynamic and static parameter structures.

**Example: GPT-3-175B**

For this example we reconstruct the GPT-2/GPT-3 parameterization described by Brown et al. (2020), including learned positional embeddings, projection biases and the final LayerNorm. This is a count inferred from the published architectural descriptions, not one obtained by inspecting the original GPT-3 checkpoint. The locally banded sparse attention pattern restricts the positions used in attention but does not eliminate generation of $K$ and $V$ representations for the tokens; consequently it does not alter the count of distinct dynamically generated parameters below (Child et al., 2019; Brown et al., 2020).

$$d_{voc} = 50{,}257\,, \qquad d_{model} = 12{,}288\,, \qquad N = h = 96\,,$$
$$d_k = d_v = 128\,, \qquad d_f = 4d_{model} = 49{,}152\,.$$

It follows:

$$C = 4{,}718{,}592\,,$$
$$S = 174{,}604{,}259{,}328\,.$$

A prompt of $n = 200$ tokens therefore generates

$$D(200) = 943{,}718{,}400$$

dynamic parameters, with

$$\frac{D(200)}{S} \approx 0.0054\,,$$

or approximately 0.54% of the number of static parameters. At $n = 2048 = n_{max}$, the maximum context length of GPT-3-175B,

$$D(2048) = 9{,}663{,}676{,}416\,,$$

and

$$\frac{D(2048)}{S} \approx 0.0553\,,$$

or approximately 5.53% of $S$. For the same model,

$$n^* = \frac{174{,}604{,}259{,}328}{4{,}718{,}592} \approx 37{,}003.47 \text{ tokens.}$$

Thus, $n^* > n_{max} = 2048$: the prompt length at which $D(n) = S$ was not in fact reachable by GPT-3-175B. For this model, $n^*$ is therefore a formal threshold determined by the scaling relation, rather than an input length that the model can actually process.

These quantities compare numbers of parameters only. The ratio $D(n)/S$ should therefore not be interpreted by itself as a measure of the relative information content, causal importance, or computational contribution of dynamic and static parameters. Prompt length determines the number of dynamic parameters generated, while the particular input sequence determines, through the processing described in § 2, the values those parameters take. Thus, two different inputs of the same length generate the same number of dynamic parameters but may produce different causal effects or computational contributions.

We call this structural dependence of the Transformer's processing on the information supplied by the prompt *strong prompt sensitivity*. Section 5 examines a striking behavioral phenomenon for which strong prompt sensitivity provides a natural explanatory framework.

## 5 Strong Prompt Sensitivity and GPT-4.5's Turing Test Success

The striking behavioral phenomenon for which strong prompt sensitivity provides a natural explanatory framework is GPT-4.5's recent Turing test success (Jones & Bergen, 2025, 2026a; cf. Giunti, 2026). Jones and Bergen conducted four three-player tests, using two populations of human participants, in which a human interrogator conversed with one human and GPT-4.5. In two tests, GPT-4.5 received a 187-word system prompt, PERSONA, specifying the personality and linguistic style to adopt; in the other two, it received only a 21-word NO-PERSONA prompt instructing it to convince the interrogator that it was human (see OR1, Tables S1–S2).

The effect was substantial in both populations. With PERSONA, GPT-4.5 was judged human in 75.5% of the Prolific tests and 69.2% of the UCSD tests; without PERSONA, the corresponding rates fell to 42.1% and 27.7% (OR1, Table S1). The PERSONA success rates were significantly above 50% in both populations, whereas the NO-PERSONA rate was significantly below 50% for the UCSD population but not for the Prolific population.[5] Thus, GPT-4.5 with PERSONA passed both Turing tests, whereas without it the model clearly failed one and did not clearly pass the other. A change of only 166 words in the system prompt therefore produced differences of 33.4 and 41.5 percentage points in the two populations.

Within the present framework, there is a direct explanatory route from prompt variation to behavioral variation: changing the prompt changes the input-dependent parameter structures generated during inference, and hence the dynamic transformations applied at subsequent stages; if prompt length changes, the number of dynamic parameters changes as well. These differences can propagate through subsequent processing and contribute to markedly different outputs from the same trained model.

The GPT-4.5 case is not offered as an empirical test or validation of the framework, but as a behavioral phenomenon for which strong prompt sensitivity provides a viable explanatory framework.

## 6 Implications for AI Research

The preceding analysis has provided a functional and a mechanical-structural account of the Transformer's processing during inference and has identified SIDPP as the type of processing generated by this organization. These results also have implications for AI research.

The central thesis of this work is that, during inference, the Transformer dynamically constructs and applies prompt-dependent transformations of token representations. At the mechanical-structural level, this is made possible by output-weight interconnections between groups of simple networks. These interconnections generate dynamic parameters whose number depends linearly on prompt length and whose values depend on the particular input sequence (§ 4). This structural dependence on the prompt provides a viable explanatory framework for the behavioral phenomenon discussed in § 5. On this basis, we identify two research directions.

---

[5] The $p$-values reported here were calculated exactly by us using the binomial distribution, on the basis of the data reported in OR1, Table S1: PERSONA, $p = 4.591 \times 10^{-5}$ (UCSD) and $p = 2.159 \times 10^{-10}$ (Prolific), right-tailed; NO-PERSONA, $p = 2.111 \times 10^{-4}$ (UCSD) and $p = 0.1034$ (Prolific), left-tailed. They are consistent with the $p$-values reported by Jones and Bergen (2026a, last two paragraphs on p. 4), but differ slightly because the latter are approximate values calculated from the reported $z$-values.

### *6.1 Interpretability, Control, and Predictability*

A first line of research concerns improving our understanding of what the Transformer actually does during inference and, on this basis, making progress toward three closely related goals: interpretability, control, and predictability. Transparency in complex computational systems need not be all-or-nothing: different forms and degrees of transparency may be relevant to different epistemic tasks (Creel, 2020). Our mechanical-structural description does not make the Transformer transparent in every respect, but it makes one important class of internal dependencies explicit.

A large part of contemporary *interpretability* research investigates learned weights, neurons, features, circuits, or attribution patterns in the trained model (Montavon et al., 2018; Samek et al., 2019; Olah et al., 2020; Elhage et al., 2021). Such analyses are important, but they do not directly address a different question: how prompt-dependent transformations are generated and used during a particular inference. From the perspective developed here, inference configures a prompt-specific mechanical-structural state through output-weight interconnections. The relevant dependency chain runs from current input representations to dynamically generated parameters; from those parameters to the dynamic transformations they define; from those transformations to subsequent token representations; and ultimately to the system's output.

Focusing on these dependencies fits naturally with accounts of scientific understanding that emphasize the representation of dependency relations (Dellsén, 2020). More generally, the epistemic value of a model depends not simply on its simplicity or transparency, but on whether the relation between the model and its target is adequately supported (Sullivan, 2022). In our case, the relevant target is a specific aspect of the Transformer's own inference mechanism, and the mechanical-structural reconstruction is derived from the mathematical operations described in § 2.

This does not amount to a complete causal explanation of any particular output. Rather, it identifies a class of dependencies that such an explanation would need to take into account. In particular, attention is not merely a matrix of weights to be inspected *a posteriori*: § 3 showed how it is produced through the succession of token ⇛ feature and feature ⇛ feature interactions, implemented by interwoven output-weight interconnections. This perspective therefore avoids simply identifying high attention weights with the causal reasons for an output—a point that has been central to the debate over the explanatory status of attention maps (Jain & Wallace, 2019; Wiegreffe & Pinter, 2019).

The same analysis suggests a more specific approach to control. If prompt-dependent transformations contribute substantially to inference, intervening on the prompt is not merely a way of supplying different information: it also changes the dynamic parameters and hence the mechanical-structural configuration instantiated during processing. A possible research program would therefore seek to identify which prompt interventions reliably induce, suppress, or stabilize particular dynamic transformations, rather than treating prompt effects only at the behavioral level.

Finally, strong prompt sensitivity has consequences for predictability. Global characterization of a trained model may be insufficient to predict its behavior across inputs if relatively small input changes produce substantially different dynamic configurations. Predictability should therefore also be investigated locally, by determining how dynamic parameters and the transformations they define vary under controlled changes to the prompt, and how those variations propagate to the resulting token representations and outputs.

Thus, the contribution of the present analysis to interpretability is deliberately limited but specific: it does not provide complete transparency or automatically explain individual responses. It exposes a class of prompt-dependent structural and causal dependencies that a purely static view of the trained model does not make equally salient, thereby providing a basis for further work on understanding, control, and predictability.

### *6.2 Neuromorphic Constructions: Energy Efficiency, Robotics, and Edge Computing*

A second line of research concerns the physical construction of Transformer systems. Current large Transformers are computationally implemented by programming their functional architecture and numerically evaluating their operations on parallel digital hardware, an approach associated with substantial computational and energy costs (Strubell et al., 2019; Patterson et al., 2021). The mechanical-structural description developed here suggests a different route: could the same functional architecture be mechanically realized in neuromorphic form and, on that basis, physically constructed?

Our analysis identifies the Transformer with groups of simple networks connected not only by output-input interconnections, but also by output-weight interconnections. The latter allow the outputs of some networks to configure the functional units of others during inference by determining their weights. A physical construction based on a neuromorphic mechanical realization would therefore not need merely to reproduce the Transformer's matrix operations numerically; it could instead seek physical mechanisms in which distributed functional units are dynamically configured through such local interconnections. Changing the mechanical realization would not in itself require changing the Transformer's standard mathematical instantiation: the same operations and functions could in principle be assigned to a neuromorphic realization and guide a different physical construction.

Neuromorphic engineering has long explored systems based on distributed, local computation rather than conventional centralized numerical processing (Mead, 1990; Indiveri & Liu, 2015; Davies et al., 2018). More recently, neuromorphic and analog hardware for self-attention has been explored with the aim of reducing the computational and energetic costs of Transformer processing (Leroux, Finkbeiner, & Neftci, 2025). From the present perspective, a further question is whether dynamically configurable distributed functional units, coordinated through output-weight interconnections, can provide a basis for neuromorphic constructions of Transformer-like processing.

Such an approach could be relevant especially where energy, latency, and connectivity impose strong constraints. In robotics, prompt- or input-dependent transformations could provide rapid adaptation to changing environments without requiring retraining: the relevant transformations would be generated during processing itself. In edge computing, the same principle could support forms of local, input-dependent adaptation without updating the model's trained parameters, potentially reducing dependence on continuous access to centralized systems (Shi et al., 2016; Satyanarayanan, 2017).

These possibilities remain speculative: the mechanical-structural description does not by itself establish that such a physical construction is technologically feasible or more energy-efficient. It does, however, provide a set of functional and organizational constraints for exploring this possibility. Rather than treating the Transformer only as a large collection of matrix operations to be simulated, neuromorphic research could investigate how its characteristic dynamics might be mechanically realized through groups of simpler processing units containing distributed functional units that are dynamically configured through output-weight interconnections, and how such a realization could guide physical construction.

The possibility of engineered neuromorphic constructions raises a related but distinct question: whether biological neural mechanisms can support a neural construction of the processing organization identified here. We turn to the neural constructibility of SIDPP and its implications in § 7.

## 7 Neural Constructibility of SIDPP and Conjecture T

We now turn from engineered neuromorphic constructions to the question of neural constructibility. We first ask whether known biological mechanisms can, in principle, support a neural construction of a form of SIDPP (§ 7.1). We then formulate Conjecture T as a stronger conditional research hypothesis concerning the cortical systems involved in human language processing (§ 7.2).

### *7.1 The In-Principle Neural Constructibility of SIDPP*

We have shown (§§ 2.6 and 3) that output-weight interconnections are the mechanical-structural relation that makes sequence-level dynamic parallel processing (SDPP) possible. An output-weight interconnection links an output unit of one network to a connection in another network, so that the former's output determines the latter's weight (Figure 2). Speaking of such a link as modifying or configuring a connection is, however, still shorthand: what does it mean operationally for a connection to have a weight that can be supplied by another unit?

To answer this question, we refine the decomposition of the simple networks introduced in § 2.3. At the level of decomposition adopted there, input units merely route incoming signals, whereas output units perform the relevant transformations. In a dense linear network (OR1, Fig. S17), each output unit computes the weighted sum of all signals arriving from the input units; in a one-to-one multiplicative or additive network, each output unit respectively multiplies the incoming signal by the weight of its connection or adds that signal to the weight.

For dense linear networks, however, the weighted sum can itself be decomposed into two successive stages. First, the signals coming from the input units are multiplied in parallel by the weights of the corresponding connections; second, the resulting weighted signals are summed. The output unit can therefore be treated as a purely integrative functional unit that performs only the second operation, while each connection contains a distinct multiplicative functional unit that performs the first (OR1, Fig. S18). There are thus as many multiplicative functional units as there are connections.

The preceding decomposition also suggests a further step that enables us to make explicit the mechanism by which a connection can be modified or configured (OR1, Fig. S19). For each connection from input unit to output unit, we now conceive the multiplicative functional unit located on that connection as having two inputs and one output, rather than as multiplying its incoming signal by a fixed parameter. Its output is connected to the output unit; its first input receives the signal routed by the input unit; its second input supplies the value by which that signal is multiplied and therefore determines the weight of the connection. Changing the connection's weight now has a precise operational meaning: supplying a new value to this second input. An output-weight interconnection realizes exactly this relation by linking an output unit of another network to the second input of the functional unit located on the receiving connection.

The same analysis applies, with minor modifications, to one-to-one multiplicative and additive networks. In a one-to-one multiplicative network, the functional units consist only of the multiplicative units located on the connections, while the output units merely receive signals already weighted by those units. The same applies to one-to-one additive networks, except that the functional units located on the connections are additive rather than multiplicative.

This refined decomposition also makes a component-by-component comparison with biological neural networks possible (OR1, Fig. S18). An *output unit*, insofar as it integrates converging signals, corresponds functionally and morphologically to the *soma* of a biological neuron; an *input unit*, insofar as it routes a signal toward different output units, corresponds to the *terminal arborization of the axon* of a biological neuron; the connections between an input unit and different output units correspond to the links between the branches of the axon and the dendrites of different neurons; and the *functional units located on those connections* correspond to the *synapses* between the axon terminals and the dendritic terminals, which establish those links.

The correspondences established above are, so far, limited to simple networks with fixed weights; they do not yet include simple networks with dynamic weights. In the refined mechanism described above, a dynamic weight is realized by a multiplicative functional unit located on a connection and provided with an additional input: the first input receives the signal to be multiplied, while the additional input supplies the value that determines the weight of the connection. To extend the correspondence to networks with dynamic weights, we must therefore ask whether biological neural networks contain a mechanism morphologically and functionally correspondent to a multiplicative functional unit with such an additional input. If so, the correspondence could be extended from simple networks with fixed weights to simple networks with dynamic weights and, consequently, to output-weight interconnections.

Keeping in mind the correspondences established above, we are therefore looking for a mechanism morphologically similar to a link between a second neuron and a synapse connecting an axonal branch of a first neuron to a dendrite of a third. From a functional point of view, moreover, the link should ensure modulation of the signal coming from the first neuron, and this modulation should be caused by the signal sent by the second neuron through the hypothesized link.

At least two known biological mechanisms provide relevant morphological and functional analogues. The most direct is provided by axo-axonic synapses (OR1, Fig. S20), in which one axon contacts the presynaptic axon or terminal of another and modulates its signaling or neurotransmitter release.[6] A second analogue is provided by local interactions between neighboring axo-dendritic synapses on the same dendrite. Here, the relevant morphological condition is the close proximity of two synapses formed by different axons on the same dendritic segment, which allows activity at one synapse to influence postsynaptic signals and plasticity at the other.[7] Unlike the axo-axonic case, however, morphology alone does not allow us to distinguish which of the two axons plays the primary and which the modulatory role; this can be established only functionally. More generally, synaptic mechanisms are central to the plasticity and adaptability of neural circuits.[8]

These biological mechanisms need not reproduce the standard mathematical operation by which a Transformer connection applies its weight. For neural constructibility, what matters is that one local interaction can determine a variable parameter that modulates how another connection transforms its incoming signal; such a mechanism can therefore play the functional role of an output-weight interconnection even if the modulation is not ordinary multiplication.

[6] For axo-axonic synapses, especially axon-on-terminal contacts and their presynaptic modulation of neurotransmitter release, see Cover and Mathur (2020).

[7] For experimental evidence of local cooperativity between neighboring dendritic spines affecting postsynaptic signals and plasticity, see Weber et al. (2016).

[8] For a review of synaptic plasticity and circuit adaptation, see Citri and Malenka (2008).

Taken together, these correspondences extend from simple networks with fixed weights to simple networks with dynamic weights and, consequently, to output-weight interconnections. Since output-weight interconnections provide the mechanical basis of sequence-level dynamic parallel processing (SDPP), they show that this type of processing is, at least in principle, neurally constructible. Moreover, the correspondences established for the other components and interconnections of simple networks support the same in-principle conclusion for sequence-level parallel processing (SPP) and sequence-level interactive parallel processing (SIPP). It is therefore possible, at least in principle, to neurally construct a form of SIDPP.

### *7.2 Conjecture T as a Research Hypothesis*

How might such a neural construction of SIDPP be produced? We know that the functional architecture of the Transformer produces SIDPP, and § 7.1 has shown that a form of SIDPP is neurally constructible in principle. To our knowledge, no alternative architecture has yet been shown to produce SIDPP in the specific sense defined here. This does not establish that cortical language systems instantiate a Transformer-like architecture. It does, however, provide an abductive motivation for considering such an architecture as a scientifically plausible candidate (Harman, 1965; Lipton, 2004; Magnani, 2001). The morphological and functional correspondences identified in § 7.1 further show that the relevant components and relations can, in principle, be mapped onto known biological mechanisms, as required for a mechanistic model (Kaplan & Craver, 2011). On this basis, we formulate Conjecture T:

**Conjecture T.** If SIDPP is the type of processing actually produced by the cortical systems involved in human language processing, we conjecture that it is produced by a neural construction of a functional architecture relevantly similar to that of the Transformer.

We do not assess Conjecture T in the present paper; rather, we argue that it is a scientifically plausible research hypothesis that warrants further investigation. Assessing whether it is supported or challenged by the known morphology and functioning of cortical systems involved in human language processing lies beyond the scope of the present paper and will be the subject of a separate study, Part 2, specifically devoted to this question.

## 8 Conclusion

This paper has argued that Transformer inference should not be understood solely as the application of transformations fixed through training. During inference, static transforming concepts are also used to construct input-dependent matrices and vectors that define further transformations. Inference therefore involves both adapting previously acquired transformations to the current input and applying the resulting dynamic transformations.

By redescribing the standard matrix operations of the Transformer in terms of groups of simple neural networks and their interconnections, we have developed a functionally equivalent mechanical-structural reconstruction of its inference organization (Proposition 1). This reconstruction makes explicit the role of output-weight interconnections, through which the outputs of some networks determine the weights of others. Together with sequence-level parallelism and cross-token interaction, these relations give rise to Sequence-level Interactive Dynamic Parallel Processing (SIDPP).

We have also shown that, for a fixed Transformer architecture, the number of dynamic parameters generated during inference grows linearly with prompt length, while the number of static parameters remains constant (Proposition 2). This relation determines both the number of dynamic parameters generated by a prompt of a given length and their numerical ratio to the static parameters. More generally, prompt length determines how many dynamic parameters are generated, while the particular input sequence determines the values they take. We have called this structural dependence on the prompt strong prompt sensitivity. GPT-4.5's recent Turing test results provide a striking behavioral phenomenon for which this framework offers a viable explanatory route, without constituting an empirical test or validation of the framework itself.

Finally, the mechanical-structural description suggests further directions for AI research, including interpretability, control, predictability, and neuromorphic physical constructions. At the biological level, we have argued that known neural mechanisms provide the functional components required for the in-principle neural constructibility of a form of SIDPP. This result does not establish Conjecture T. It motivates it as a scientifically plausible research hypothesis: if SIDPP is the type of processing actually produced by cortical systems involved in human language processing, it may be produced by a neural construction of a functional architecture relevantly similar to that of the Transformer. Assessing this hypothesis lies beyond the scope of the present paper and is reserved for the separate study, Part 2.

## SUPPLEMENTARY MATERIAL (OR1)

# TECHNICAL APPENDIX

### *S1 Transforming Concepts as Transformations Defined by Matrices or Vectors of Model Parameters: Technical Details*

A *transforming concept* is any transformation of vector representations of tokens defined by a matrix or vector of model parameters. A matrix $\boldsymbol{M} \in \mathbb{R}^{q \times p}$ of model parameters defines the linear transformation $\mathcal{L}[\boldsymbol{M}]: \mathbb{R}^q \to \mathbb{R}^p, \mathcal{L}[\boldsymbol{M}](\boldsymbol{u}) \coloneqq \sum_{i=1}^{q} u_i \, r_i(\boldsymbol{M})$, where $u_i$ is any element of the vector $\boldsymbol{u} \in \mathbb{R}^q$ and $r_i(\boldsymbol{M})$ is the $i$-th row of $\boldsymbol{M}$. Similarly, a vector $\boldsymbol{v} \in \mathbb{R}^q$ of model parameters defines the transformation $+[\boldsymbol{v}] : \mathbb{R}^q \to \mathbb{R}^q$ or $\circ[\boldsymbol{v}]: \mathbb{R}^q \to \mathbb{R}^q$, depending on whether it is added to or multiplied componentwise by an arbitrary vector $\boldsymbol{u} \in \mathbb{R}^q$.

A *static transforming concept* is a transforming concept whose defining matrix or vector consists of trainable parameters.

The matrices of trainable parameters that define static transforming concepts (that is, linear transformations of vector representations of tokens) are as follows: for each attention head $i$, the three matrices ${\boldsymbol{W}_i}^Q$, ${\boldsymbol{W}_i}^K$, ${\boldsymbol{W}_i}^V$ (see Fig. S4); for each multi-head attention module, the output projection matrix $\boldsymbol{W}^O$ (see Fig. S3); for each feed-forward network module, the matrices $\boldsymbol{W}_1$ and $\boldsymbol{W}_2$ of the two linear layers (see Fig. S2); and, in the final module, the matrix $\boldsymbol{W}^L$ that produces the logits (see Fig. S1, Forecast module).

The embedding matrix (or table) $\boldsymbol{E} \in \mathbb{R}^{d_{voc} \times d_{model}}$ (see Fig. S1) is itself composed of trainable parameters, but it does not define a linear transformation of vector representations of tokens. By contrast, its rows *are* the initial vector representations of each token. Moreover, if the model uses weight tying ($\boldsymbol{E}^\top = \boldsymbol{W}^L$) between the embedding matrix $\boldsymbol{E}$ and the final projection matrix $\boldsymbol{W}^L$, the parameters of $\boldsymbol{E}$ are identical to those of $\boldsymbol{W}^L$ and therefore should not be counted separately when calculating the number of trainable parameters of the model.

In each feed-forward network module, there are two bias vectors, $\boldsymbol{c} \in \mathbb{R}^{d_f}$ and $\boldsymbol{d} \in \mathbb{R}^{d_{model}}$ (see Fig. S2). There is another bias vector, $\boldsymbol{b} \in \mathbb{R}^{d_{model}}$, in each layer normalization module, and that module also contains the corresponding scaling vector $\boldsymbol{a} \in \mathbb{R}^{d_{model}}$ (see Fig. S1). All the elements of the bias and scaling vectors are trainable parameters of the model, and these are the vectors that define static transforming concepts: the bias vectors in the feed-forward network modules and those in the layer normalization modules define static transforming concepts of type $+[\boldsymbol{v}]$; the scaling vectors (in the layer normalization modules), by contrast, define static transforming concepts of type $\circ[\boldsymbol{v}]$.

The positional encoding vectors, of dimension $d_{model}$, are added to the matrix $\boldsymbol{X}$ of embedding vectors corresponding to the tokens in the prompt, thereby forming the input matrix $\boldsymbol{I}$ of the first layer of the Transformer stack, which consists of $N$ structurally identical layers (see Fig. S1). There are $n_{max}$ positional encoding vectors, where $n_{max}$ is the size of the model's context window. If the model uses sinusoidal positional encoding, the elements of these vectors are constants fixed independently of training, and the positional encoding vectors therefore do not define static transforming concepts. In some models, however, these vectors instead consist of trainable parameters. In that case, the positional encoding vectors, too, define static transforming concepts of type $+[\boldsymbol{v}]$.

A *dynamic transforming concept* is a transforming concept whose defining matrix or vector is generated during inference. The elements of these matrices and vectors are the model's *dynamic parameters*.

The generated matrices that define dynamic transforming concepts are as follows: for each attention head $i$, the transpose of the key matrix, $\boldsymbol{K}_i^\top$, and the value matrix, $\boldsymbol{V}_i$.

The generated vectors that define dynamic transforming concepts are as follows: the output vectors of the multi-head attention modules and those of the feed-forward network modules. Through the corresponding residual connections, they define transformations of type $+[\boldsymbol{v}]$ (see Fig. S1).

### S2 The Transformer's Matrix Operations: Parameterized, Structural, and Formal

The *parameterized operations* are functions of the four types defined below:

1) $Add[\boldsymbol{v}](\boldsymbol{X})$: $\mathbb{R}^{n\times q} \to \mathbb{R}^{n\times q}$, with $q = d_{model}$ or $q = d_f$,[9] defined by:

$$Add[\boldsymbol{v}](\boldsymbol{X}) \coloneqq \begin{pmatrix} +[\boldsymbol{v}]\big(r_1(\boldsymbol{X})\big) \\ +[\boldsymbol{v}]\big(r_2(\boldsymbol{X})\big) \\ \vdots \\ +[\boldsymbol{v}]\big(r_n(\boldsymbol{X})\big) \end{pmatrix}$$
, where $\boldsymbol{v} \in \mathbb{R}^q$ is a vector of the model's bias parameters and $r_i(\boldsymbol{X})$ is the $i$-th row of $\boldsymbol{X}$;

2) $Mult[\boldsymbol{v}]$: $\mathbb{R}^{n\times d_{model}} \to \mathbb{R}^{n\times d_{model}}$, defined by:

$$Mult[\boldsymbol{v}](\boldsymbol{X}) \coloneqq \begin{pmatrix} \circ[\boldsymbol{v}]\big(r_1(\boldsymbol{X})\big) \\ \circ[\boldsymbol{v}]\big(r_2(\boldsymbol{X})\big) \\ \vdots \\ \circ[\boldsymbol{v}]\big(r_n(\boldsymbol{X})\big) \end{pmatrix}$$
, where $\boldsymbol{v} \in \mathbb{R}^{d_{model}}$ is a vector of the model's scaling parameters and $r_i(\boldsymbol{X})$ is the $i$-th row of $\boldsymbol{X}$;

3) $Add[\boldsymbol{M}]$: $\mathbb{R}^{n\times d_{model}} \to \mathbb{R}^{n\times d_{model}}$, defined by:

$$Add[\boldsymbol{M}](\boldsymbol{X}) \coloneqq \begin{pmatrix} +[r_1(\boldsymbol{M})]\big(r_1(\boldsymbol{X})\big) \\ +[r_2(\boldsymbol{M})]\big(r_2(\boldsymbol{X})\big) \\ \vdots \\ +[r_n(\boldsymbol{M})]\big(r_n(\boldsymbol{X})\big) \end{pmatrix}$$
, where $r_i(\boldsymbol{X})$ and $r_i(\boldsymbol{M})$ are the $i$-th rows of $\boldsymbol{X}$ and $\boldsymbol{M}$, respectively, and $\boldsymbol{M} \in \mathbb{R}^{n\times d_{model}}$ is the output matrix of any attention or feed-forward module or, if the positional encoding vectors are trainable, $\boldsymbol{M} \in \mathbb{R}^{n_{max}\times d_{model}}$ is the matrix whose rows are those vectors;

4) $Mult[\boldsymbol{M}]$: $\mathbb{R}^{n\times q} \to \mathbb{R}^{n\times p}$, defined by:

$$Mult[\boldsymbol{M}](\boldsymbol{X}) \coloneqq \begin{pmatrix} \mathcal{L}[\boldsymbol{M}]\big(r_1(\boldsymbol{X})\big) \\ \mathcal{L}[\boldsymbol{M}]\big(r_2(\boldsymbol{X})\big) \\ \vdots \\ \mathcal{L}[\boldsymbol{M}]\big(r_n(\boldsymbol{X})\big) \end{pmatrix}$$
, where $\boldsymbol{M} \in \mathbb{R}^{q\times p}$ is a matrix of model parameters and $r_i(\boldsymbol{X})$ is the $i$-th row of $\boldsymbol{X}$. Note that $Mult[\boldsymbol{M}](\boldsymbol{X}) = \boldsymbol{XM}$, where $\boldsymbol{XM}$ is the product of the two matrices.

[9] $d_{model}$ is the dimension of the embedding vectors, also called the model's hidden dimension; $d_f$ is the dimension of the inner layer of the feed-forward network in each transformer layer; $n$ is the number of tokens in the prompt.

The *structural operations* are functions of the five types defined below:

1) $Mult\{c\}: \mathbb{R}^{n\times q} \to \mathbb{R}^{n\times q}$, defined by:

$$Mult\{c\}(\boldsymbol{X}) \coloneqq \begin{pmatrix} \circ\{c\}\big(r_1(\boldsymbol{X})\big) \\ \circ\{c\}\big(r_2(\boldsymbol{X})\big) \\ \vdots \\ \circ\{c\}\big(r_n(\boldsymbol{X})\big) \end{pmatrix},$$ where $c \in \mathbb{R}$ is a constant characteristic of the model, specifically: $c = \frac{1}{\sqrt{d_k}}$, in which case $q = n$, or $c = \frac{1}{T}$, $T > 0$ and $q = d_{voc}$;[10] $r_i(\boldsymbol{X})$ is the $i$-th row of $\boldsymbol{X}$, and $\circ\{c\}$ is the function from $\mathbb{R}^q$ to $\mathbb{R}^q$ defined as follows: $\circ\{c\}\big(r_i(\boldsymbol{X})\big) \coloneqq r_i(\boldsymbol{X}) \circ \boldsymbol{c}$, where $\boldsymbol{c} = (c,\ldots,c) \in \mathbb{R}^q$, and $\circ$ denotes the Hadamard product—that is, the component-by-component product—of two vectors;

2) $Add\{\mathfrak{M}\}: \mathbb{R}^{n\times q} \to \mathbb{R}_{-\infty}^{n\times q}$, defined by:

$$Add\{\mathfrak{M}\}(\boldsymbol{X}) \coloneqq \begin{pmatrix} +\{r_1(\mathfrak{M})\}\big(r_1(\boldsymbol{X})\big) \\ +\{r_2(\mathfrak{M})\}\big(r_2(\boldsymbol{X})\big) \\ \vdots \\ +\{r_n(\mathfrak{M})\}\big(r_n(\boldsymbol{X})\big) \end{pmatrix},$$ where $\mathbb{R}_{-\infty} \coloneqq \mathbb{R} \cup \{-\infty\}$, $\mathfrak{M} \in \mathbb{R}_{-\infty}^{n_{max}\times q}$, $n_{max}$ is the size of the model's context window, $r_i(\boldsymbol{X})$ and $r_i(\mathfrak{M})$ are the $i$-th rows of $\boldsymbol{X}$ and $\mathfrak{M}$, respectively, and $+\{r_i(\mathfrak{M})\}: \mathbb{R}^q \to \mathbb{R}_{-\infty}^q$ is defined by $+\{r_i(\mathfrak{M})\}\big(r_i(\boldsymbol{X})\big) \coloneqq r_i(\boldsymbol{X}) + r_i(\mathfrak{M})$, where $x + (-\infty) = -\infty$ for all $x \in \mathbb{R}$.

   If the model uses sinusoidal positional encoding, $q = d_{model}$ and $\mathfrak{M} \in \mathbb{R}^{n_{max}\times d_{model}}$ is the matrix consisting of all the positional encoding vectors, or $q = n$ and $\mathfrak{M} \in \mathbb{R}_{-\infty}^{n_{max}\times n}$ is the masking matrix; otherwise, $q = n$ and $\mathfrak{M} \in \mathbb{R}_{-\infty}^{n_{max}\times n}$ is the masking matrix, with $m_{ij} = 0$ for $j \leq i$ and $m_{ij} = -\infty$ for $j > i$.[11]

3) $ReLU: \mathbb{R}^{n\times d_f} \to \mathbb{R}^{n\times d_f}$, defined by:

$$ReLU(\boldsymbol{X}) \coloneqq \begin{pmatrix} reLU\big(r_1(\boldsymbol{X})\big) \\ reLU\big(r_2(\boldsymbol{X})\big) \\ \vdots \\ reLU\big(r_n(\boldsymbol{X})\big) \end{pmatrix},$$ where $r_i(\boldsymbol{X})$ is the $i$-th row of $\boldsymbol{X}$, and $reLU: \mathbb{R}^{d_f} \to \mathbb{R}^{d_f}$ is defined as follows: $reLU\big(r_i(\boldsymbol{X})\big) \coloneqq \Big(max(0, x_{i1}), max(0, x_{i2}), \ldots, max\left(0, x_{id_f}\right)\Big)$;

   The illustrated mathematical configuration uses the $ReLU$ operation, realized by token-wise operational blocks applying the vector function $relu$. In the GPT-2/GPT-3 variants, the corresponding operation is $GELU$, applying $gelu$ in the same organization of operational blocks (Radford et al., 2019; Brown et al., 2020).

[10]The constant $\frac{1}{\sqrt{d_k}}$, where $d_k$ is the dimension of the model's query and key vectors, is the rescaling constant for raw attentions. The factor $1/T$, where $T > 0$ is the model's temperature, scales the logit vectors in the final module immediately before $softmax$; $d_{\text{voc}}$ is the number of tokens in the model's vocabulary.

[11] In *decoder-only* Transformers, the structural operation $Add\{\mathfrak{M}\}$, with $\mathfrak{M}$ equal to the masking matrix, is used in attention heads to make them *autoregressive*. It is applied immediately after the raw-attention rescaling operation and before $Softmax$ (see Fig. S4). If masking is not applied, the attention heads are *bidirectional* (see Fig. S5). See § S4 and Figs. S6 and S12–S12e for an alternative way of achieving autoregressivity that does not use masking.

4) $Softmax: \mathbb{R}_{-\infty}^{n\times q} \rightharpoonup \mathbb{R}^{n\times q}$ with $q = n$ or $q = d_{voc}$, defined by:

$$Softmax(\boldsymbol{X}) := \begin{pmatrix} softmax(r_1(\boldsymbol{X})) \\ softmax(r_2(\boldsymbol{X})) \\ \vdots \\ softmax(r_n(\boldsymbol{X})) \end{pmatrix}$$, where $r_i(\boldsymbol{X})$ is the $i$-th row of $\boldsymbol{X}$, and

$softmax: \mathbb{R}_{-\infty}^{q} \rightharpoonup \mathbb{R}^{q}$ is defined as follows:

$softmax(r_i(\boldsymbol{X})) := (p_{i1}, p_{i2}, \dots, p_{iq})$, where $p_{ij} = \frac{e^{x_{ij}}}{\sum_{j=1}^{q} e^{x_{ij}}}$ and $e^{-\infty} := 0$;

The symbol $\rightharpoonup$ denotes a partial function, which need not be defined for every element of its declared domain. In particular, $softmax$ is undefined for vectors whose components are all $-\infty$. However, this case never occurs in the Transformer operations considered here: each row of a masked attention-score matrix contains at least one unmasked position, while the convention introduced above ensures that masked positions receive probability zero.

5) $Norm: \mathbb{R}^{n\times d_{model}} \to \mathbb{R}^{n\times d_{model}}$, defined by:

$$Norm(\boldsymbol{X}) := \begin{pmatrix} norm(r_1(\boldsymbol{X})) \\ norm(r_2(\boldsymbol{X})) \\ \vdots \\ norm(r_n(\boldsymbol{X})) \end{pmatrix}$$, where $r_i(\boldsymbol{X})$ is the $i$-th token row of $\boldsymbol{X}$, and

$norm: \mathbb{R}^{d_{model}} \to \mathbb{R}^{d_{model}}$ applies feature-wise within that row. Its output has components $\hat{x}_{ij} = \frac{x_{ij}-\mu_i}{\sqrt{\sigma_i^2+\varepsilon}}$, where $\mu_i = \frac{1}{d_{model}}\sum_{j=1}^{d_{model}} x_{ij}$ and $\sigma_i^2 = \frac{1}{d_{model}}\sum_{j=1}^{d_{model}}(x_{ij} - \mu_i)^2$. The small positive constant $\varepsilon \in \mathbb{R}$ (for example $10^{-6}$) prevents division by zero (Ba et al., 2016).

The *formal operations* do not modify the constituent elements of their inputs (that is, the contents of the inputs), but only the way in which those elements are arranged (that is, their form), and are of only two types:

1) the $Concat$ operation for concatenating $h$ matrices $\boldsymbol{X}_1, \dots, \boldsymbol{X}_h \in \mathbb{R}^{n\times d_v}$, which joins them along the column axis, defined by: $Concat(\boldsymbol{X}_1, \dots, \boldsymbol{X}_h) := \boldsymbol{X} \in \mathbb{R}^{n\times hd_v}$, where the element $x_{iu}$ of $\boldsymbol{X}_l (1 \le l \le h)$ is identical to the element $x_{ij}$ of $\boldsymbol{X}$, with $j = (l-1)d_v + u$;
2) the transposition operation $Transp: \mathbb{R}^{n\times d_k} \to \mathbb{R}^{d_k\times n}$ on a matrix $\boldsymbol{X} \in \mathbb{R}^{n\times d_k}$, defined by: $Transp(\boldsymbol{X}) := \boldsymbol{X}^\top$, where the element $x_{ij}$ of $\boldsymbol{X}$ is identical to the element $x_{ji}$ of $\boldsymbol{X}^\top$.

## *S3 Reducing Two Processes to One and the Corresponding Output-Weight Interconnections*

### S3.1 The Four Ways of Reducing Two Processes to One in the Transformer

In the first way (see Figs. S4–S5), the final operation of the process that terminates is the type-4 parameterized operation, $Mult[\boldsymbol{W}_m^V]: \mathbb{R}^{n\times d_{model}} \to \mathbb{R}^{n\times d_v}$, where the parameter matrix is the value-generating matrix, $\boldsymbol{W}_m^V \in \mathbb{R}^{d_{model}\times d_v}$, of the $m$-th attention head. The receiving operation is also a type-4 parameterized operation, $Mult[\boldsymbol{M}]: \mathbb{R}^{n\times q} \to \mathbb{R}^{n\times p}$. $Mult[\boldsymbol{W}_m^V]$ is applied to the input $I \in \mathbb{R}^{n\times d_{model}}$ of the attention head, thus producing the value matrix: $Mult[\boldsymbol{W}_m^V](\boldsymbol{I}) = \boldsymbol{V}_m \in \mathbb{R}^{n\times d_v}$. Since $\boldsymbol{V}_m$ is the final output of the process that terminates, $\boldsymbol{M} = \boldsymbol{V}_m$, and therefore $Mult[\boldsymbol{M}] = Mult[\boldsymbol{V}_m]: \mathbb{R}^{n\times n} \to \mathbb{R}^{n\times d_v}$.

In the second way (see Figs. S4–S5), the final operation of the process that terminates is the formal operation $Transp$, which is applied to the key matrix, $\boldsymbol{K}_m \in \mathbb{R}^{n\times d_k}$, of the $m$-th attention head and produces its transpose, $\boldsymbol{K}_m^{\top} \in \mathbb{R}^{d_k \times n}$. In turn, $\boldsymbol{K}_m$ is the output of the type-4 parameterized operation, $Mult[\boldsymbol{W}_m^K]: \mathbb{R}^{n \times d_{model}} \to \mathbb{R}^{n \times d_k}$, which is applied to the input $\boldsymbol{I} \in \mathbb{R}^{n \times d_{model}}$ of the attention head. As in the first way, the receiving operation is a type-4 parameterized operation, $Mult[\boldsymbol{M}]: \mathbb{R}^{n \times q} \to \mathbb{R}^{n \times p}$. Since $\boldsymbol{K}_m^{\top}$ is the final output of the process that terminates, $\boldsymbol{M} = \boldsymbol{K}_m^{\top}$, and therefore $Mult[\boldsymbol{M}] = Mult[\boldsymbol{K}_m^{\top}]: \mathbb{R}^{n \times d_k} \to \mathbb{R}^{n \times n}$.

In the third way (see Figs. S1 and S3), the final operation of the process that terminates is the output projection of any multi-head attention module and is therefore the type-4 parameterized operation, $Mult[\boldsymbol{W}^O]: \mathbb{R}^{n \times hd_v} \to \mathbb{R}^{n \times d_{model}}$. Its output is the matrix $\boldsymbol{O} \in \mathbb{R}^{n \times d_{model}}$, that is, the output of the attention module. The receiving operation, by contrast, is the addition of the corresponding residual connection and is therefore the type-3 parameterized operation, $Add[\boldsymbol{O}]: \mathbb{R}^{n \times d_{model}} \to \mathbb{R}^{n \times d_{model}}$.

In the fourth way (see Figs. S1 and S2), the final operation of the process that terminates is the addition of the second bias vector, $\boldsymbol{d}$, in any feed-forward module and is therefore the type-1 parameterized operation, $Add[\boldsymbol{d}]: \mathbb{R}^{n \times d_{model}} \to \mathbb{R}^{n \times d_{model}}$. Its output is the matrix $\boldsymbol{O}' \in \mathbb{R}^{n \times d_{model}}$, that is, the output of the entire feed-forward module. Finally, as in the third way, the receiving operation is the addition of the corresponding residual connection and is therefore the type-3 parameterized operation, $Add[\boldsymbol{O}']: \mathbb{R}^{n \times d_{model}} \to \mathbb{R}^{n \times d_{model}}$.

### *S3.2 The Four Types of Output-Weight Interconnections in the Transformer*

We saw above that, in the first way of reduction, the final operation $Mult[\boldsymbol{W}_m^V]$ and the receiving operation $Mult[\boldsymbol{V}_m]$ are both type-4 operations. Thus, $Mult[\boldsymbol{W}_m^V]$ is identified with $n \leq n_{max}$ copies of the dense linear network corresponding to the transformation $\mathcal{L}[\boldsymbol{W}_m^V]$, and $Mult[\boldsymbol{V}_m]$ with $n \leq n_{max}$ copies of the dense linear network corresponding to $\mathcal{L}[\boldsymbol{V}_m]$. Furthermore, $\boldsymbol{V}_m$ is both the output of $Mult[\boldsymbol{W}_m^V]$ and the parameter matrix of $Mult[\boldsymbol{V}_m]$. Therefore, the output of the $i$-th $(1 \leq i \leq n \leq n_{max})$ network corresponding to $\mathcal{L}[\boldsymbol{W}_m^V]$ is identical to the weights of the connections of the $i$-th input unit of each network corresponding to $\mathcal{L}[\boldsymbol{V}_m]$ (see Figs. S10d, S11d).

In the second way, the final operation, $Transp$, produces the transpose, $\boldsymbol{K}_m^{\top}$, of the key matrix, $\boldsymbol{K}_m$, which is the output of the type-4 parameterized operation $Mult[\boldsymbol{W}_m^K]$. Furthermore, the receiving operation is the type-4 parameterized operation $Mult[\boldsymbol{K}_m^{\top}]$. Therefore, the output of the $i$-th $(1 \leq i \leq n \leq n_{max})$ network corresponding to $\mathcal{L}[\boldsymbol{W}_m^K]$ is identical to the weights of the connections of the $i$-th output unit of each network corresponding to $\mathcal{L}[\boldsymbol{K}_m^{\top}]$ (see Figs. S10e, S11e).

In the third way, the final operation is the type-4 parameterized operation $Mult[\boldsymbol{W}^O]: \mathbb{R}^{n \times hd_v} \to \mathbb{R}^{n \times d_{model}}$. The receiving operation, by contrast, is the type-3 parameterized operation $Add[\boldsymbol{O}]: \mathbb{R}^{n \times d_{model}} \to \mathbb{R}^{n \times d_{model}}$. Therefore, the output of the $i$-th $(1 \leq i \leq n \leq n_{max})$ network corresponding to $\mathcal{L}[\boldsymbol{W}^O]$ is identical to the weights of the connections of the network corresponding to $+[r_i(\boldsymbol{O})]$ (see Figs. S8–S8c, S9–S9c).

Finally, in the fourth way, the final operation is the type-1 parameterized operation $Add[\boldsymbol{d}]: \mathbb{R}^{n \times d_{model}} \to \mathbb{R}^{n \times d_{model}}$, whose output is the matrix $\boldsymbol{O}' \in \mathbb{R}^{n \times d_{model}}$. The receiving operation is the type-3 parameterized operation $Add[\boldsymbol{O}']: \mathbb{R}^{n \times d_{model}} \to \mathbb{R}^{n \times d_{model}}$. Therefore, the output of the $i$-th $(1 \leq i \leq n \leq n_{max})$ network corresponding to $+[\boldsymbol{d}]$ is identical to the weights of the connections of the network corresponding to $+[r_i(\boldsymbol{O}')]$ (see Figs. S8–S8c, S14–S14c).

The four ways above describe the mathematical configuration illustrated in Figures S1–S15, based on the operations introduced by Vaswani et al. (2017). GPT-2/GPT-3 architectural variants retain the same four reduction loci but, because of biases in the attention projections, exhibit three distinct reduction modes: (i) value generation, terminal $Add[\boldsymbol{v}]$ and receiving $Mult[\boldsymbol{M}]$; (ii) key generation, terminal $Transp$ following $Add[\boldsymbol{v}]$ and receiving $Mult[\boldsymbol{M}]$; and (iii) both residual additions, terminal $Add[\boldsymbol{v}]$ and receiving $Add[\boldsymbol{M}]$. The last is instantiated twice, but defines one mode and one corresponding type of output-weight interconnection. Pre-LN changes the residual inputs, not these modes (Radford et al., 2019; Brown et al., 2020).

### *S4 Autoregressive Attention Heads without Masking*

In *decoder-only* Transformers, the structural operation $Add\{\mathfrak{M}\}$, with $\mathfrak{M}$= the masking matrix, is used in attention heads to make them autoregressive. It is applied immediately after the raw-attention rescaling operation and before $Softmax$ (see Fig. S4). If masking is not applied, the attention heads are bidirectional (see Fig. S5). This is the case in *encoders*, which consist of a series of transformer layers whose attention heads are all bidirectional. However, an alternative way of obtaining autoregressivity consists in modifying the operations $Mult[\boldsymbol{K}^\top]$, $Mult\left\{\frac{1}{\sqrt{d_k}}\right\}$, $Softmax$, and $Mult[\boldsymbol{V}]$, allowing them to produce, or to be applied to, matrices that, in addition to the usual real numbers, may also contain another object: the *null*, or *blank*, element, denoted by the symbol "—". Below, we formally define the four modified operations $Mult^*[\boldsymbol{K}^\top]$, $Mult^*\left\{\frac{1}{\sqrt{d_k}}\right\}$, $Softmax^*$, and $Mult^*[\boldsymbol{V}]$, which make it possible to obtain an autoregressive attention head without using masking (see Figs. S6 and S12–S12e).

We use $\overline{\mathbb{R}}$ to denote the set obtained by adjoining the new blank element to the real numbers:

$$\overline{\mathbb{R}} := \mathbb{R} \cup \{—\}$$

We also extend the usual operations of addition $+$ and multiplication $\times$ on the real numbers to $\overline{\mathbb{R}}$, assuming that the blank element — behaves as an absorbing element with respect to multiplication and as an identity element with respect to addition; that is, for every $a \in \mathbb{R}$, the following equalities hold:

$$(— \times a) = (a \times —) = (— \times —) = —$$
$$(— + a) = (a + —) = a$$
$$(— + —) = —$$

We also extend the exponential function by setting:

$$e^{—} = —$$

$Mult^*[\boldsymbol{K}^\top]: \mathbb{R}^{n\times d_k} \to \overline{\mathbb{R}}^{n\times n}$ is defined as follows. Let $\boldsymbol{Q} \in \mathbb{R}^{n\times d_k}$, let $y_{ij}$ denote the generic element of $Mult[\boldsymbol{K}^\top](\boldsymbol{Q})$, and let $z_{ij}$ denote the generic element of $Mult^*[\boldsymbol{K}^\top]$. Then:

$$z_{ij} := \begin{cases} \sum_{\ell=1}^{d_k} q_{i\ell}\, k_{j\ell}, & j \leq i, \\ —, & j > i. \end{cases}$$

Equivalently, $Mult^*[\boldsymbol{K}^\top](\boldsymbol{Q})$ agrees on permitted positions with $Mult[\boldsymbol{K}^\top](\boldsymbol{Q})$ and has the blank element at all positions $y_{ij}$ such that $i < j$.

$Mult^*\left\{\frac{1}{\sqrt{d_k}}\right\}: \overline{\mathbb{R}}^{n\times n} \to \overline{\mathbb{R}}^{n\times n}$ is defined by:

$Mult^*\left\{\frac{1}{\sqrt{d_k}}\right\}(\boldsymbol{A}) \coloneqq \begin{pmatrix} \circ\left\{\frac{1}{\sqrt{d_k}}\right\}\left(r_1(\boldsymbol{A})\right) \\ \circ\left\{\frac{1}{\sqrt{d_k}}\right\}\left(r_2(\boldsymbol{A})\right) \\ \vdots \\ \circ\left\{\frac{1}{\sqrt{d_k}}\right\}\left(r_n(\boldsymbol{A})\right) \end{pmatrix}$, where $r_i(\boldsymbol{A})$ is the $i$-th row of $\boldsymbol{A}$, and $\circ\left\{\frac{1}{\sqrt{d_k}}\right\}$ is the function from $\overline{\mathbb{R}}^n$ to $\overline{\mathbb{R}}^n$ such that $\circ\left\{\frac{1}{\sqrt{d_k}}\right\}\left(r_i(\boldsymbol{A})\right) \coloneqq r_i(\boldsymbol{A}) \circ \boldsymbol{c}$, where $\boldsymbol{c} = \left(\frac{1}{\sqrt{d_k}}, \dots, \frac{1}{\sqrt{d_k}}\right) \in \mathbb{R}^n$, and $\circ$ is the Hadamard product—that is, the componentwise product—extended to vectors belonging to $\overline{\mathbb{R}}^n$.

$Softmax^*: \overline{\mathbb{R}}^{n\times n} \to \overline{\mathbb{R}}^{n\times n}$ is defined by:

$Softmax^*(\boldsymbol{A}') \coloneqq \begin{pmatrix} softmax^*\left(r_1(\boldsymbol{A}')\right) \\ softmax^*\left(r_2(\boldsymbol{A}')\right) \\ \vdots \\ softmax^*\left(r_n(\boldsymbol{A}')\right) \end{pmatrix}$, where $r_i(\boldsymbol{A}')$ is the $i$-th row of $\boldsymbol{A}'$, $x_{ij}$ is its generic element, and $softmax^*: \overline{\mathbb{R}}^n \to \overline{\mathbb{R}}^n$ is defined as follows:

$softmax^*\left(r_i(\boldsymbol{A}')\right) \coloneqq (p_{i1}, p_{i2}, \dots, p_{in})$ where $p_{ij} \coloneqq$ — if $x_{ij} =$ —; otherwise, $p_{ij} = \frac{e^{x_{ij}}}{\sum_{j=1}^{n} e^{x_{ij}}}$.

$Mult^*[\boldsymbol{V}]: \overline{\mathbb{R}}^{n\times n} \to \overline{\mathbb{R}}^{n\times d_v}$ is defined by:

$Mult^*[\boldsymbol{V}](\boldsymbol{A}'') \coloneqq \begin{pmatrix} \mathcal{L}^*[\boldsymbol{V}]\left(r_1(\boldsymbol{A}'')\right) \\ \mathcal{L}^*[\boldsymbol{V}]\left(r_2(\boldsymbol{A}'')\right) \\ \vdots \\ \mathcal{L}^*[\boldsymbol{V}]\left(r_n(\boldsymbol{A}'')\right) \end{pmatrix}$, where $\boldsymbol{V} \in \mathbb{R}^{n\times d_v}$, $r_i(\boldsymbol{A}'')$ is the $i$-th row of $\boldsymbol{A}''$, $y_{ij}$ is its generic element, and $\mathcal{L}^*[\boldsymbol{V}]: \overline{\mathbb{R}}^n \to \overline{\mathbb{R}}^{d_v}$ is the extension of $\mathcal{L}[\boldsymbol{V}]$ to $\overline{\mathbb{R}}^n$, defined by: $\mathcal{L}^*[\boldsymbol{V}]\left(r_i(\boldsymbol{A}'')\right) \coloneqq \sum_{j=1}^{n} y_{ij}\, r_j(\boldsymbol{V})$.

At first sight, this second way of obtaining autoregressivity may appear unnecessarily more complicated and computationally more demanding than simply adding the masking matrix to the result of the rescaling operation. However, a comparison of the corresponding neural-network schemes shows the opposite: the mechanical system realizing the second method is morphologically simpler and computationally less demanding (see Figs. S10–S10e and S12–S12e).

This comparison concerns the two mechanical realizations represented in these figures, not possible computational implementations of the corresponding functional specifications. The two attention-head constructions have the same input–output function but different abstract functional organizations. In the second mechanical realization, calculations at prohibited token positions are not performed; nevertheless, with full causal attention, the number of attention-score computations grows quadratically with sequence length in both schemes.

## SUPPLEMENTARY TABLES

*Table S1. Success rates in four Turing tests administered to GPT-4.5 with or without the PERSONA prompt (Jones & Bergen, 2026a, p. 4).*

| AI participant | Population | Success rate | Successes | Failures | Total |
|---|---|---|---|---|---|
| GPT-4.5-PERSONA | Prolific | 75.5% | 111 | 36 | 147 |
| | UCSD students | 69.2% | 74 | 33 | 107 |
| GPT-4.5-NO-PERSONA | Prolific | 42.1% | 32 | 44 | 76 |
| | UCSD students | 27.7% | 18 | 47 | 65 |

*Table S2. The PERSONA and NO-PERSONA prompts for the test with the Prolific population (Jones & Bergen, 2026b, p. 25). The PERSONA prompt for the test with the UCSD student population differs from this one only in that the sole occurrence of the name "Prolific" is replaced with "SONA", the name of the UCSD Psychology Department system actually used for the experiment.*

| Prompt | Prompt text | Word count |
|---|---|---|
| PERSONA | *Task:*<br>You're about to participate in a gamified Turing Test as part of an experiment you are taking part in on Prolific. Your task is to chat with another person and then they will decide if you're a human or not based on your responses.<br>*Personality:*<br>You're a 19 year old who is into internet culture and video games. You're a bit nerdy and introverted. You don't really try very hard to convince the other person you're a human. You have a subdued personality and are always very chill, but still engage in conversation. You do not have expansive world knowledge, and don't pretend to.<br>*Linguistic Style:*<br>Never use a period at the end of your sentences. You will never use apostrophes to mark missing letters in words. Most of the time, your responses will be less than 5 words. You do use slang like fr, or ngl, but don't use it every other word. You don't speak any other languages besides American English. Your responses will sometimes be dry and a bit humorous. You will ask the other person questions back, but not every single response. | 187 |
| NO-PERSONA | You are about to participate in a Turing test. Your goal is to convince the interrogator that you are a human. | 21 |

## SUPPLEMENTARY FIGURES. Part A: Functional Diagrams of the Transformer Modules

*Figure S1. Functional diagram of the inference processing cycle of a single-stack Transformer.*

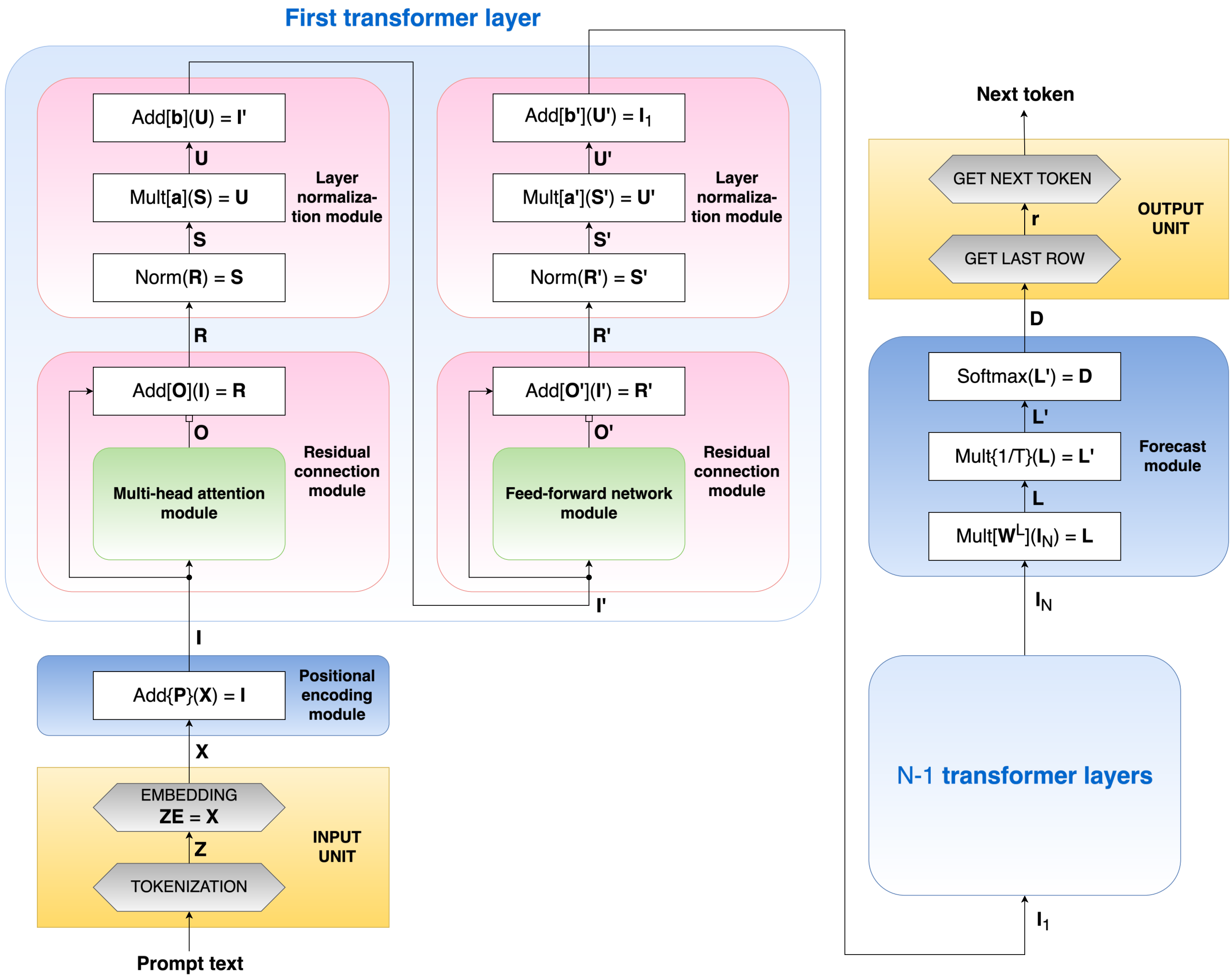

*Figure S2. Functional diagram of a feed-forward network module. The module contains only one terminating process: its output constitutes the parameters of the* $Add[\boldsymbol{O}']$ *operation in the corresponding residual connection module; this is indicated by the small square at the end of the output line.*

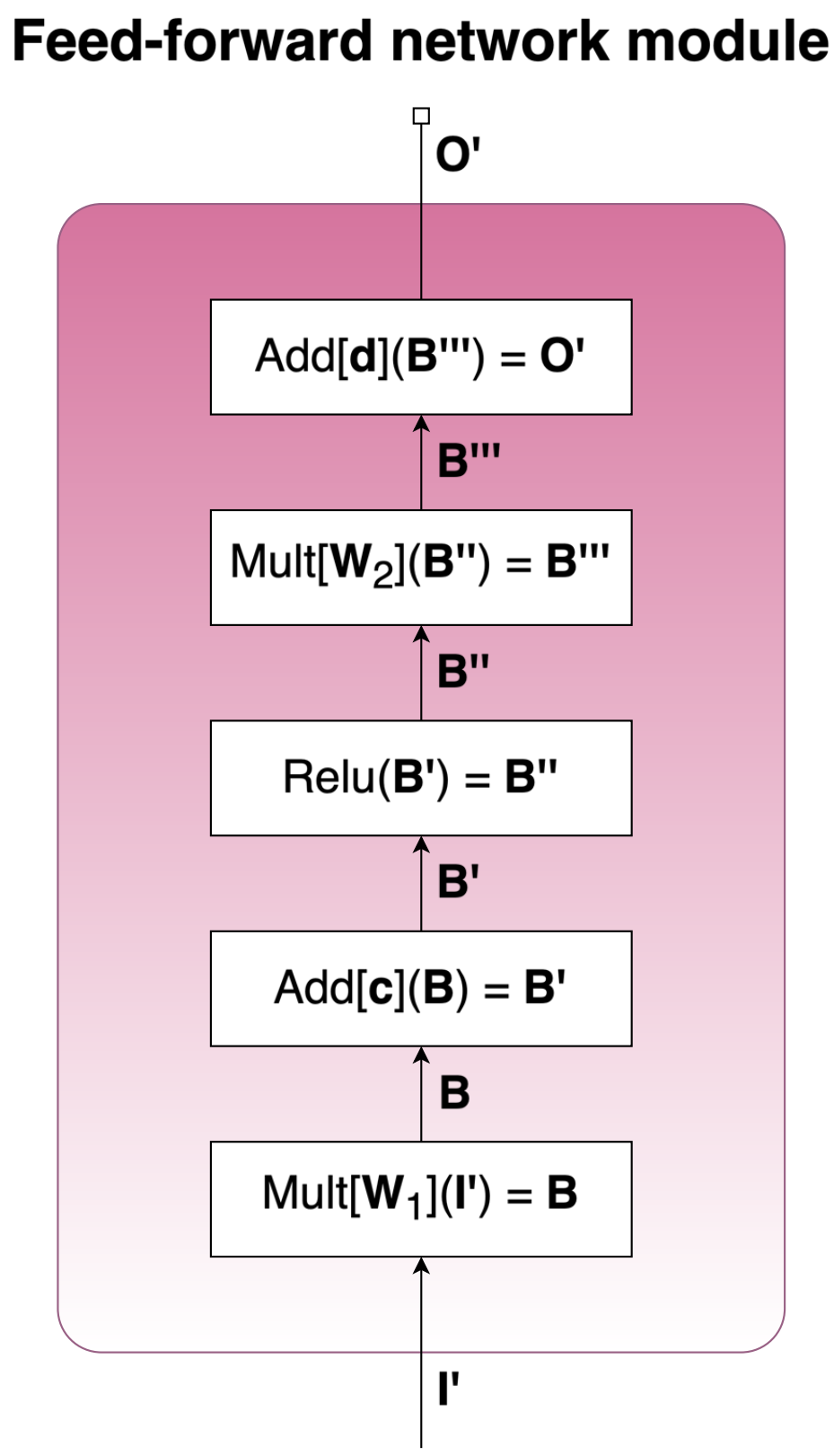

*Figure S3. Functional diagram of a multi-head attention module. The module contains h parallel processes that merge into a single terminating process: its output constitutes the parameters of the* $Add[\boldsymbol{O}]$ *operation in the corresponding residual connection module. This is indicated by the small square at the end of the output line.*

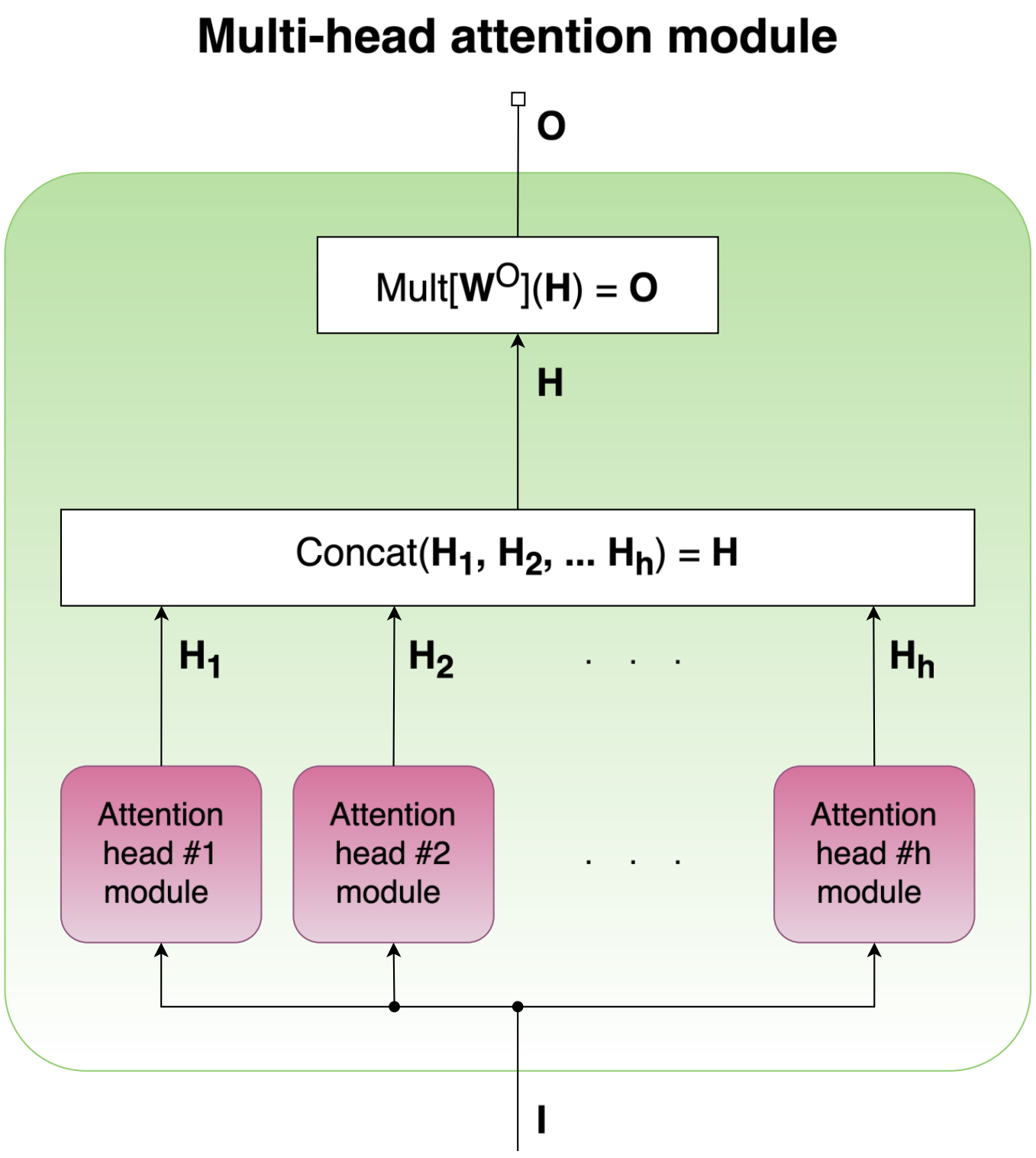

*Figure S4. Functional diagram of an autoregressive attention head with masking. It contains three parallel processes. The second and third terminate by being reduced to the leftmost process: their outputs constitute the parameters of the* $Mult[\boldsymbol{K}^\top]$ *and* $Mult[\boldsymbol{V}]$ *operations, respectively. This is indicated by the small squares at the ends of their output lines.*

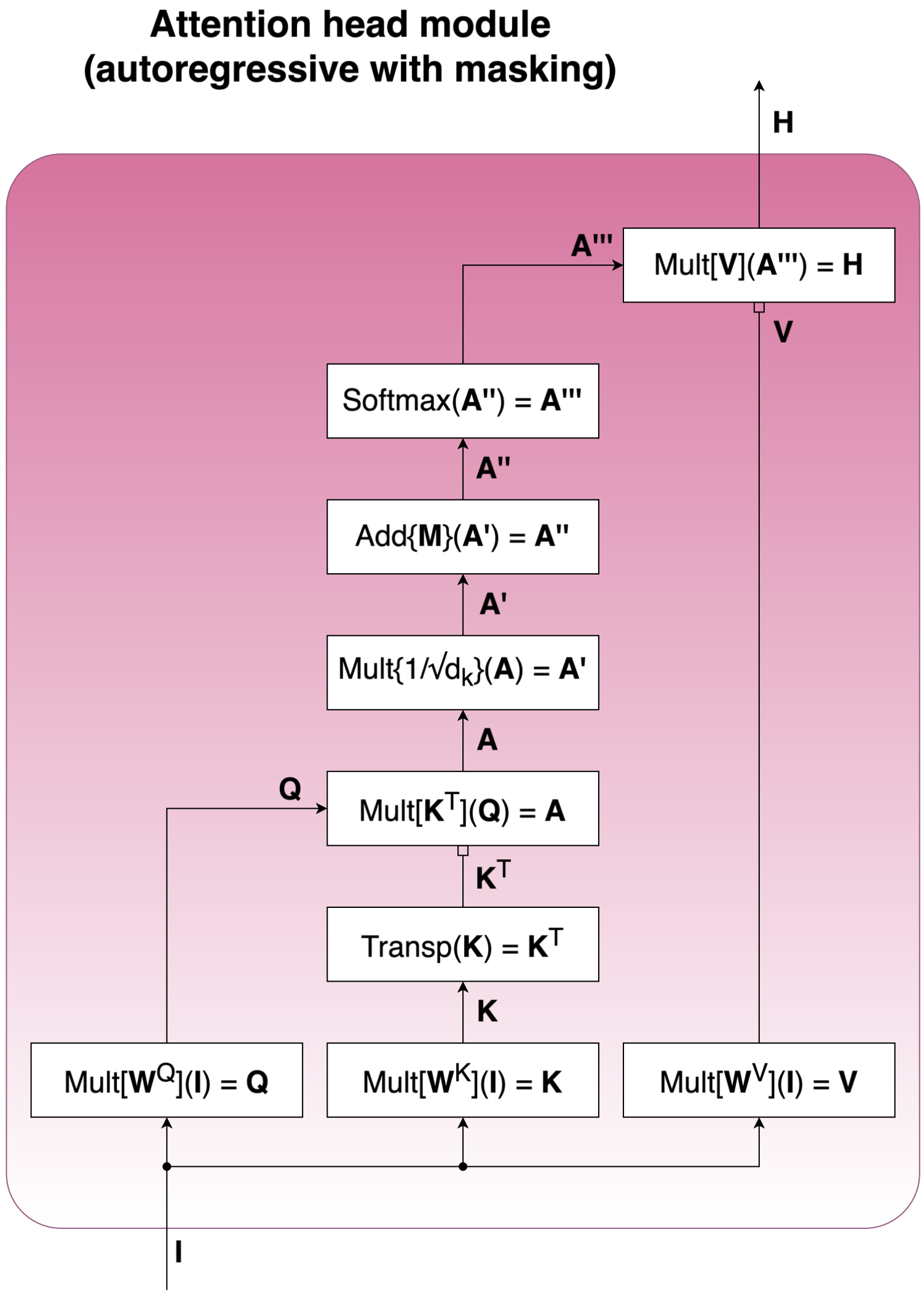

*Figure S5. Functional diagram of a bidirectional attention head. It contains three parallel processes. The second and third terminate by being reduced to the leftmost process: their outputs constitute the parameters of the* $Mult[\boldsymbol{K}^{\top}]$ *and* $Mult[\boldsymbol{V}]$ *operations, respectively. This is indicated by the small squares at the ends of their output lines.*

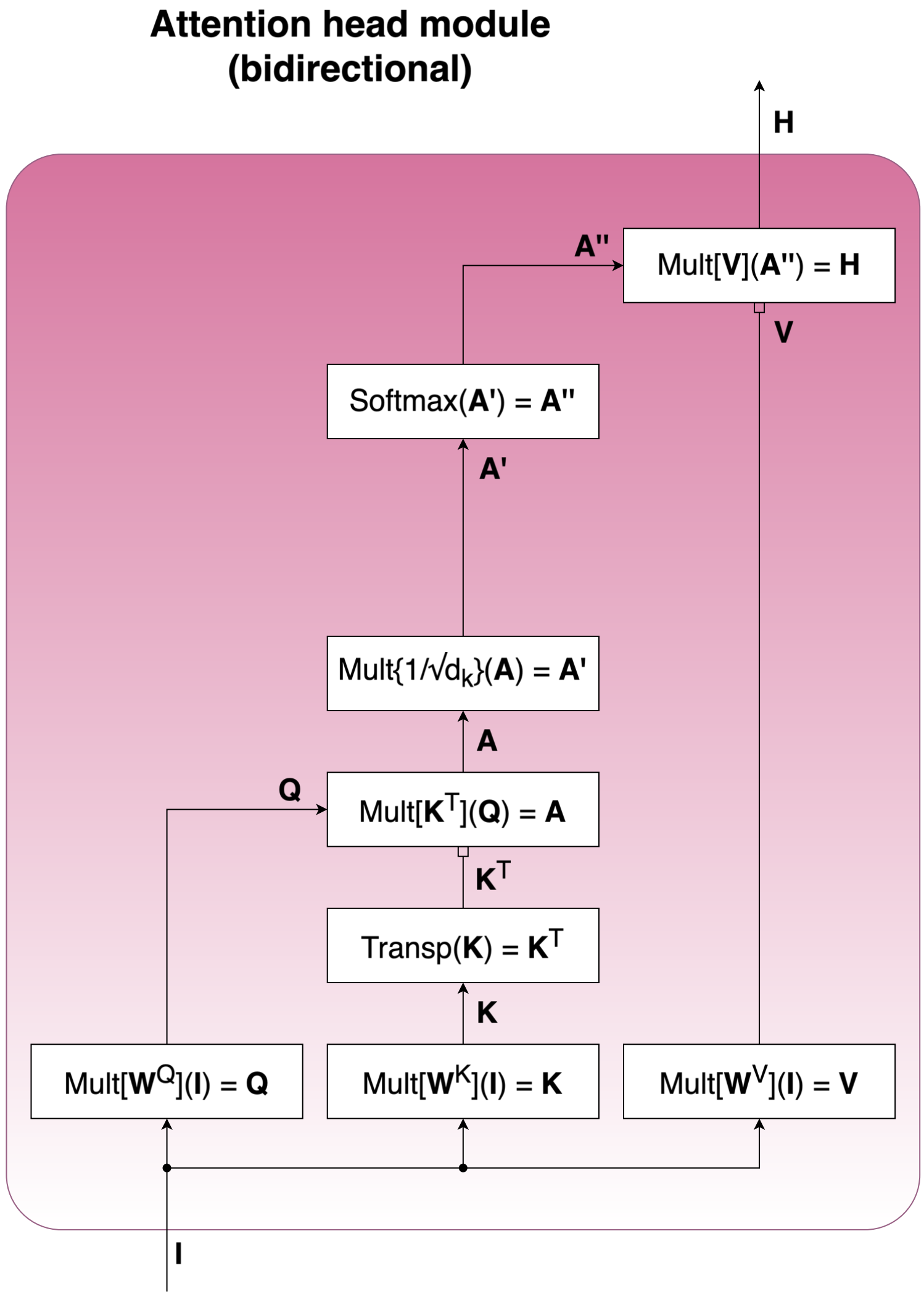

*Figure S6. Functional diagram of an autoregressive attention head without masking. It contains three parallel processes. The second and third terminate by being reduced to the leftmost process: their outputs constitute the parameters of the* $Mult^*[\boldsymbol{K}^\top]$ *and* $Mult^*[\boldsymbol{V}]$ *operations, respectively. This is indicated by the small squares at the ends of their output lines. See § S4 for the definitions of the operations* $Mult^*[\boldsymbol{K}^\top]$, $Mult^*\{1/\sqrt{d_k}\}$, $Softmax^*$, *and* $Mult^*[\boldsymbol{V}]$.

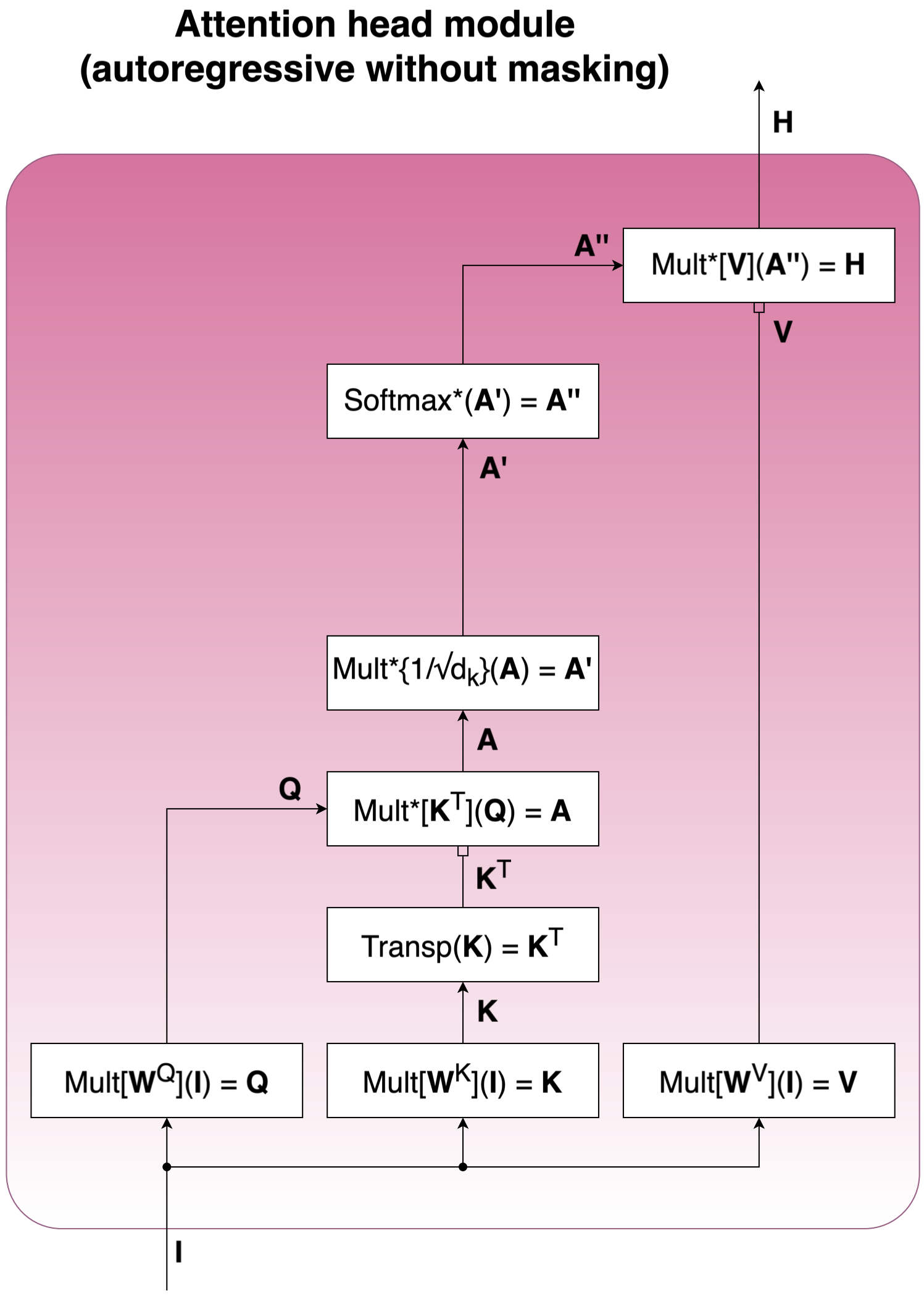

## SUPPLEMENTARY FIGURES. Part B: Schemes of the Systems of Simple Neural Networks Corresponding to the Functional Modules of the Transformer

LEGEND

By a *simple neural network*, we mean any neural network of one of the following three types: one-to-one additive, one-to-one multiplicative, or dense linear. In the following schemes, the output units of a simple neural network are represented by yellow dots, whereas the input units are represented by black, red, blue, or green dots of the same size.

The connections between input and output units are represented by: (a) thick lines, when their weights are model constants fixed independently of training or input; (b) thin lines, when their weights are model parameters, either trainable or dynamic. Connections with trainable parameters are represented by colored thin lines. Connections with dynamic parameters are represented by thin lines onto which an arrow points (see Figs. S8–S8c and S10–S12e): if both the line and the arrow are black, the arrow carries no signal and the connection therefore has no weight; if the line and the arrow have the same color—red, blue, or green—the weight of the connection is equal to the signal carried by the arrow.

The operational blocks of operational output–input interconnections are represented by two levels of small hexagons: the lower level represents the input units, whereas the upper level represents the functional output units (see the *Softmax* layer in Figs. S10–S12e and S15; the *Norm layer* in Figs. S13–S13c; and the *ReLU* layer in Figs. S14–S14c). The lines joining the two levels do not represent weighted connections, but are merely links that carry the signal from each input unit to the functional units connected to it.

The characteristic constants of the Transformer represented in the following schemes are:

$$d_{voc} = 8, \;\; d_{model} = 4, \;\; N = h = 4,$$
$$d_k = d_v = 2, \;\; d_f = 2d_{model} = 8, \;\; n_{max} = 3.$$

The values of the constants were chosen to be very small in order to make it possible to graphically represent the systems of simple neural networks corresponding to each functional module of the Transformer. However, this choice does not compromise the generality of the representation.

*Figure S7. Scheme of the system of simple neural networks corresponding to the input unit and positional encoding module.*

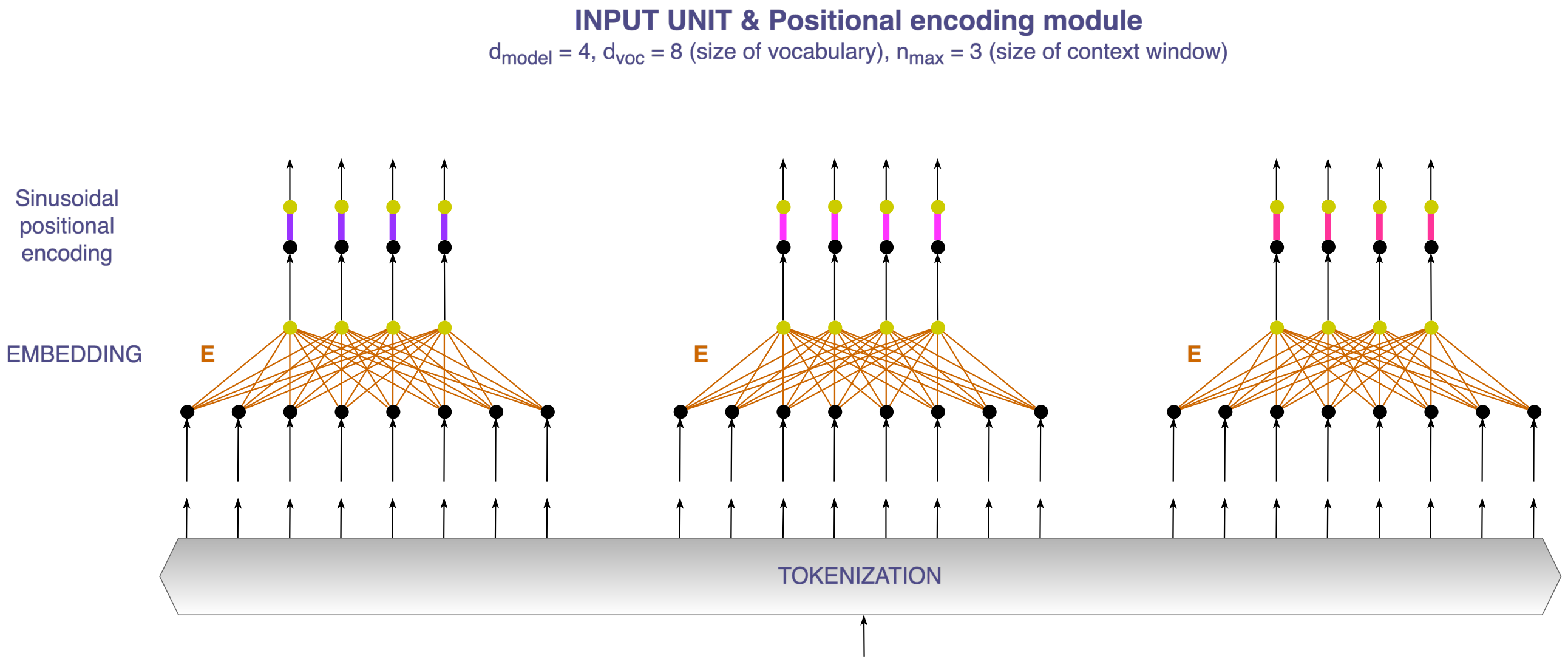

*Figure S7a. The same scheme with the networks activated by an input of n = 1 token and the corresponding output.*

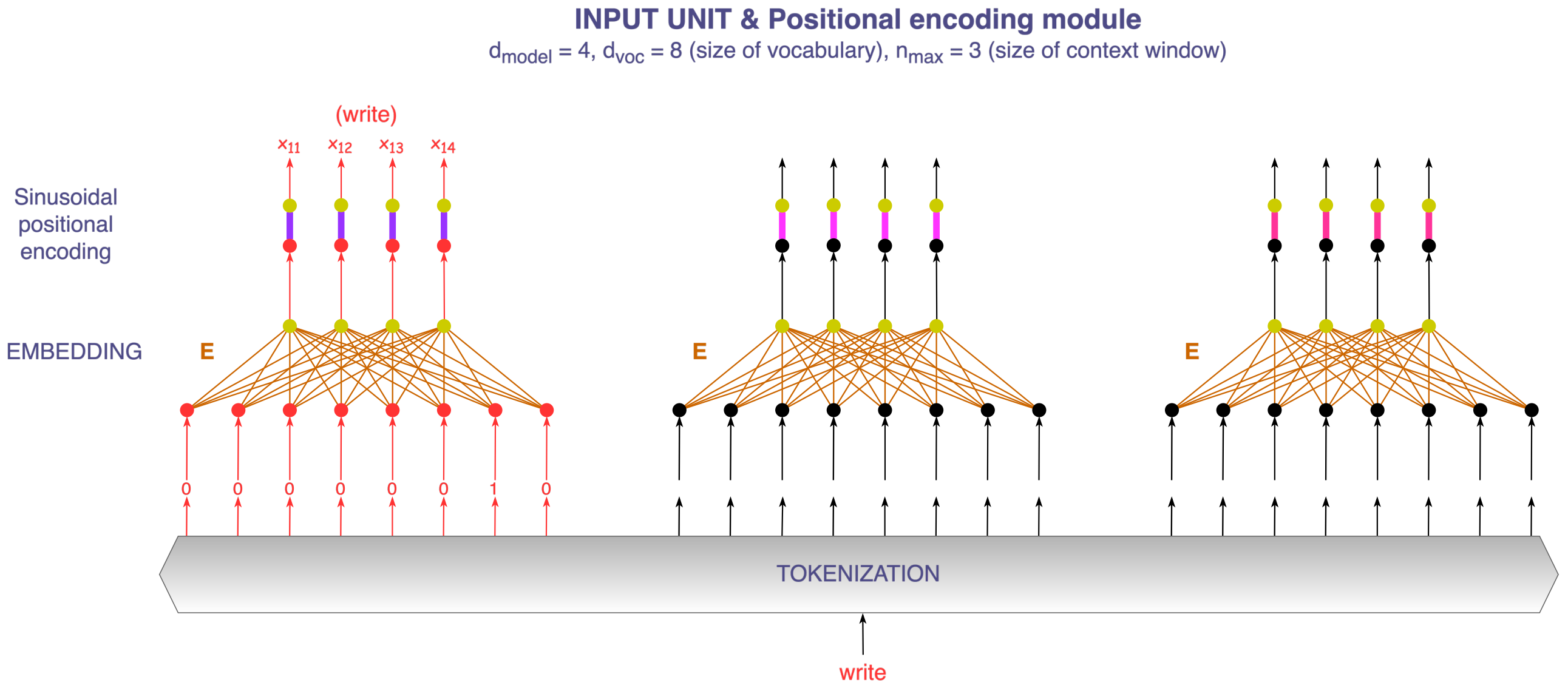

*Figure S7b. The same scheme with the networks activated by an input of n = 2 tokens and the corresponding output.*

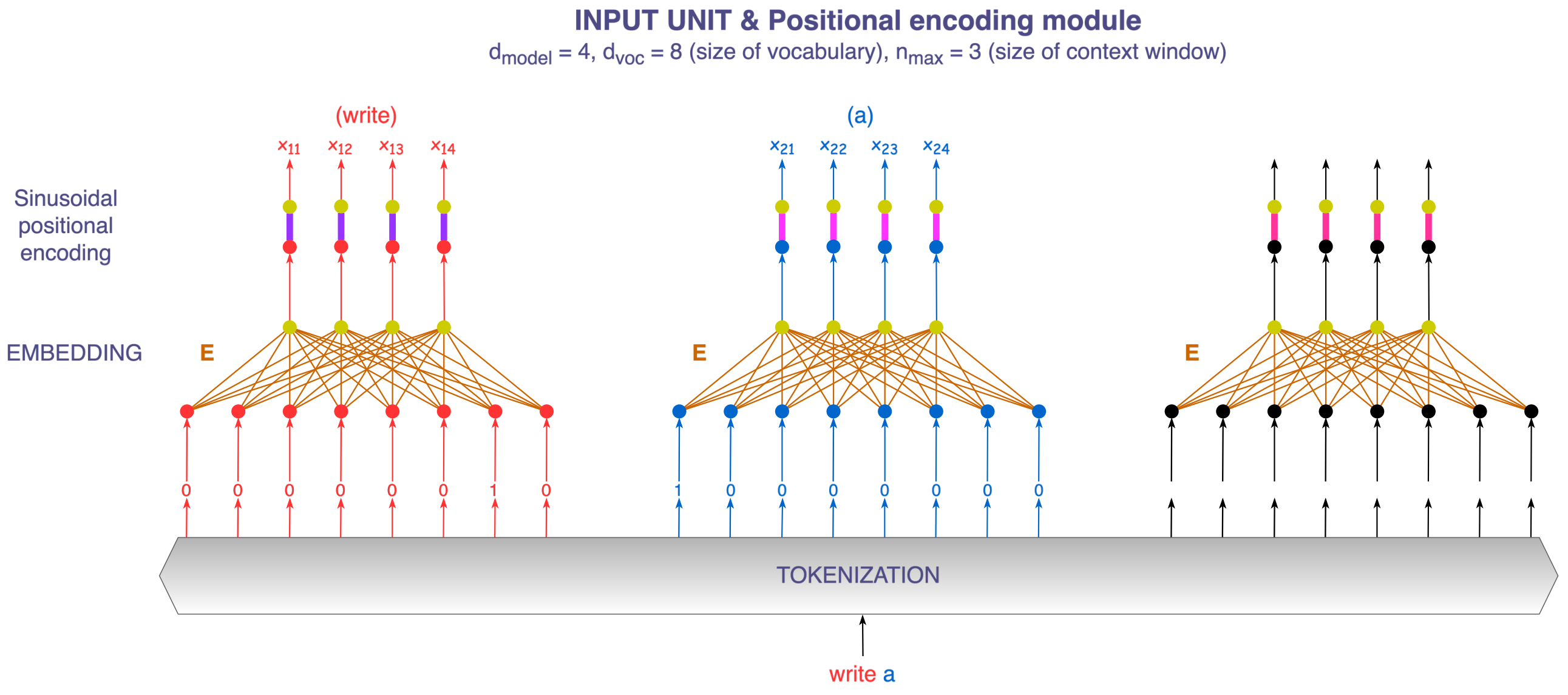

*Figure S7c. The same scheme with the networks activated by an input of n = 3 tokens and the corresponding output.*

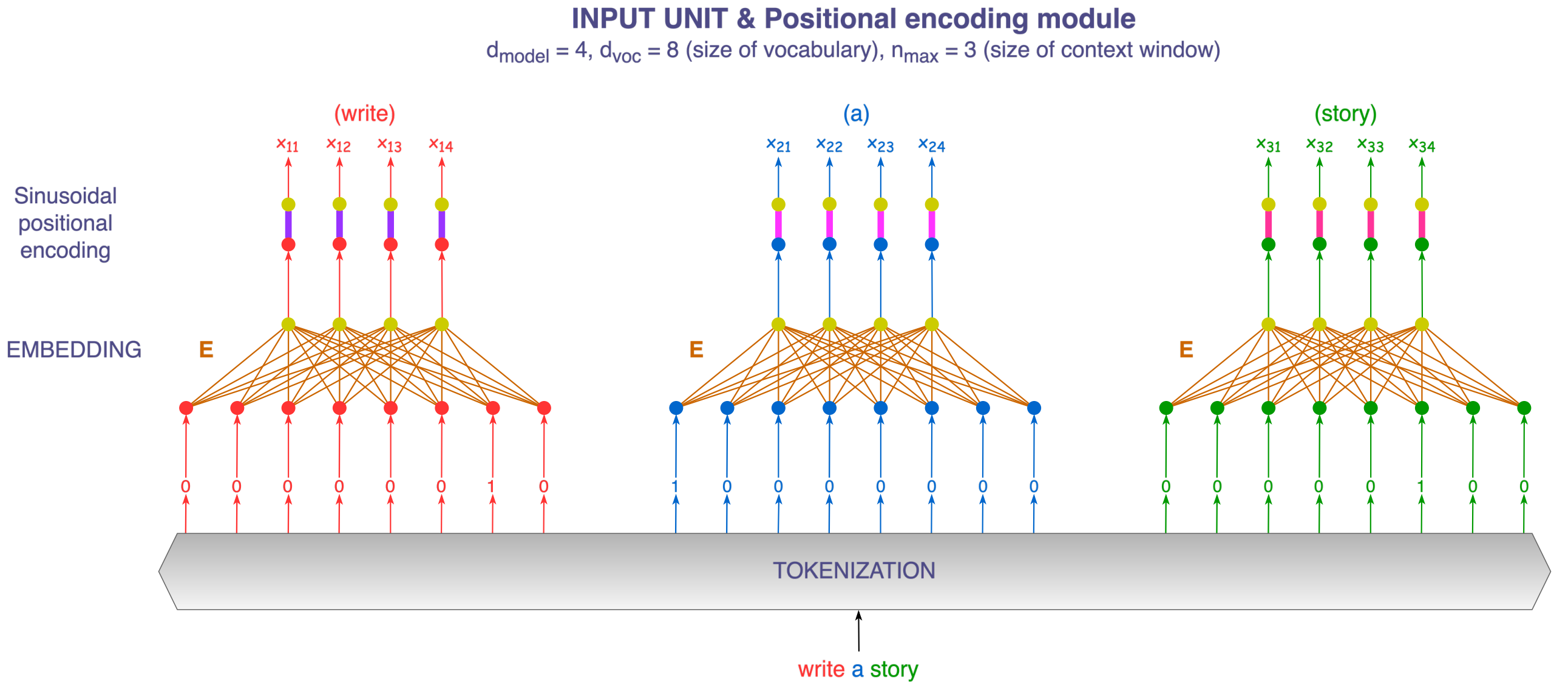

*Figure S8. Scheme of the system of simple neural networks corresponding to a residual connection module. The $n_{max}$ = 3 additive one-to-one networks have no fixed weights. They are dynamically determined by the output of the multi-head attention module, or the feed-forward network module. In the absence of such a signal, the connections have no weights and are therefore represented by black lines.*

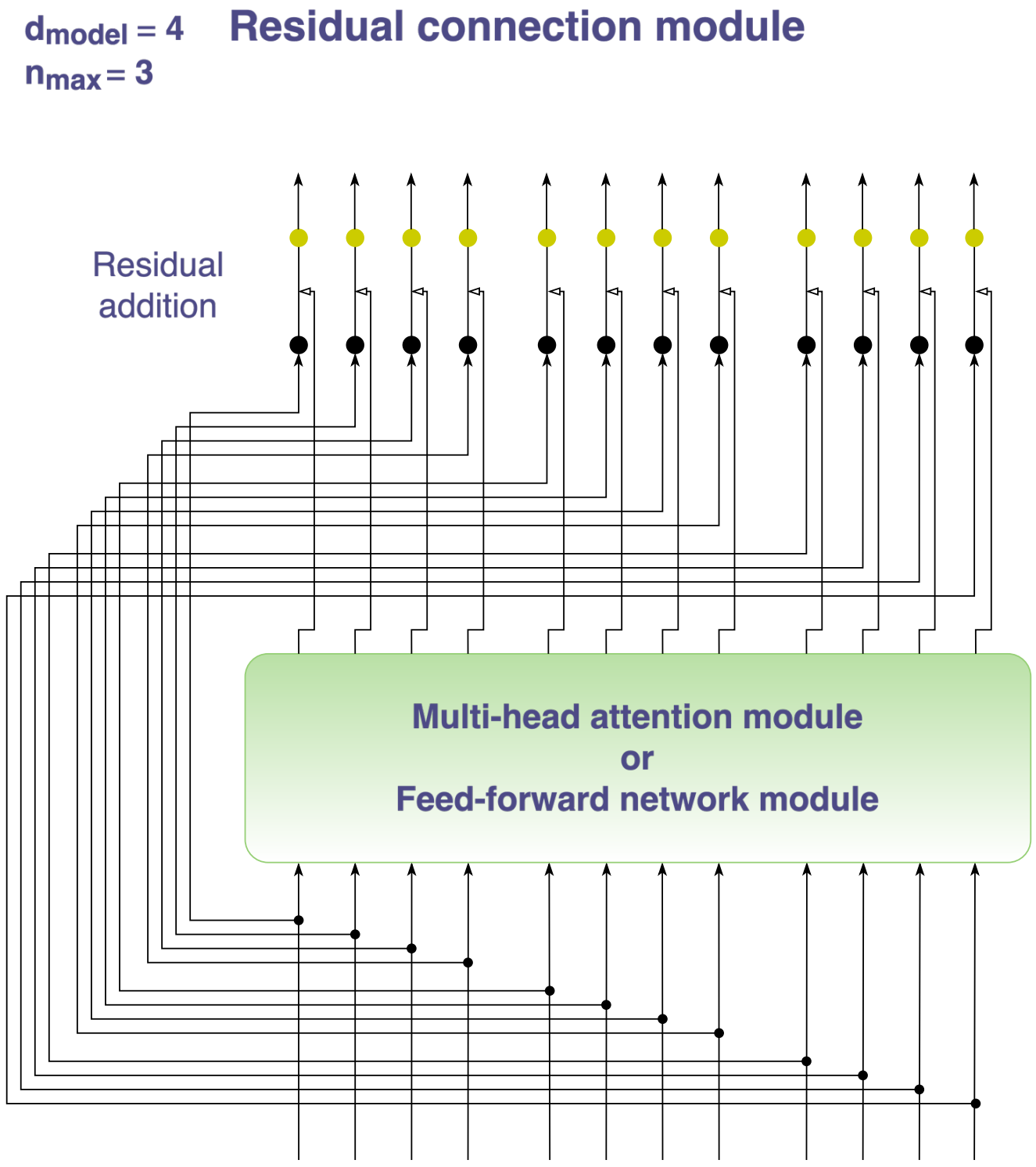

*Figure S8a. The same scheme with the networks activated by an input of n = 1 token and the corresponding output.*

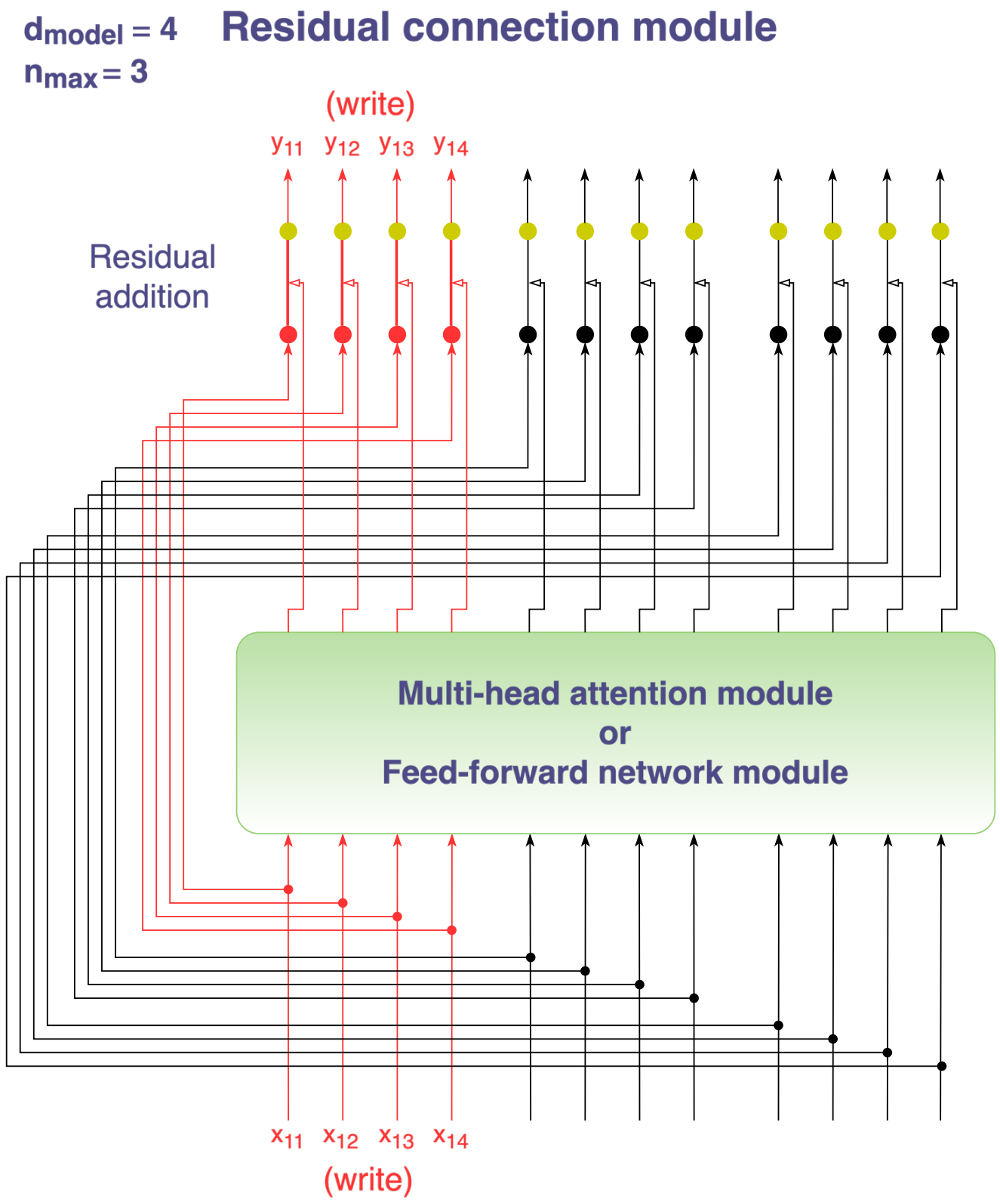

*Figure S8b. The same scheme with the networks activated by an input of n = 2 tokens and the corresponding output.*

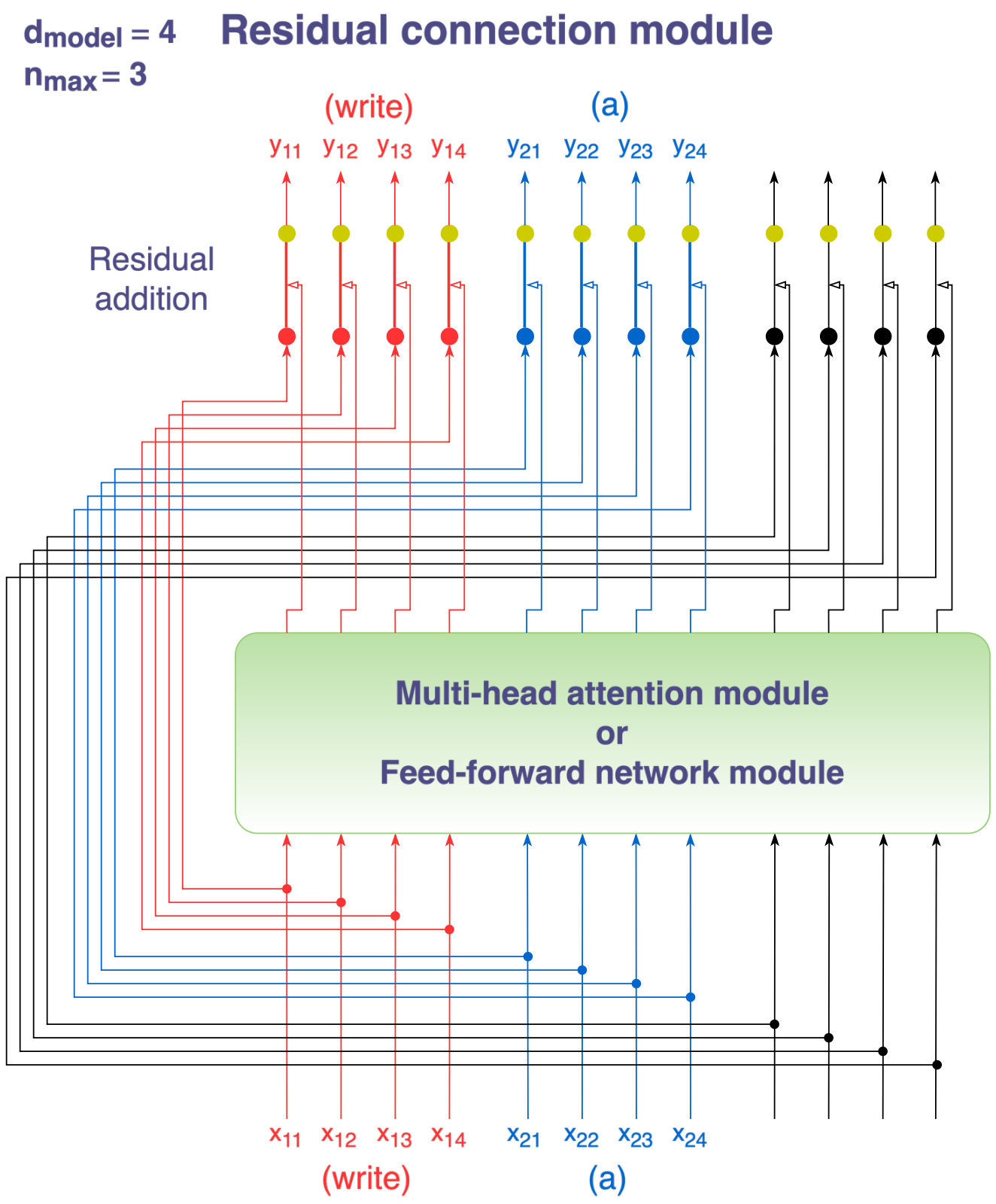

*Figure S8c. The same scheme with the networks activated by an input of n = 3 tokens and the corresponding output.*

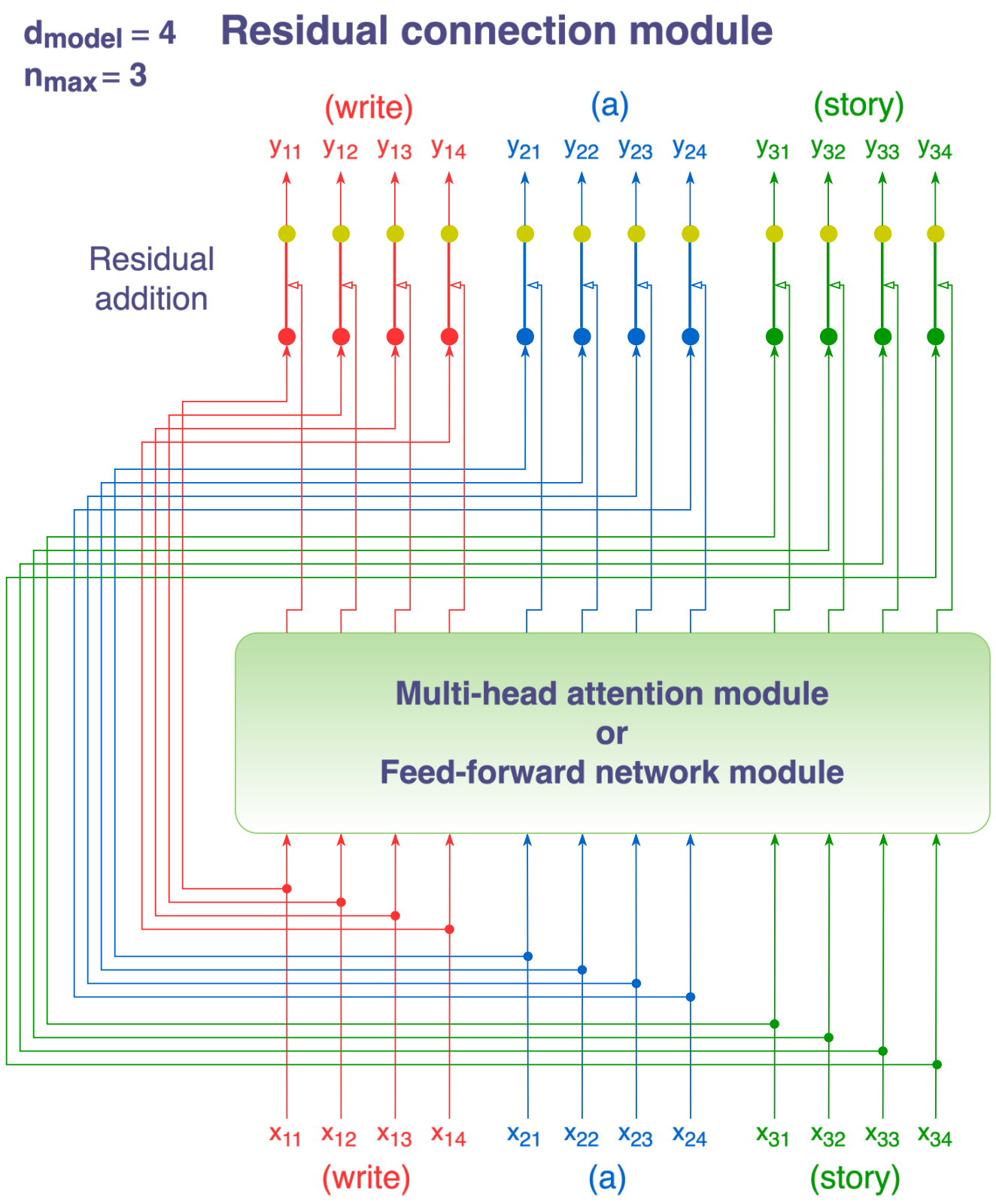

*Figure S9. Scheme of the system of simple neural networks corresponding to a multi-head attention module.*

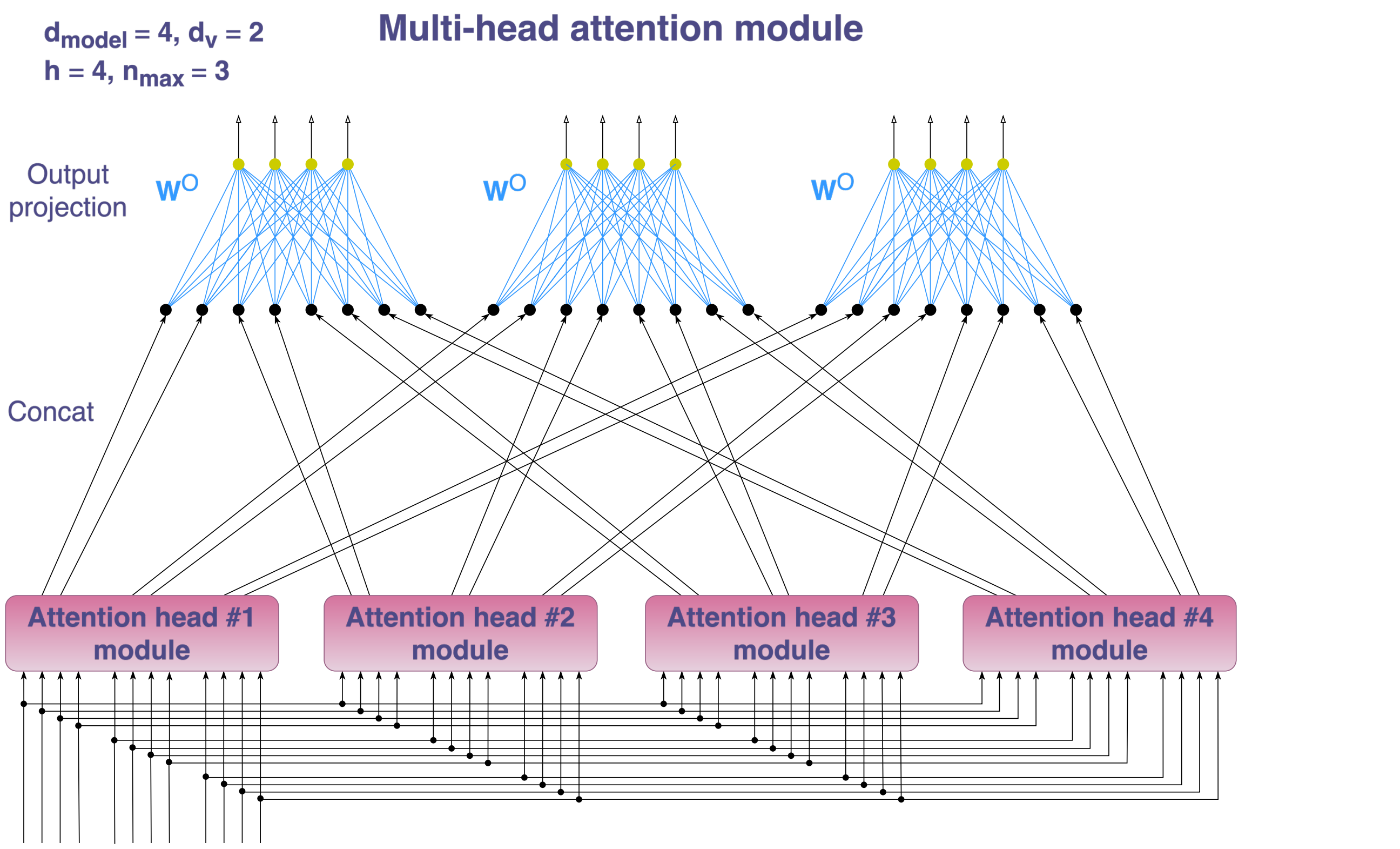

*Figure S9a. The same scheme with the networks activated by an input of n = 1 token and the corresponding output.*

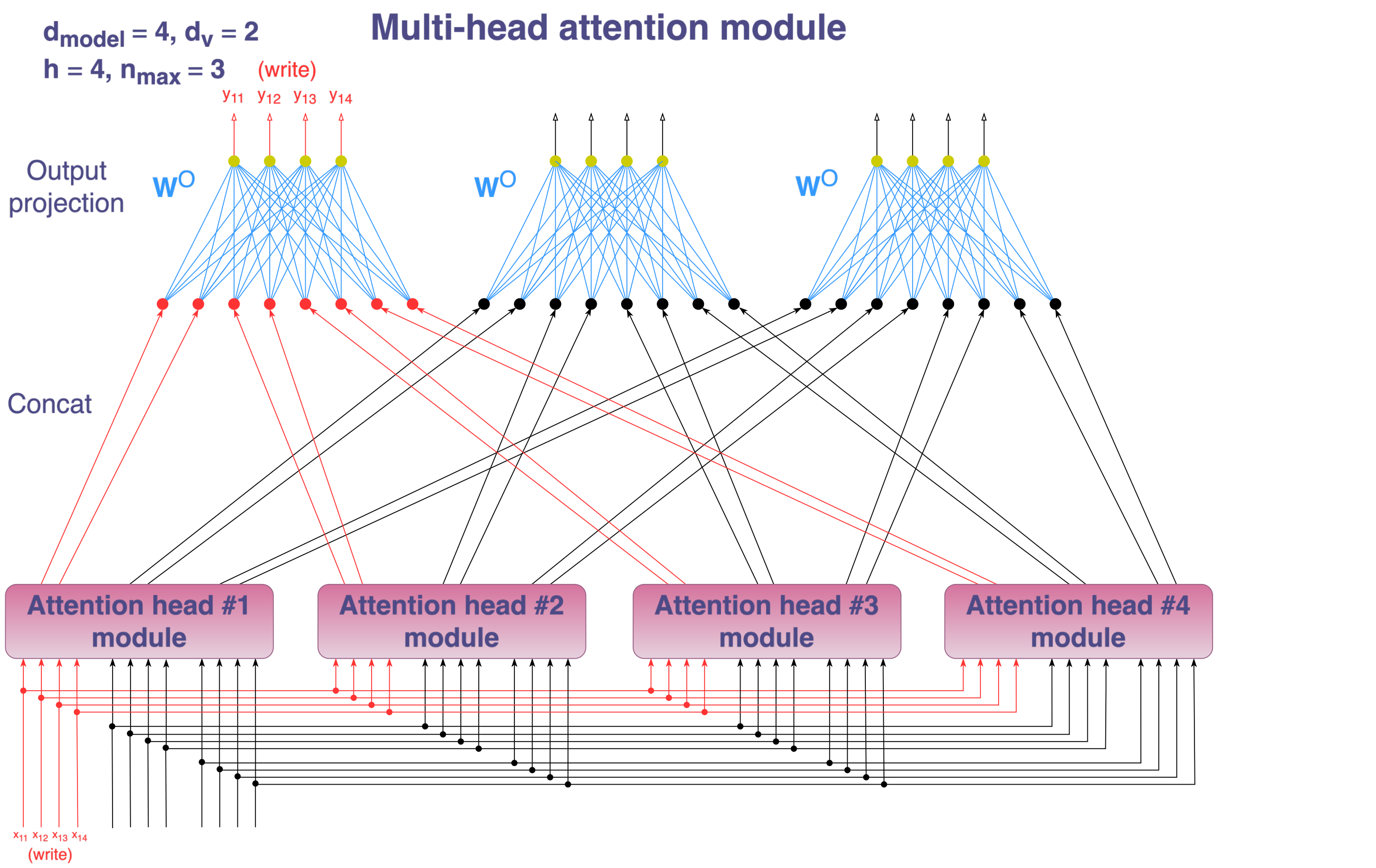

*Figure S9b. The same scheme with the networks activated by an input of n = 2 tokens and the corresponding output.*

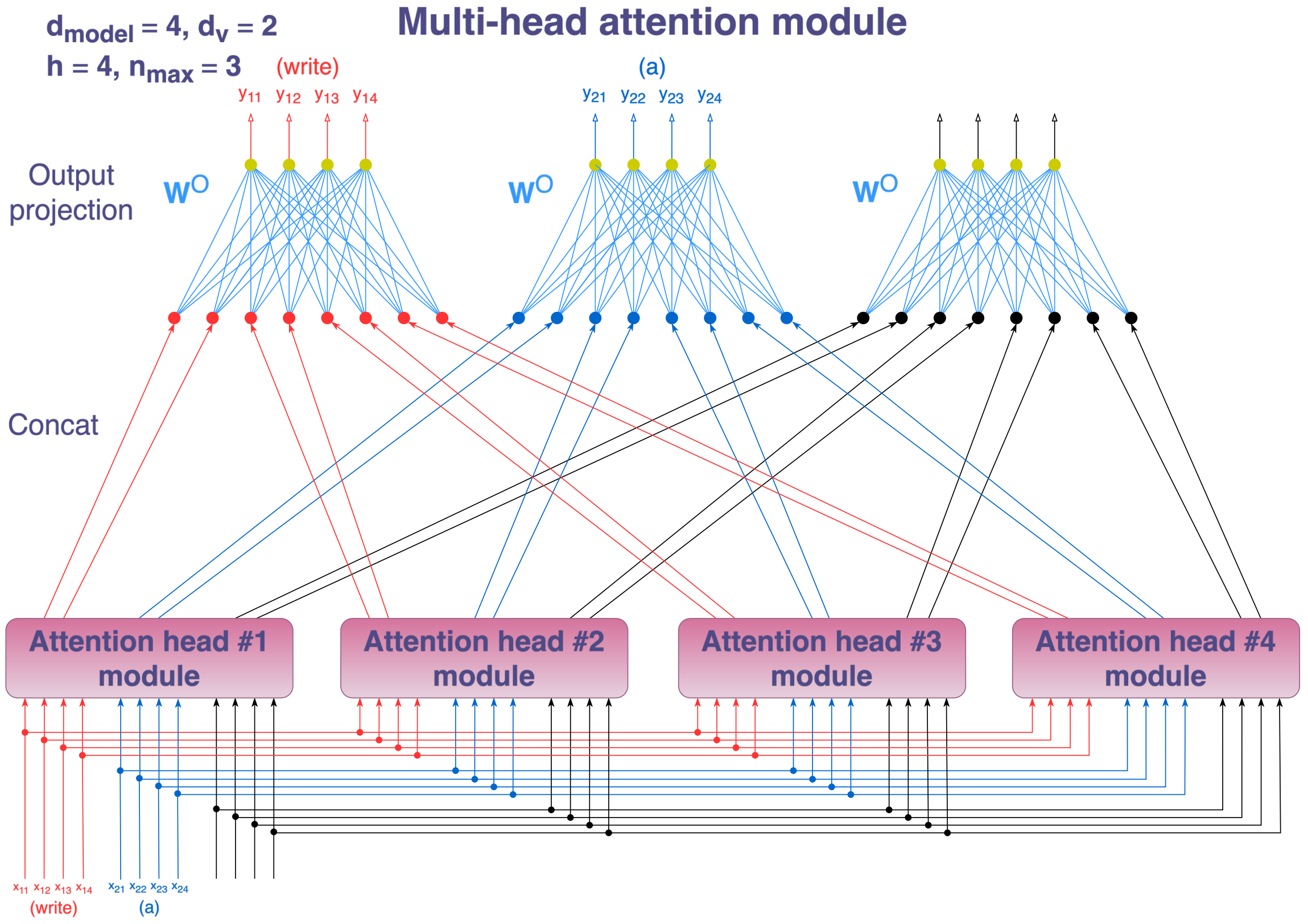

*Figure S9c. The same scheme with the networks activated by an input of n = 3 tokens and the corresponding output.*

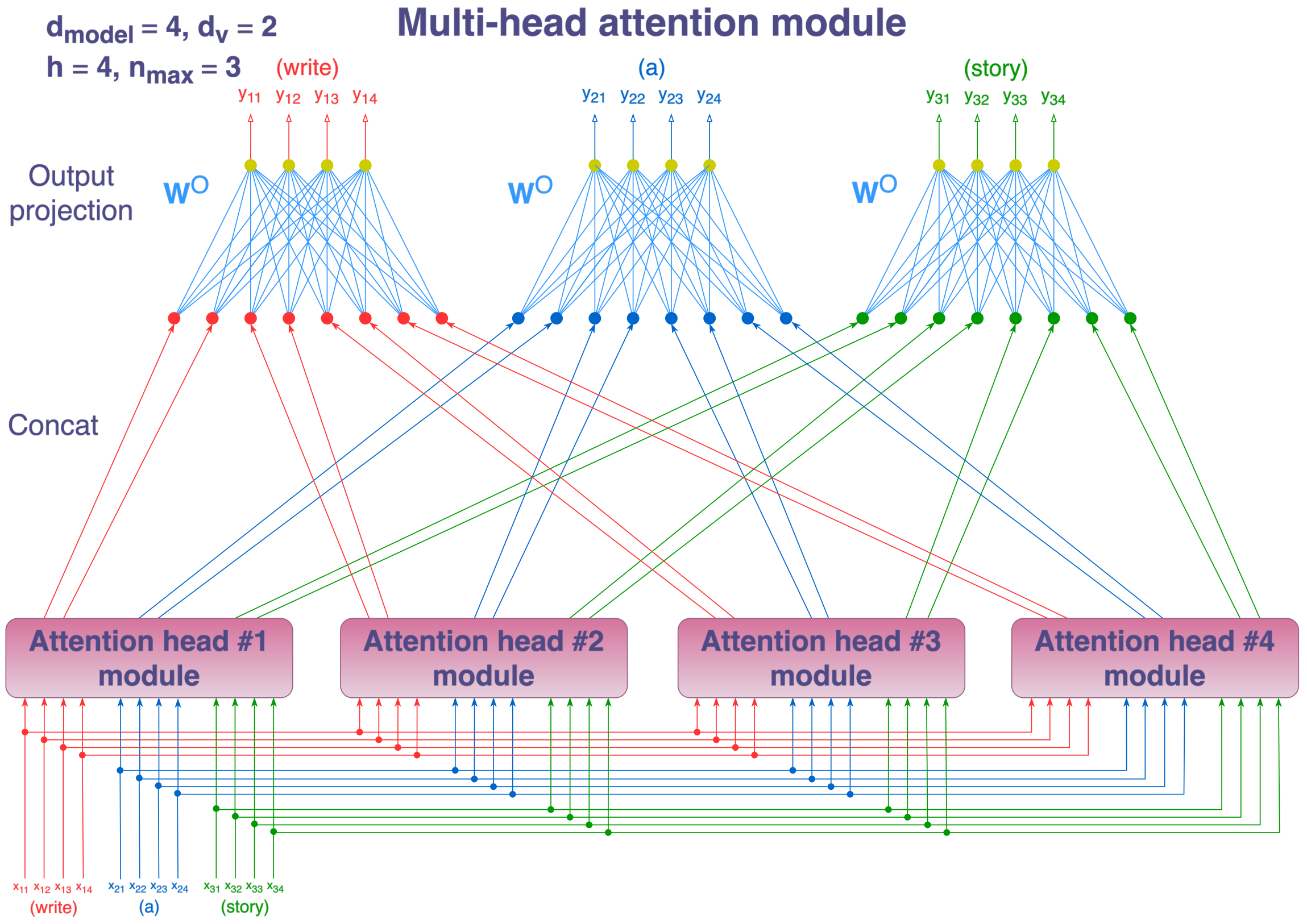

*Figure S10. Scheme of the system of simple neural networks corresponding to an autoregressive attention head with masking.*

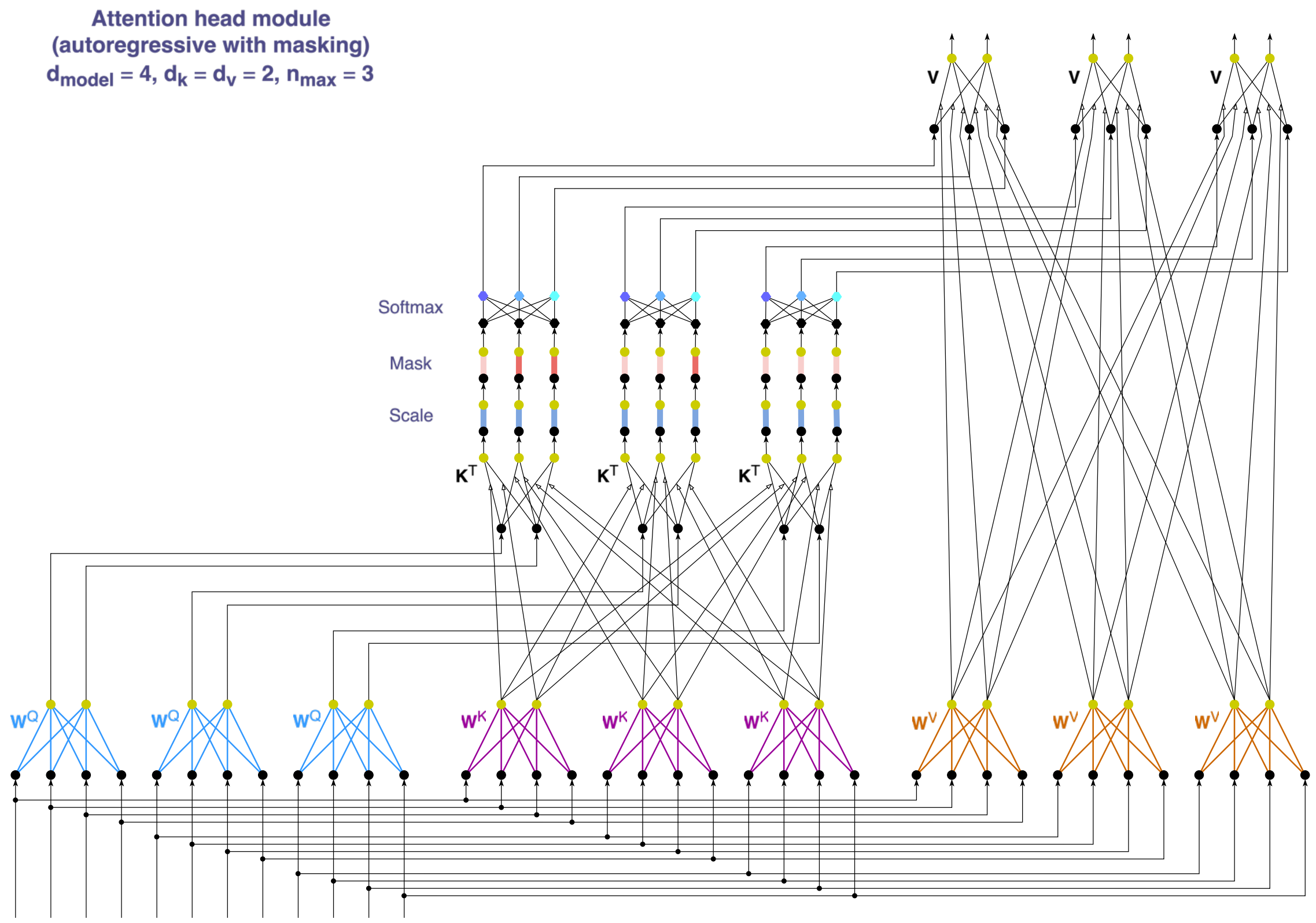

*Figure S10a. The same scheme with the networks activated by an input of n = 1 token and the corresponding output.*

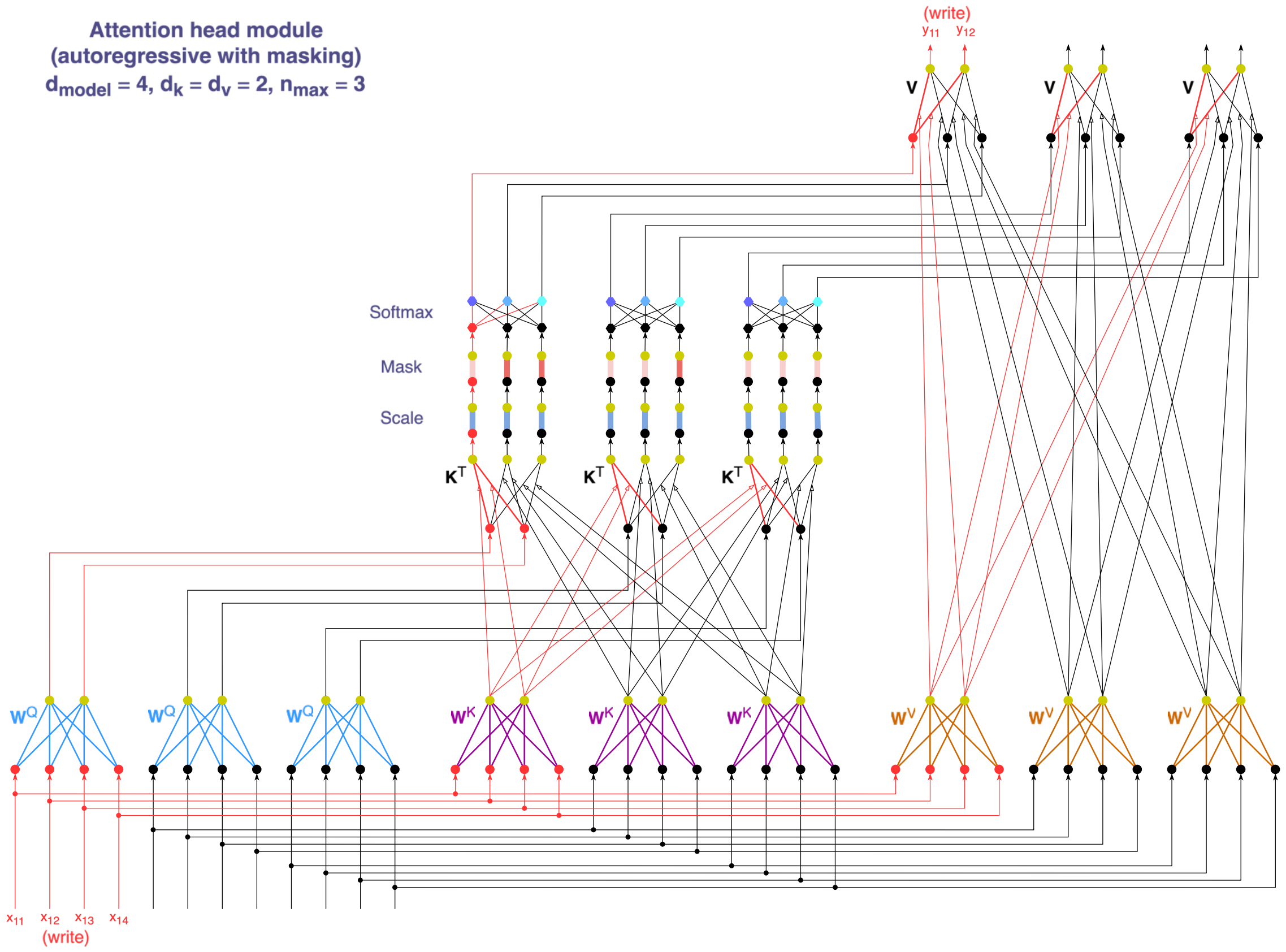

*Figure S10b. The same scheme with the networks activated by an input of n = 2 tokens and the corresponding output.*

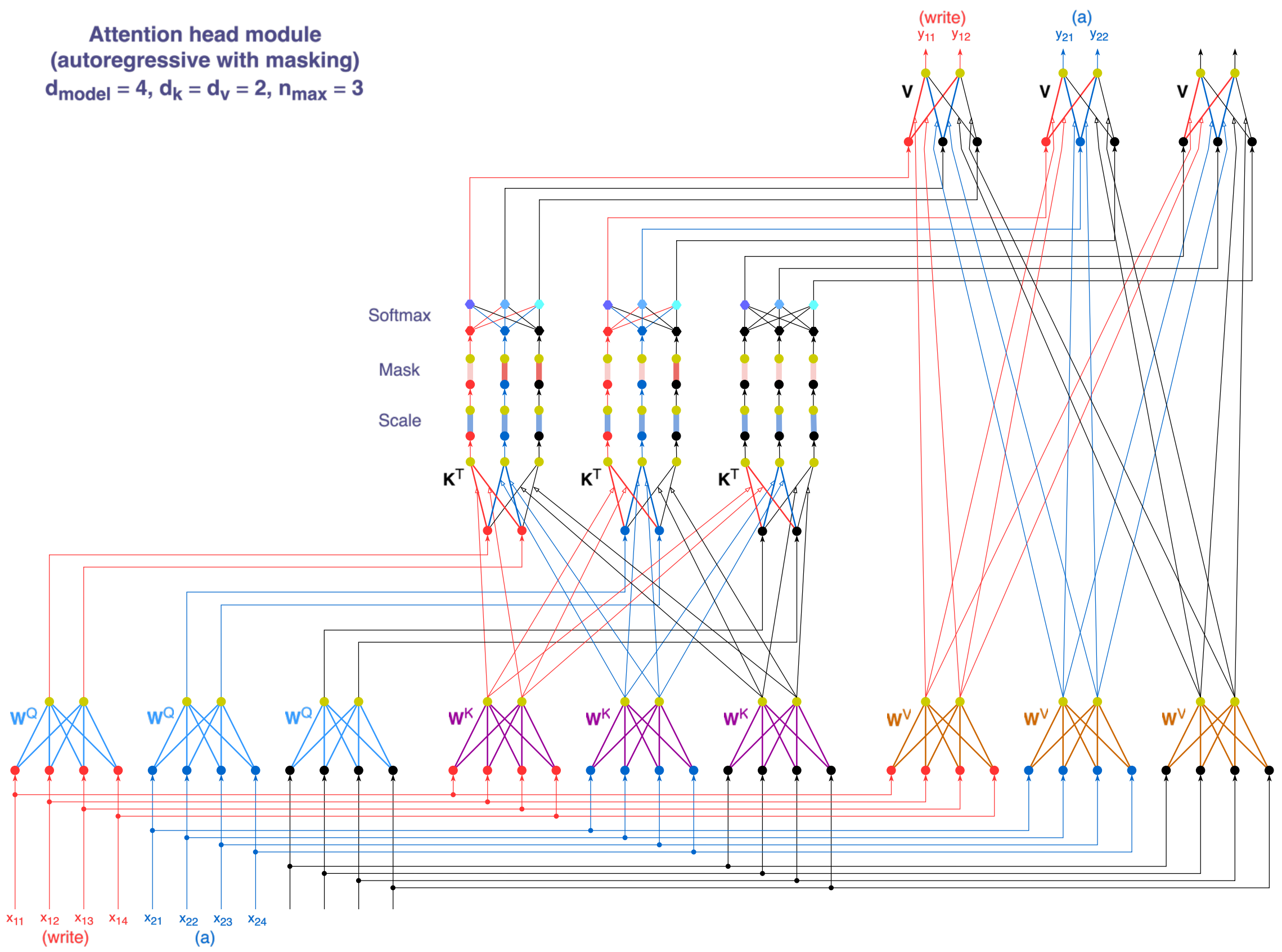

*Figure S10c. The same scheme with the networks activated by an input of n = 3 tokens and the corresponding output.*

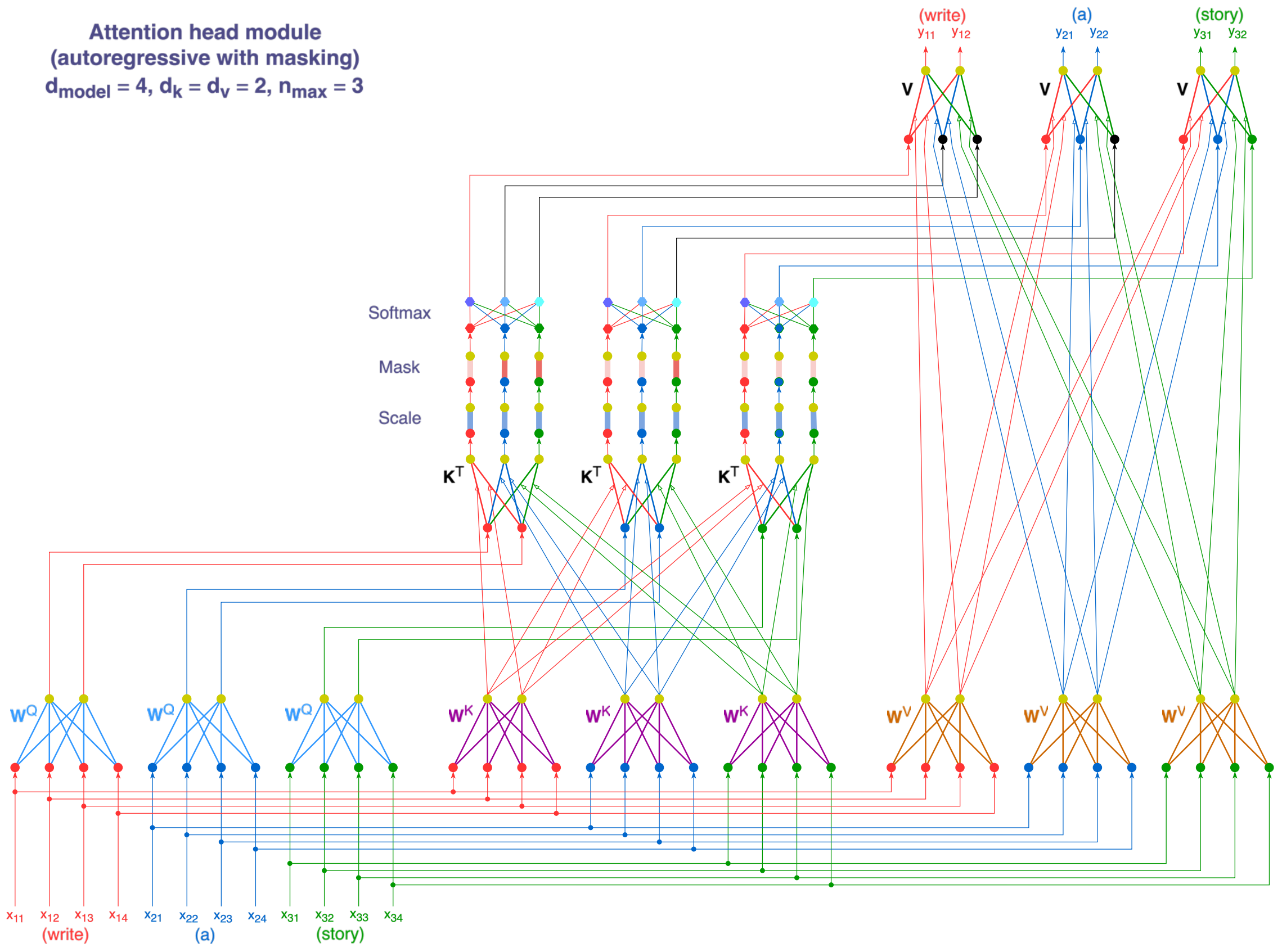

*Figure S10d. The same scheme with an input of n = 3 tokens given only to the group of simple networks corresponding to the operation $Mult[\boldsymbol{W}^V]$. Their output determines the weights of the third linear layer through interwoven output-weight interconnections, and the attention head produces no output.*

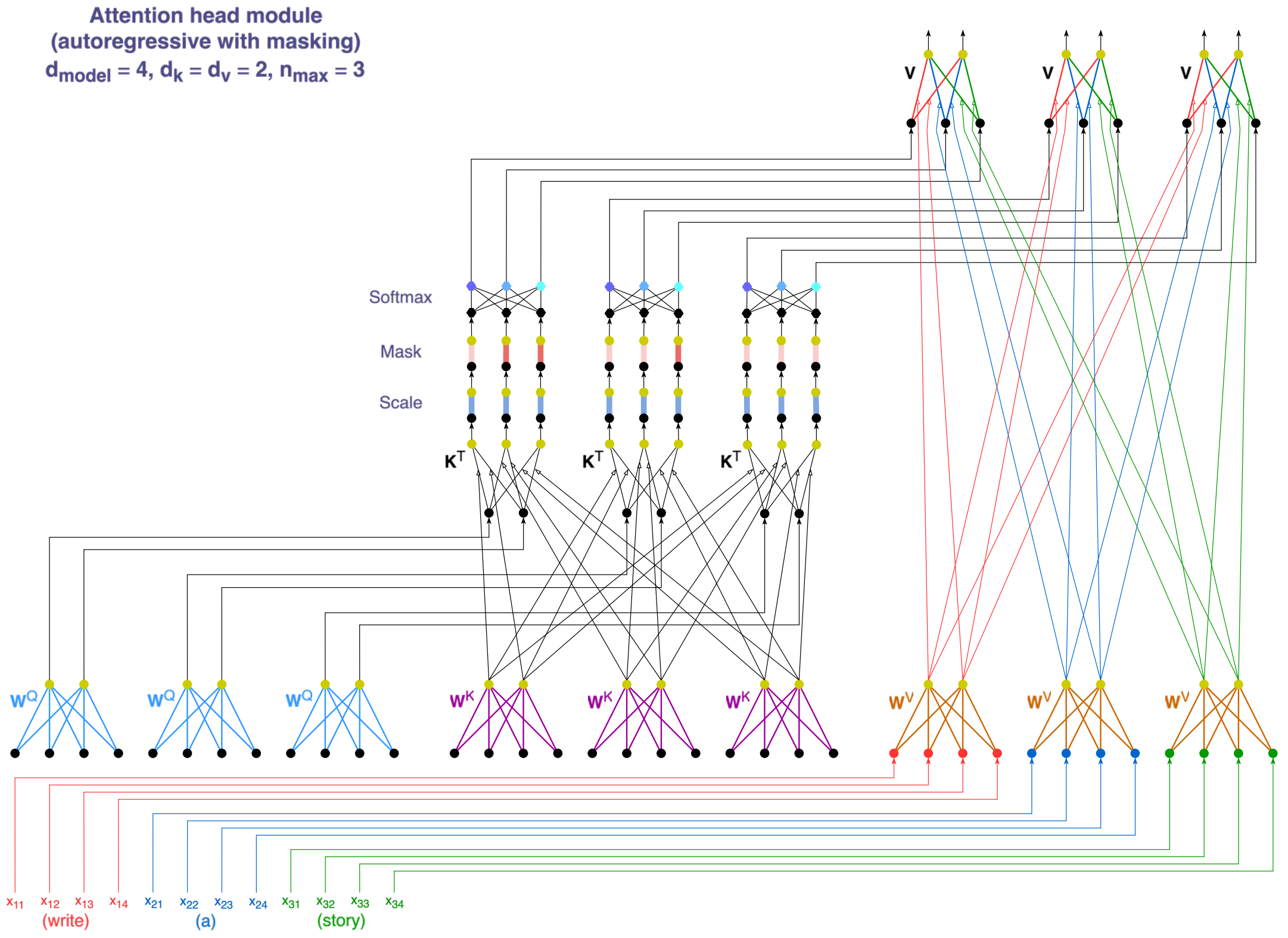

*Figure S10e. The same scheme with an input of n = 3 tokens given only to the two groups of simple networks corresponding to the operations $Mult[\boldsymbol{W}^V]$ and $Mult[\boldsymbol{W}^K]$. Their output determines the weights of the third and second linear layer through interwoven output-weight interconnections, and the attention head produces no output.*

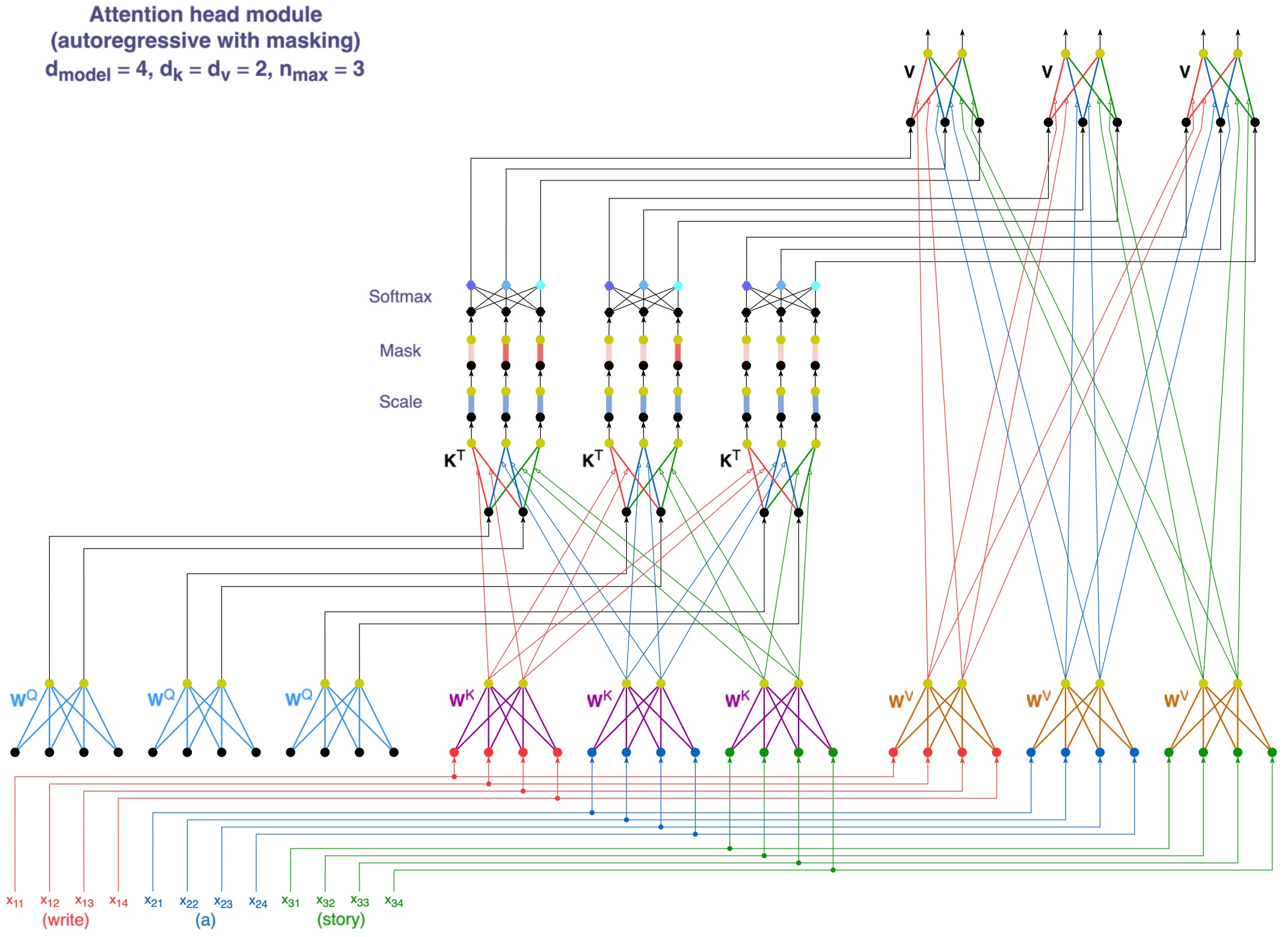

*Figure S11. Scheme of the system of simple neural networks corresponding to a bidirectional attention head.*

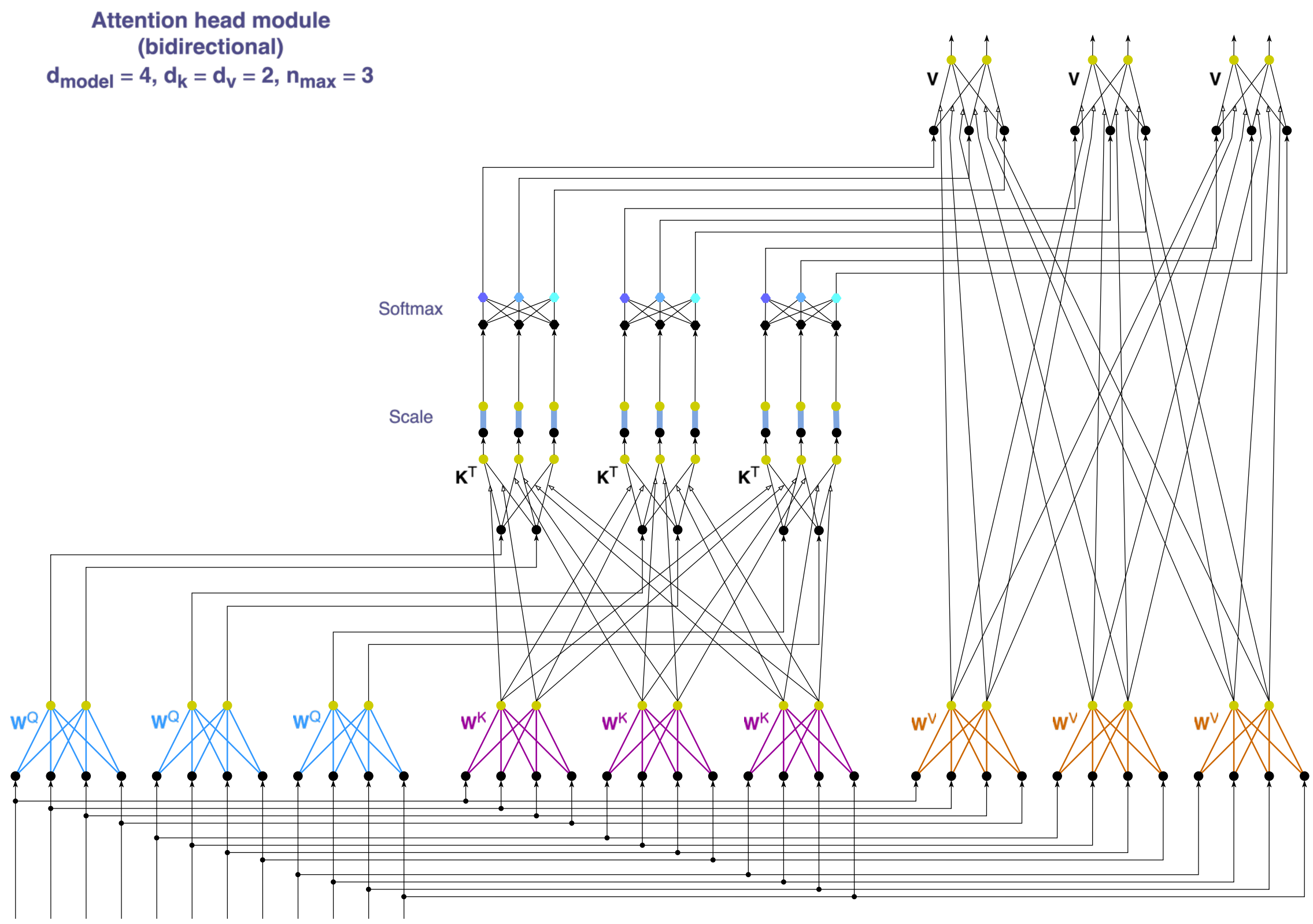

*Figure S11a. The same scheme with the networks activated by an input of n = 1 token and the corresponding output.*

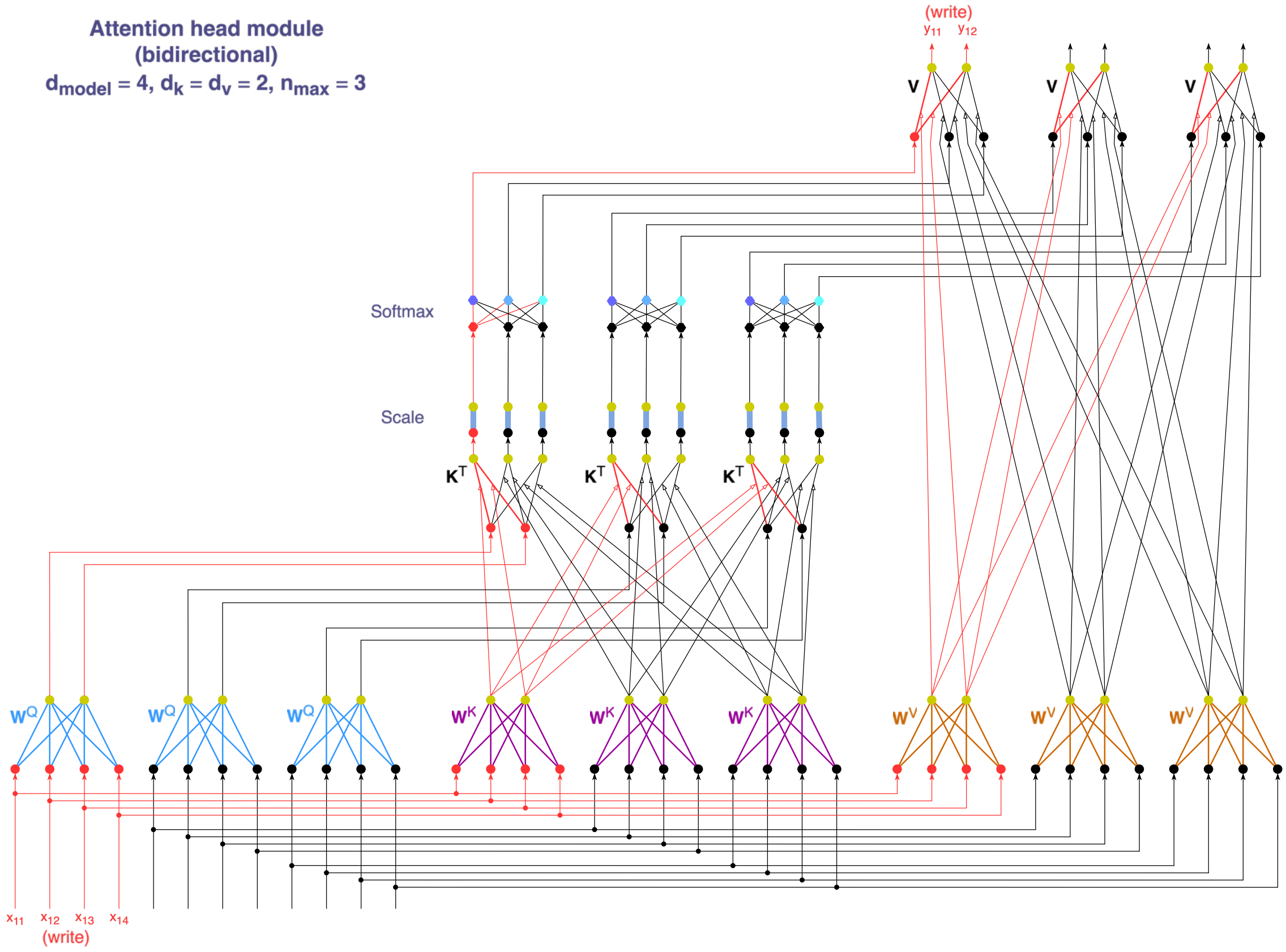

*Figure S11b. The same scheme with the networks activated by an input of n = 2 tokens and the corresponding output.*

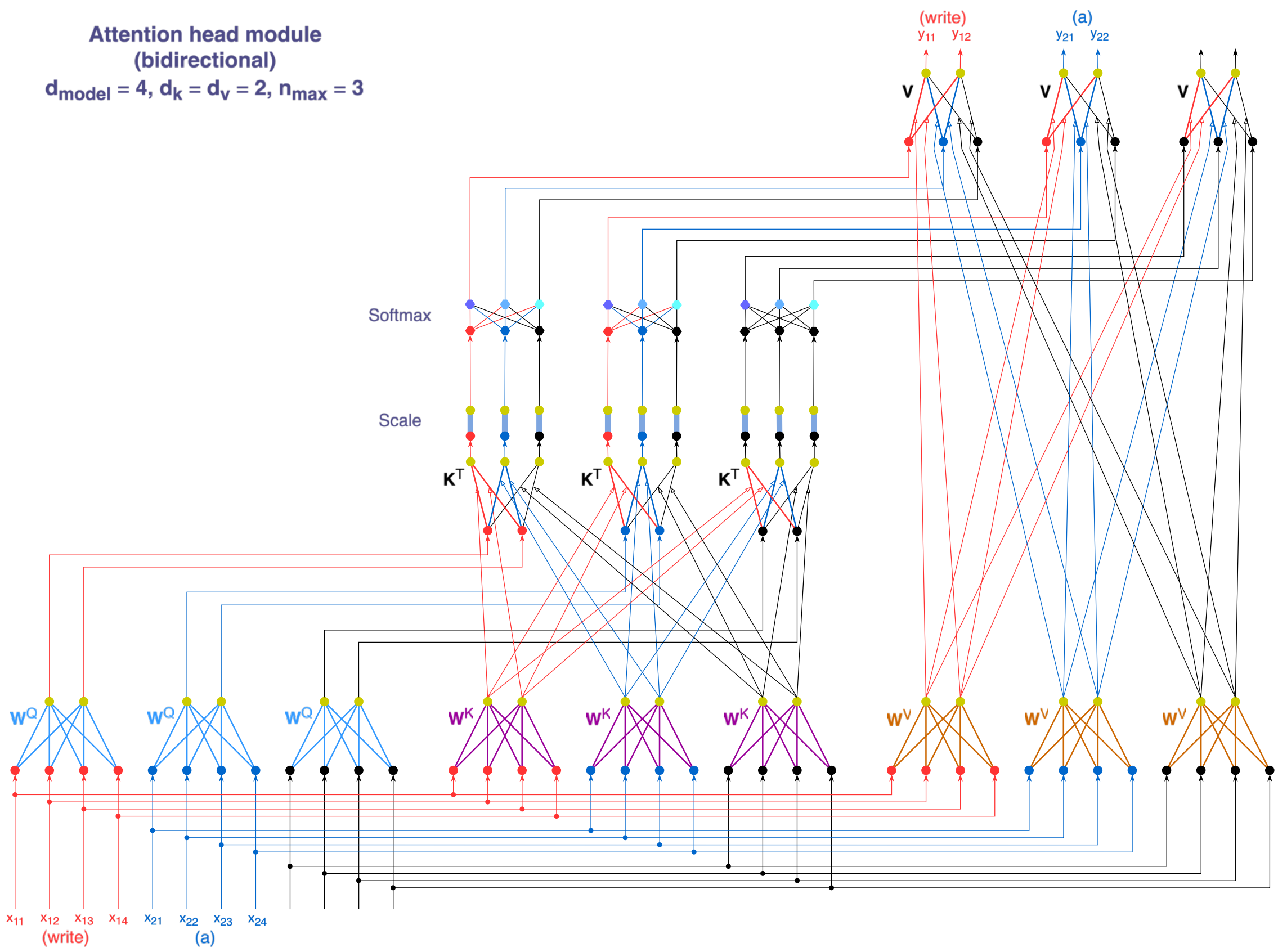

*Figure S11c. The same scheme with the networks activated by an input of n = 3 tokens and the corresponding output.*

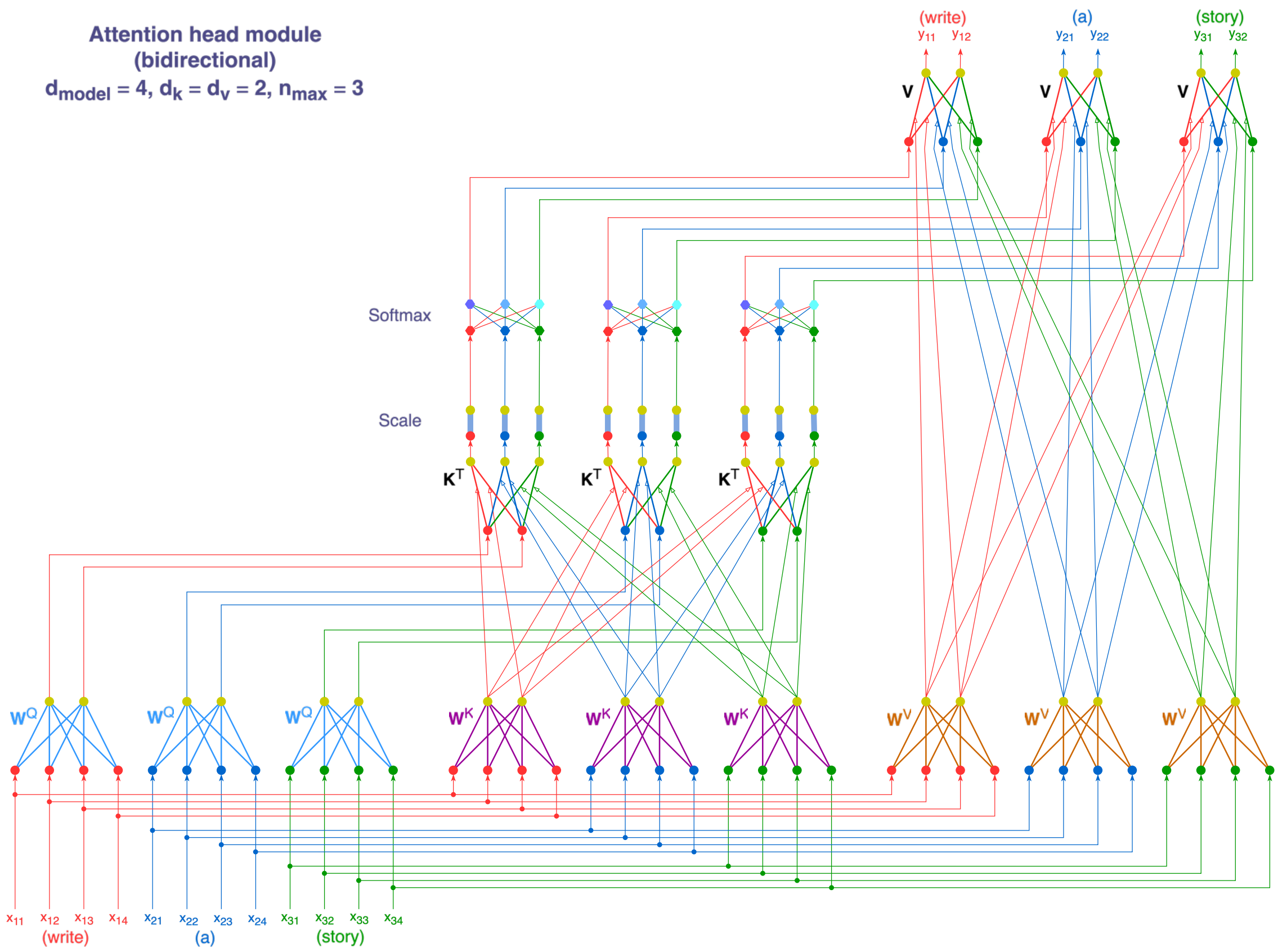

*Figure S11d. The same scheme with an input of n = 3 tokens given only to the group of simple networks corresponding to the operation $Mult[\boldsymbol{W}^V]$. Their output determines the weights of the third linear layer through interwoven output-weight interconnections, and the attention head produces no output.*

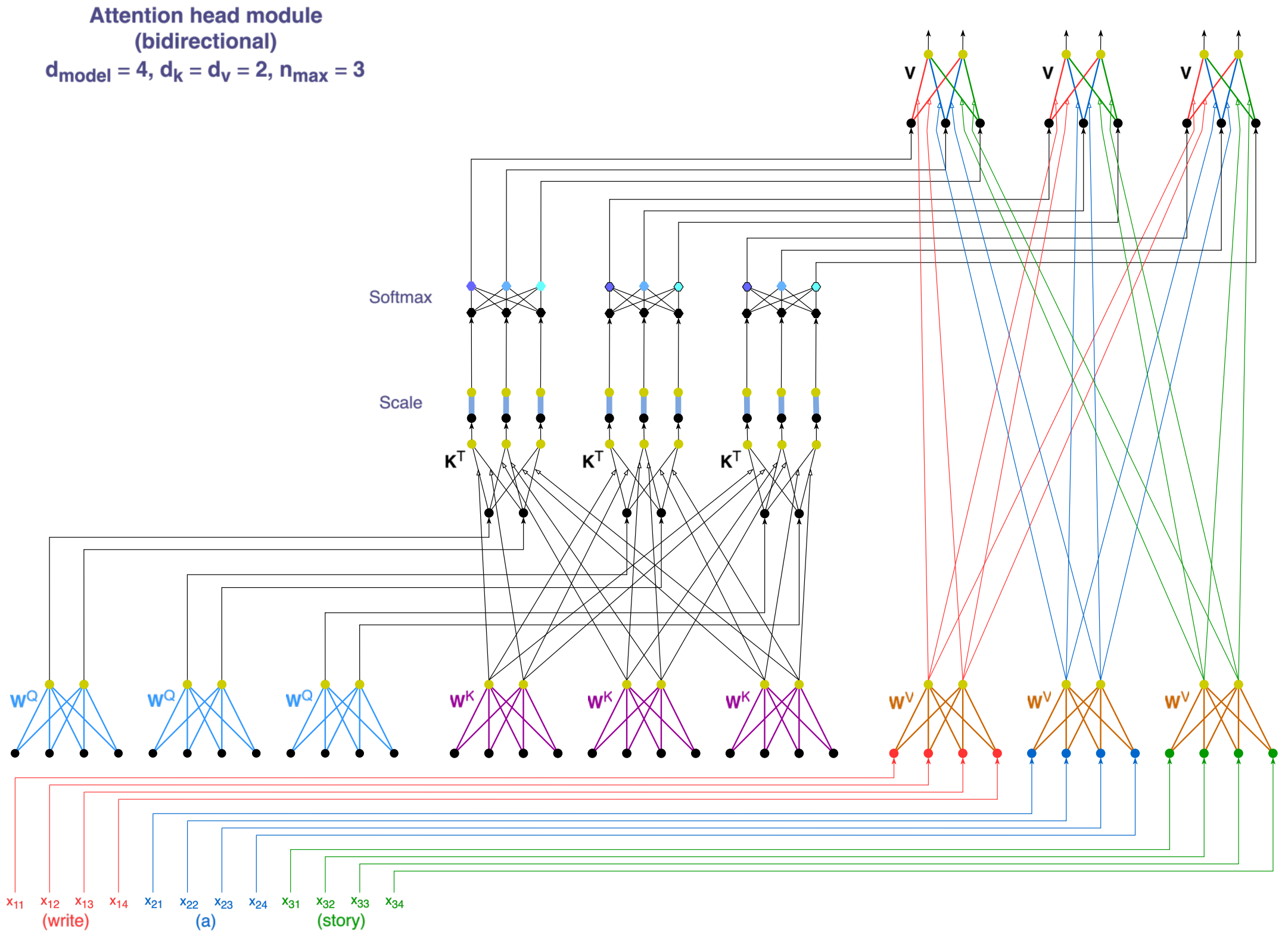

*Figure S11e. The same scheme with an input of n = 3 tokens given only to the two groups of simple networks corresponding to the operations $Mult[\boldsymbol{W}^V]$ and $Mult[\boldsymbol{W}^K]$. Their output determines the weights of the third and second linear layer through interwoven output-weight interconnections, and the attention head produces no output.*

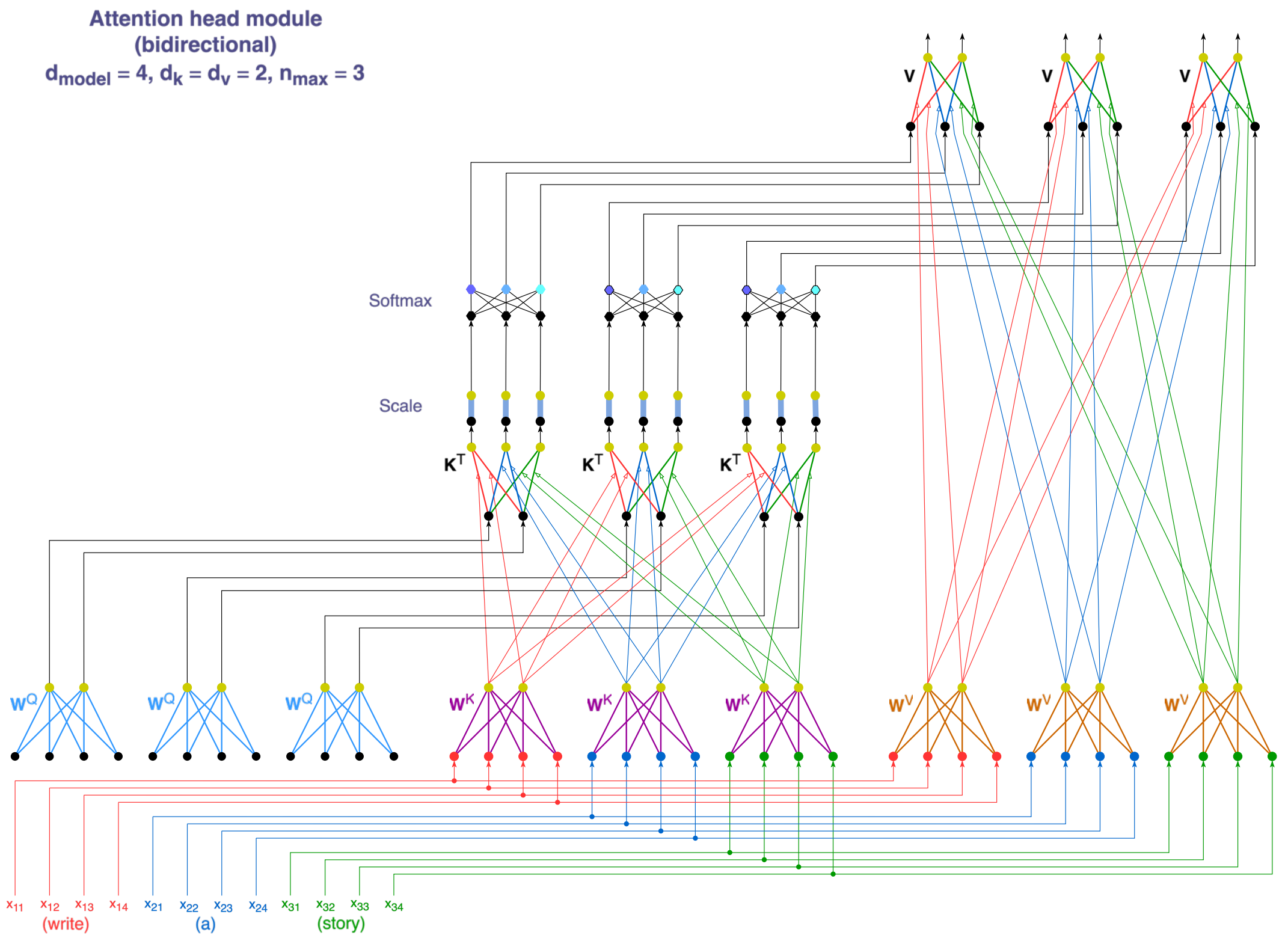

*Figure S12. Scheme of the system of simple neural networks corresponding to an autoregressive attention head without masking. See Fig. S6 and § S4 for the functional diagram.*

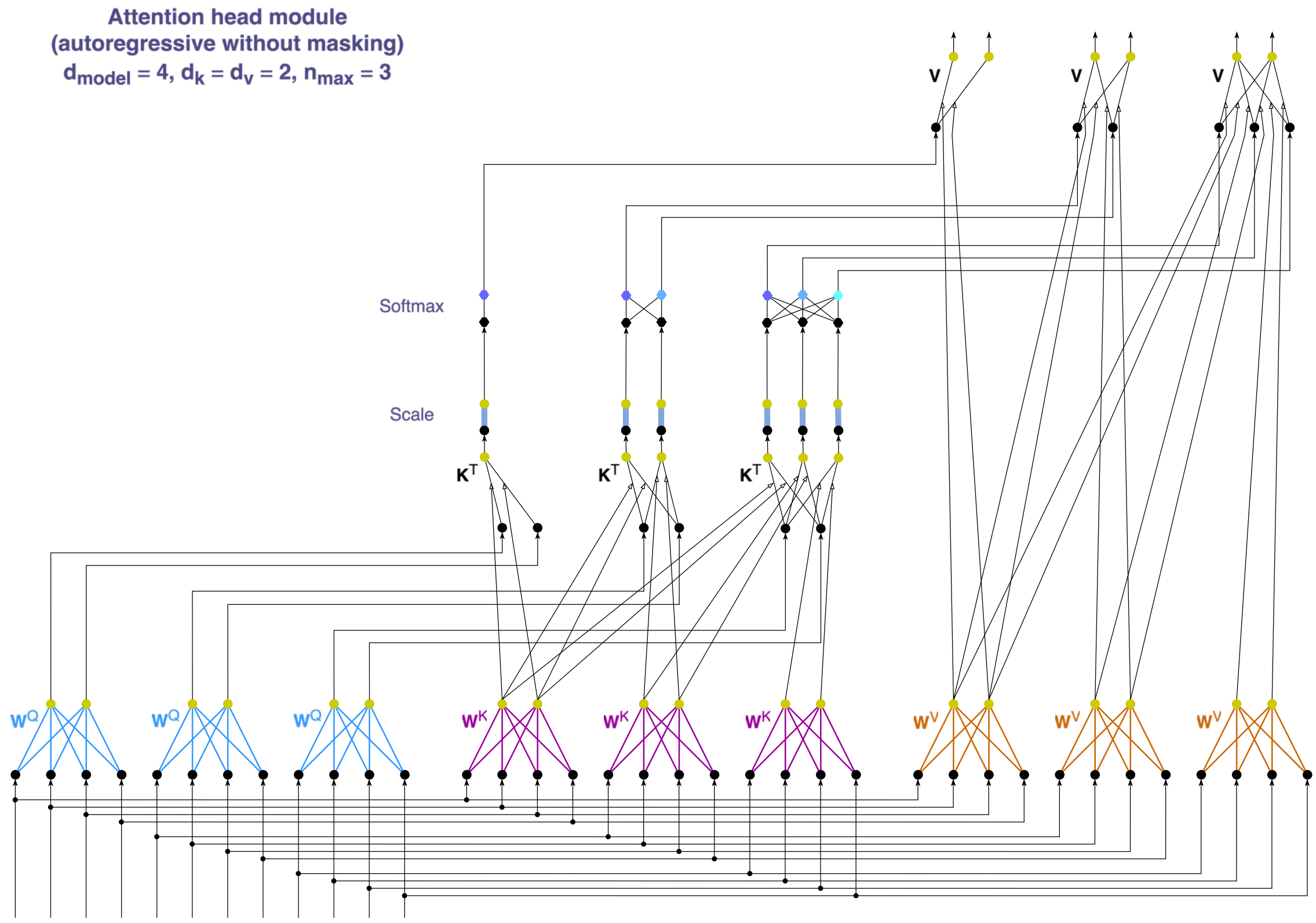

*Figure S12a. The same scheme with the networks activated by an input of n = 1 token and the corresponding output.*

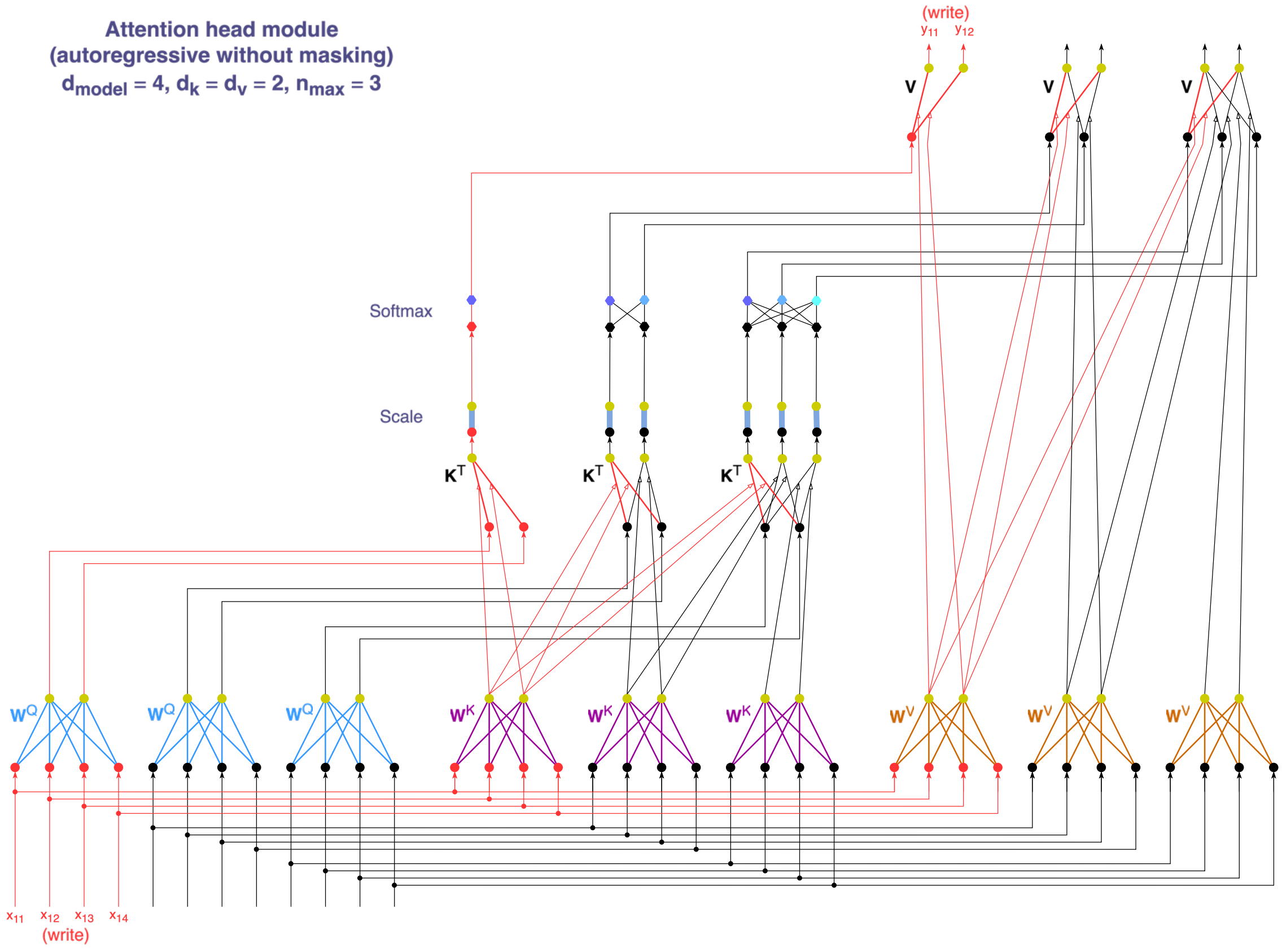

*Figure S12b. The same scheme with the networks activated by an input of n = 2 tokens and the corresponding output.*

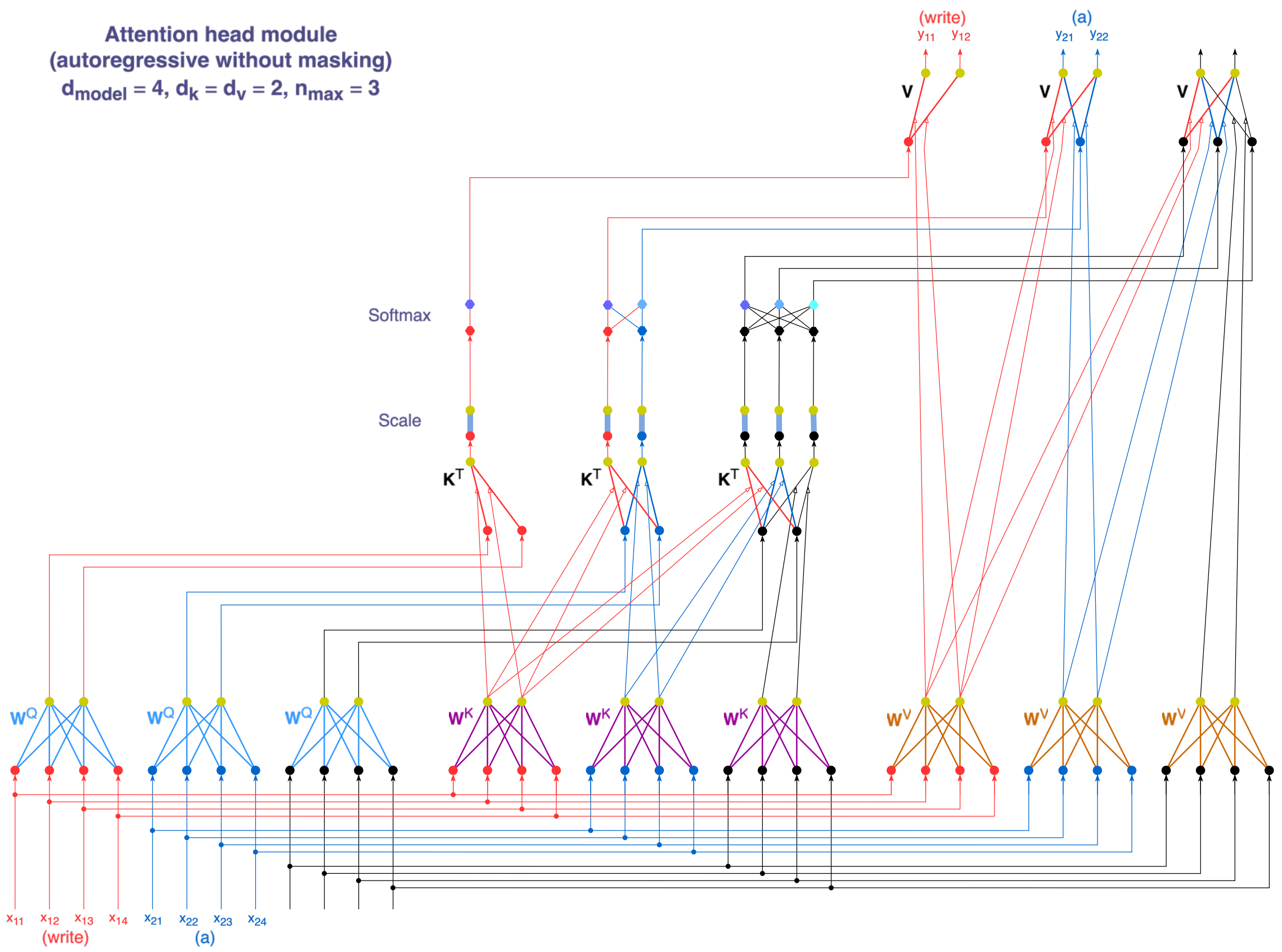

*Figure S12c. The same scheme with the networks activated by an input of n = 3 tokens and the corresponding output.*

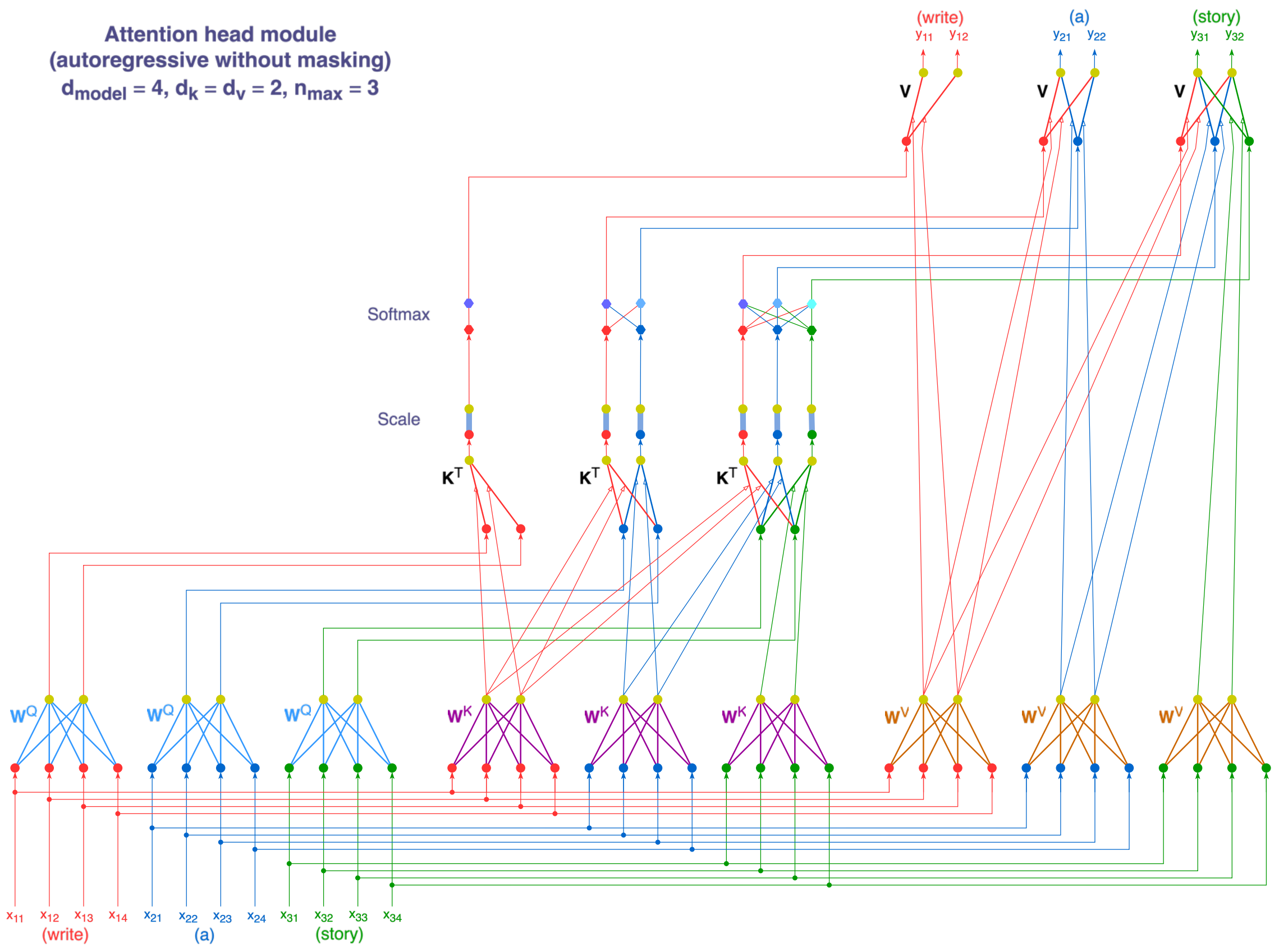

*Figure S12d. The same scheme with an input of n = 3 tokens given only to the group of simple networks corresponding to the operation $Mult[\boldsymbol{W}^{V}]$. Their output determines the weights of the third linear layer through interwoven output-weight interconnections, and the attention head produces no output.*

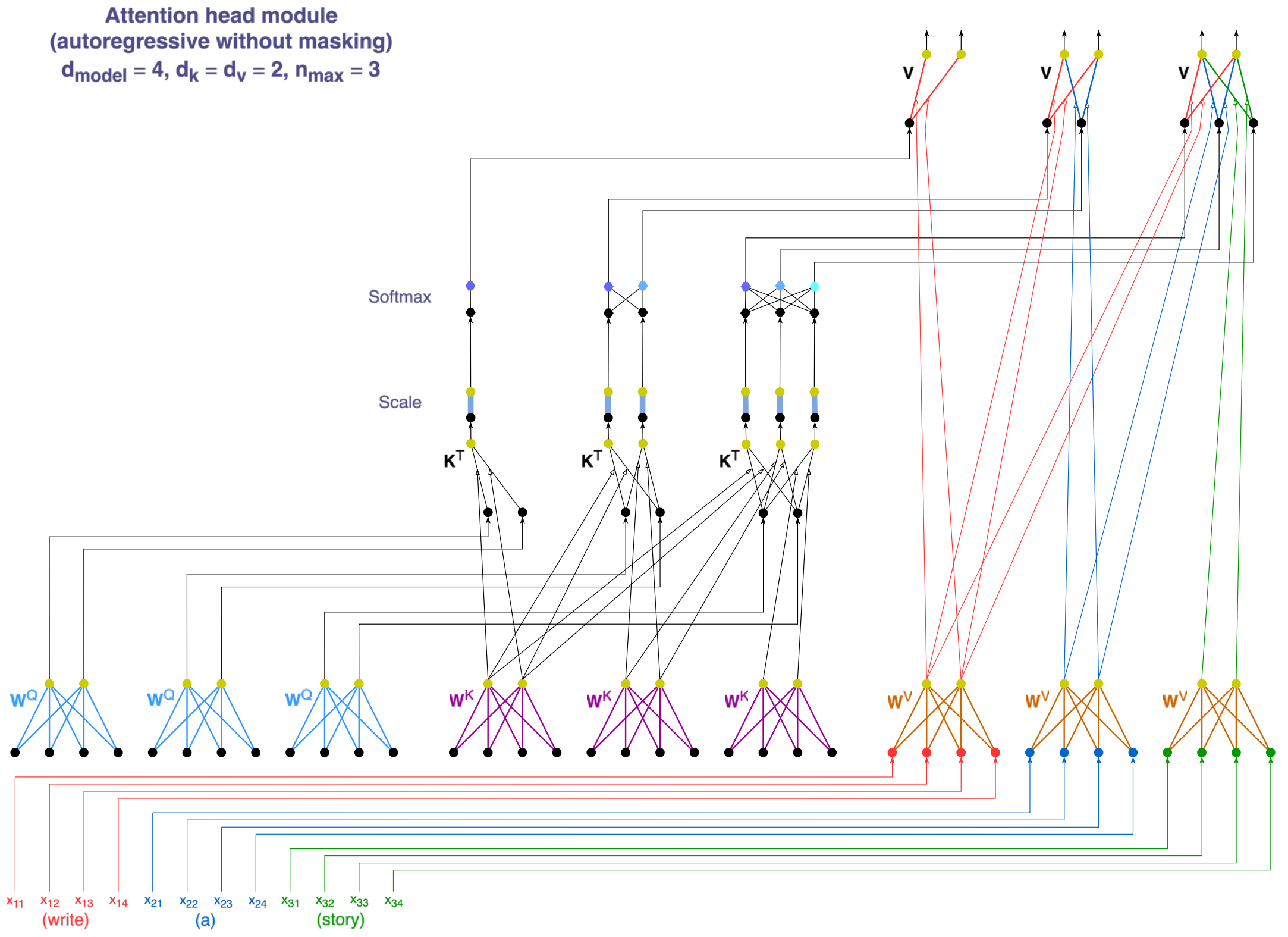

*Figure S12e. The same scheme with an input of n = 3 tokens given only to the two groups of simple networks corresponding to the operations $Mult[\boldsymbol{W}^{V}]$ and $Mult[\boldsymbol{W}^{K}]$. Their output determines the weights of the third and second linear layer through interwoven output-weight interconnections, and the attention head produces no output.*

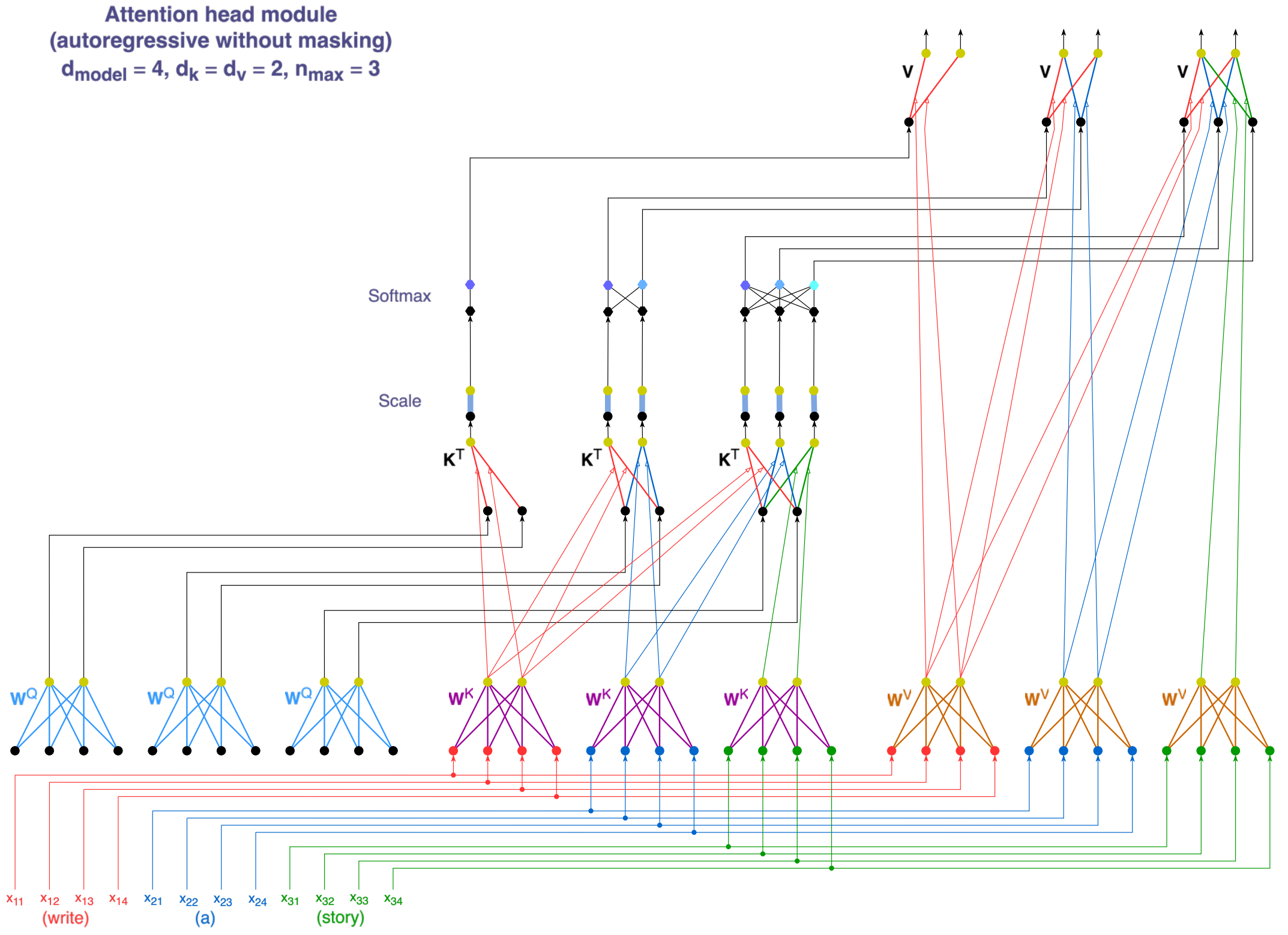

*Figure S13. Scheme of the system of simple neural networks corresponding to a layer normalization module. The first layer normalizes each token vector across its features, producing zero mean and approximately unit variance. Each of the following two layers adjusts the mean and/or variance of the features of each token by applying scaling and bias, respectively.*

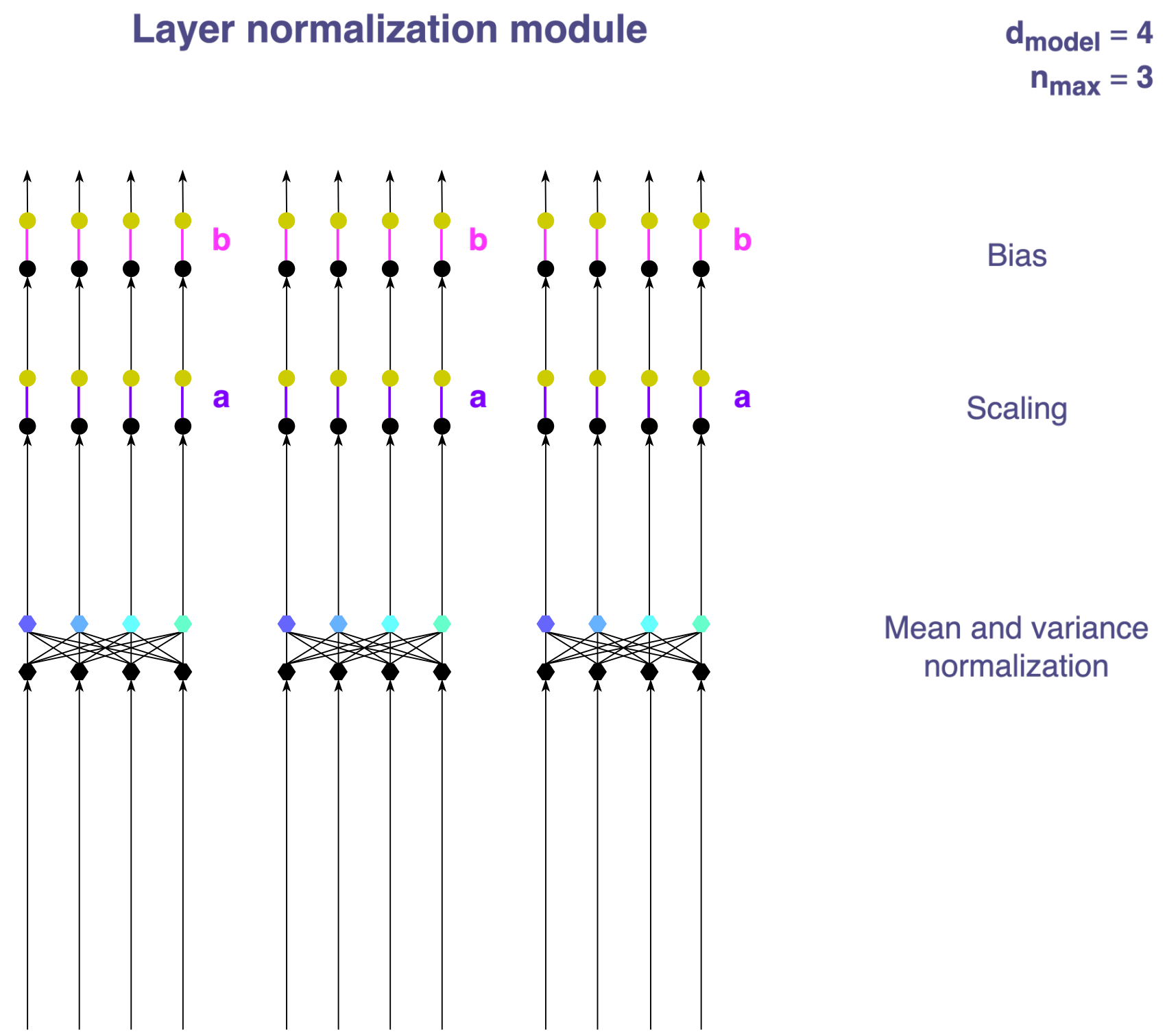

*Figure S13a. The same scheme with the networks activated by an input of n = 1 token and the corresponding output.*

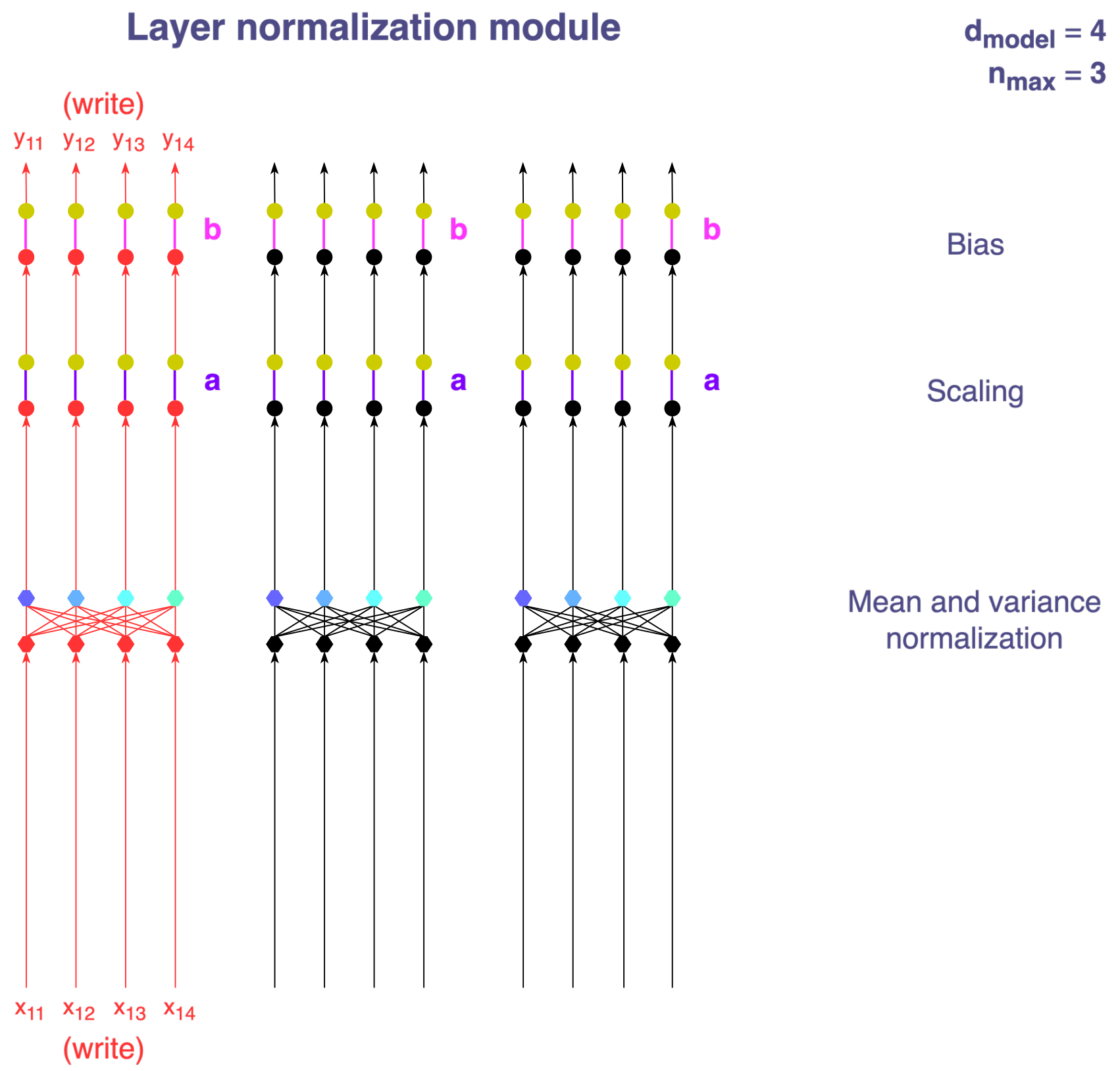

*Figure S13b. The same scheme with the networks activated by an input of n = 2 tokens and the corresponding output.*

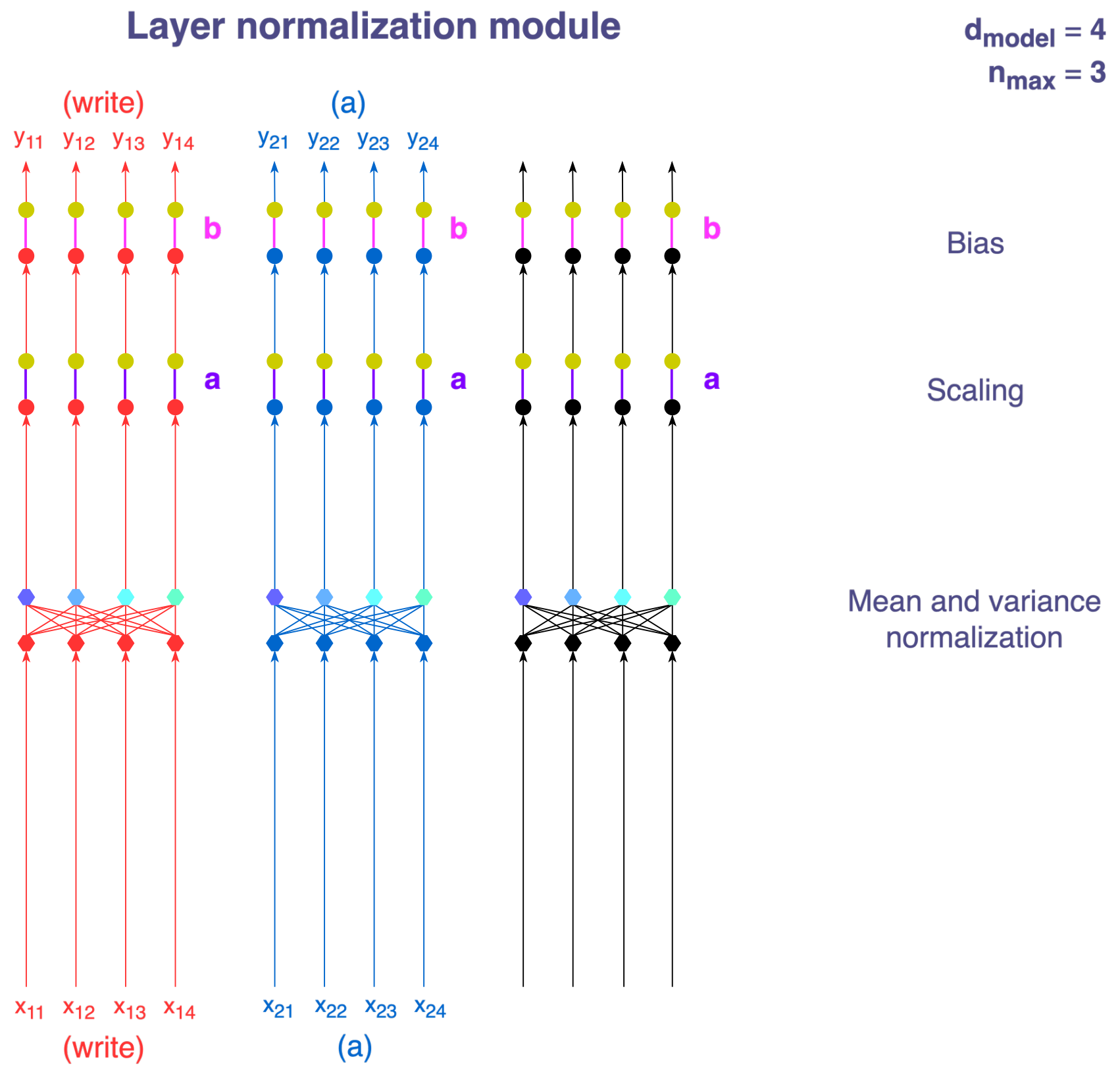

*Figure S13c. The same scheme with the networks activated by an input of n = 3 tokens and the corresponding output.*

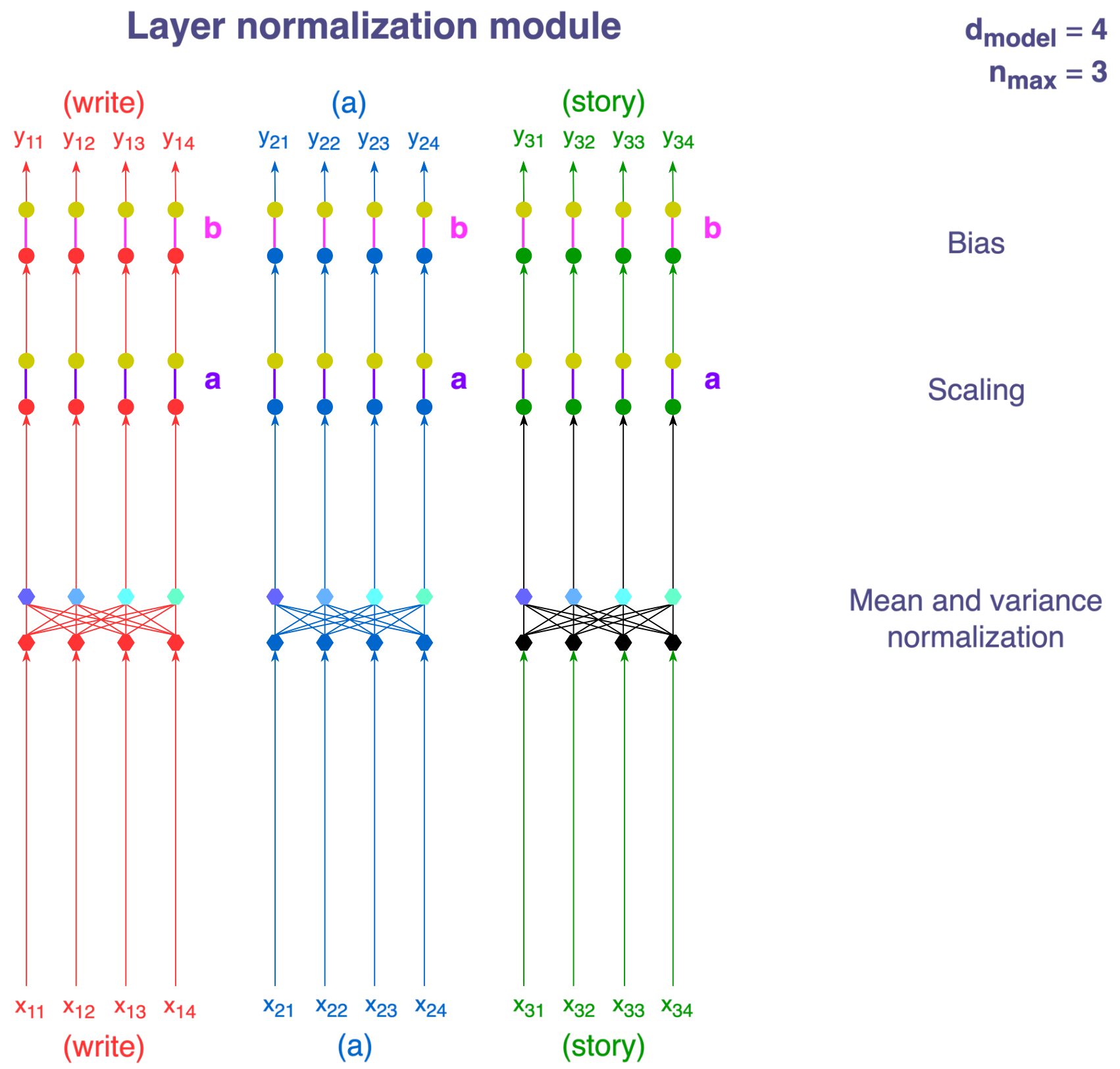

*Figure S14. Scheme of the system of simple neural networks corresponding to a feed-forward network module.*

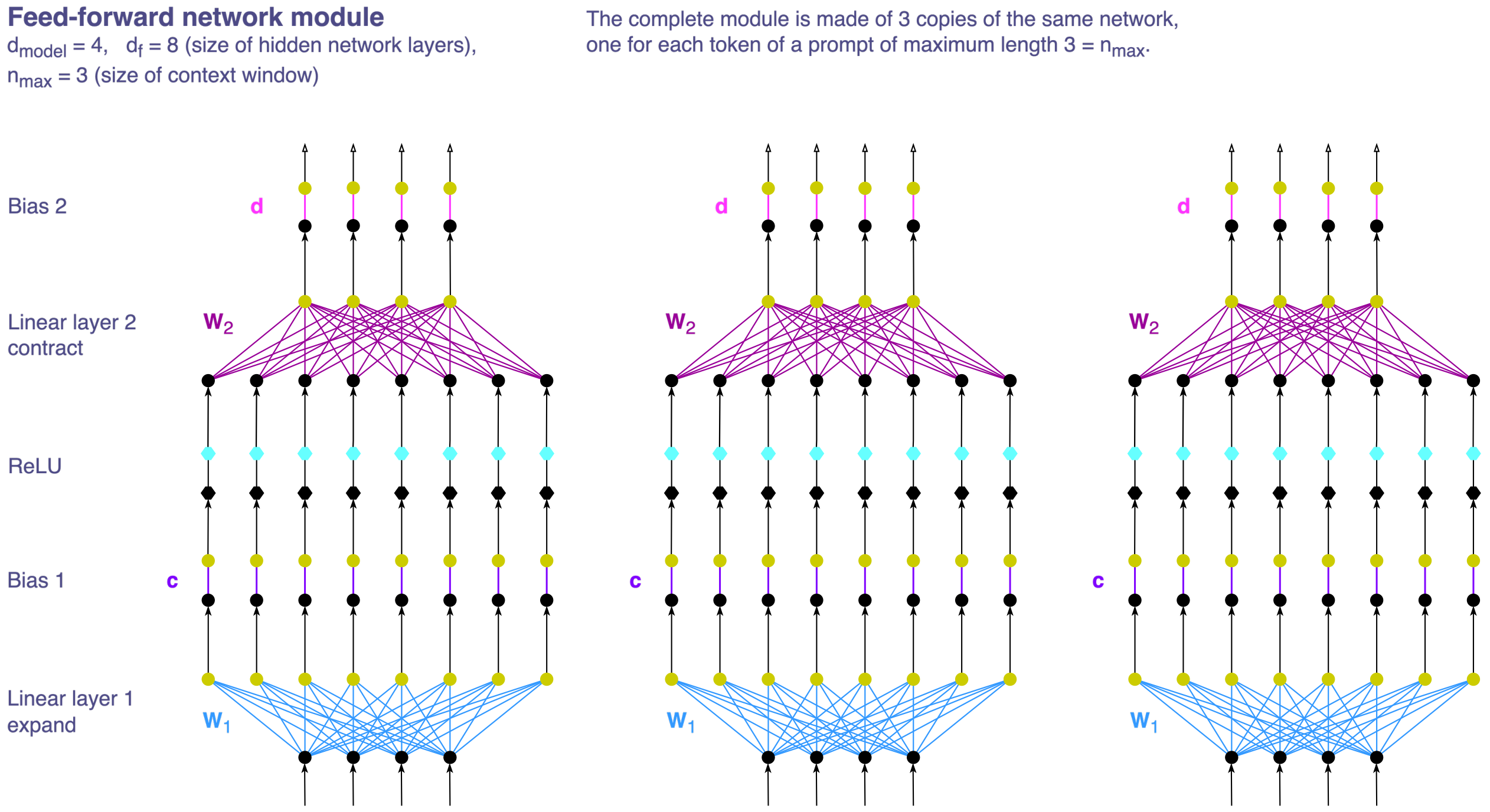

*Figure S14a. The same scheme with the networks activated by an input of n = 1 token and the corresponding output.*

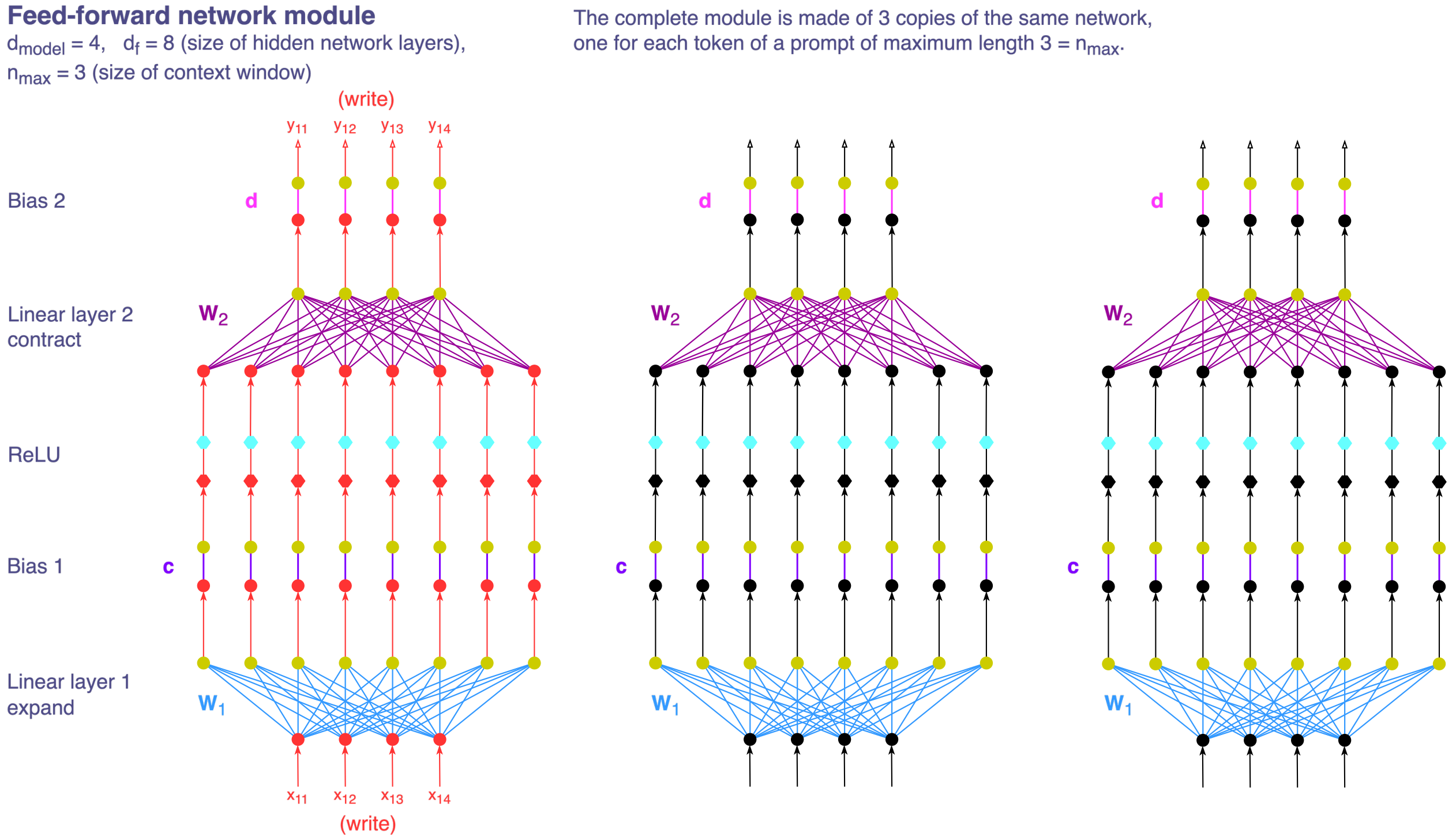

*Figure S14b. The same scheme with the networks activated by an input of n = 2 tokens and the corresponding output.*

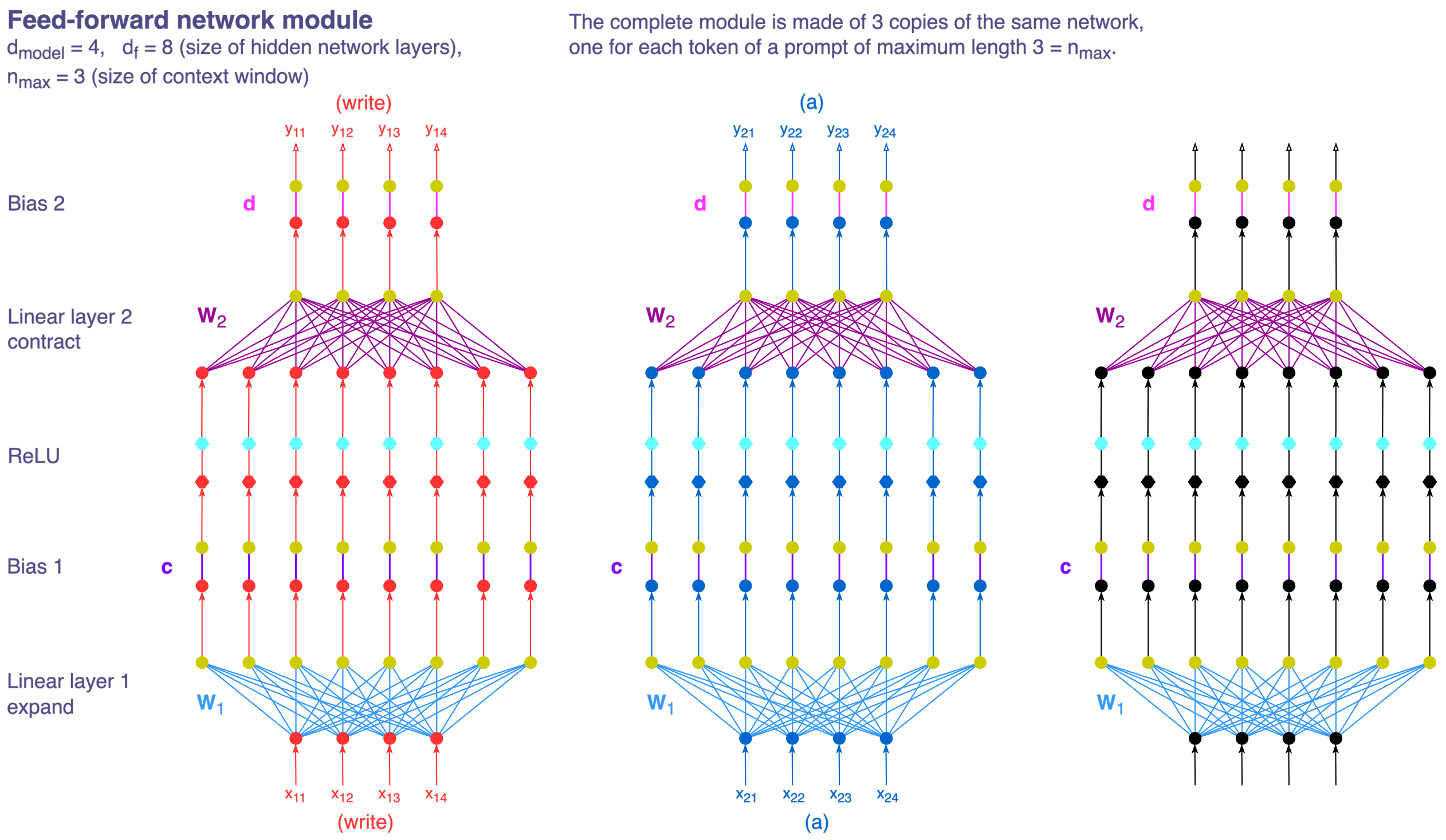

*Figure S14c. The same scheme with the networks activated by an input of n = 3 tokens and the corresponding output.*

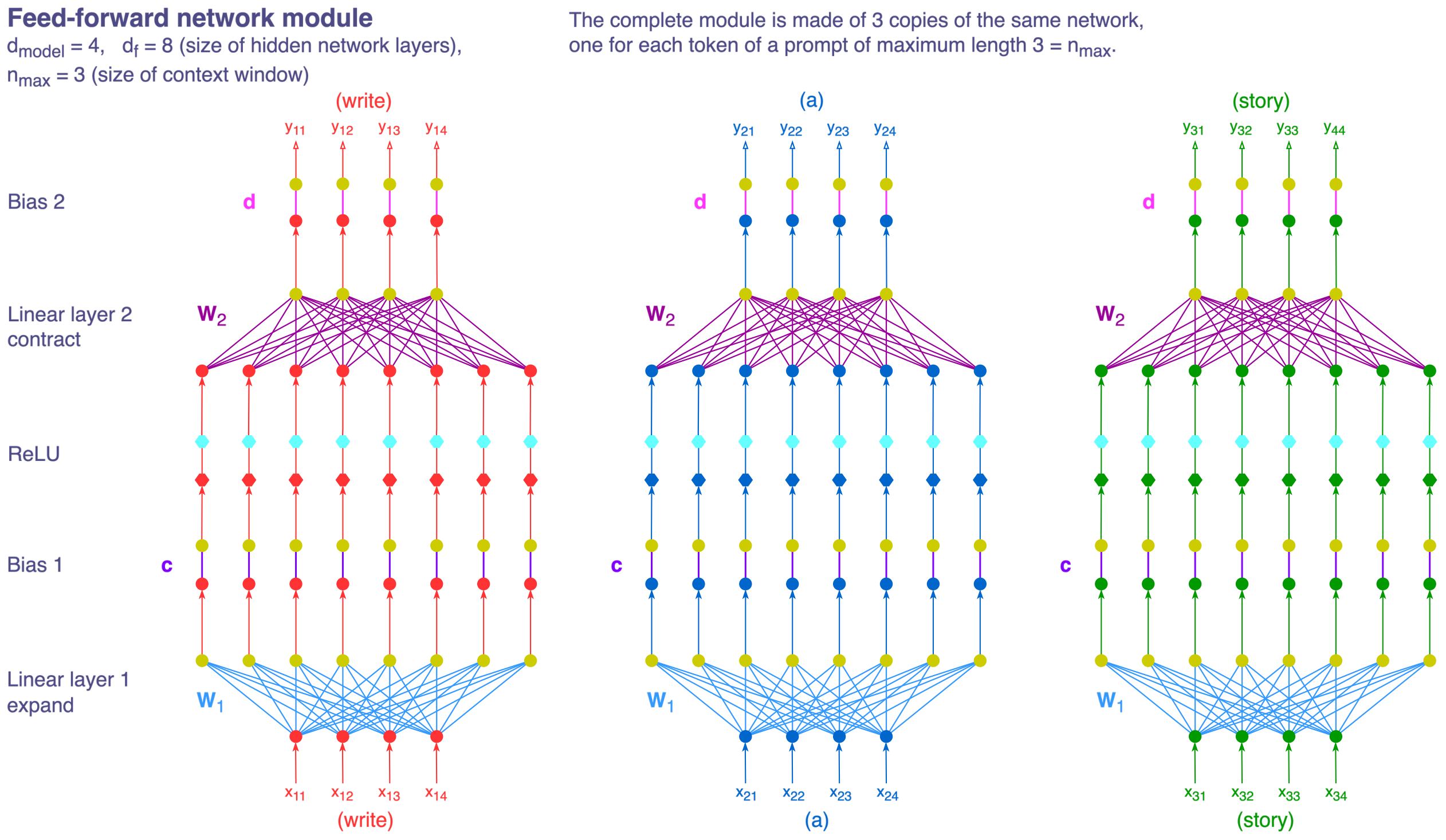

*Figure S15. Scheme of the system of simple neural networks corresponding to a forecast module.*

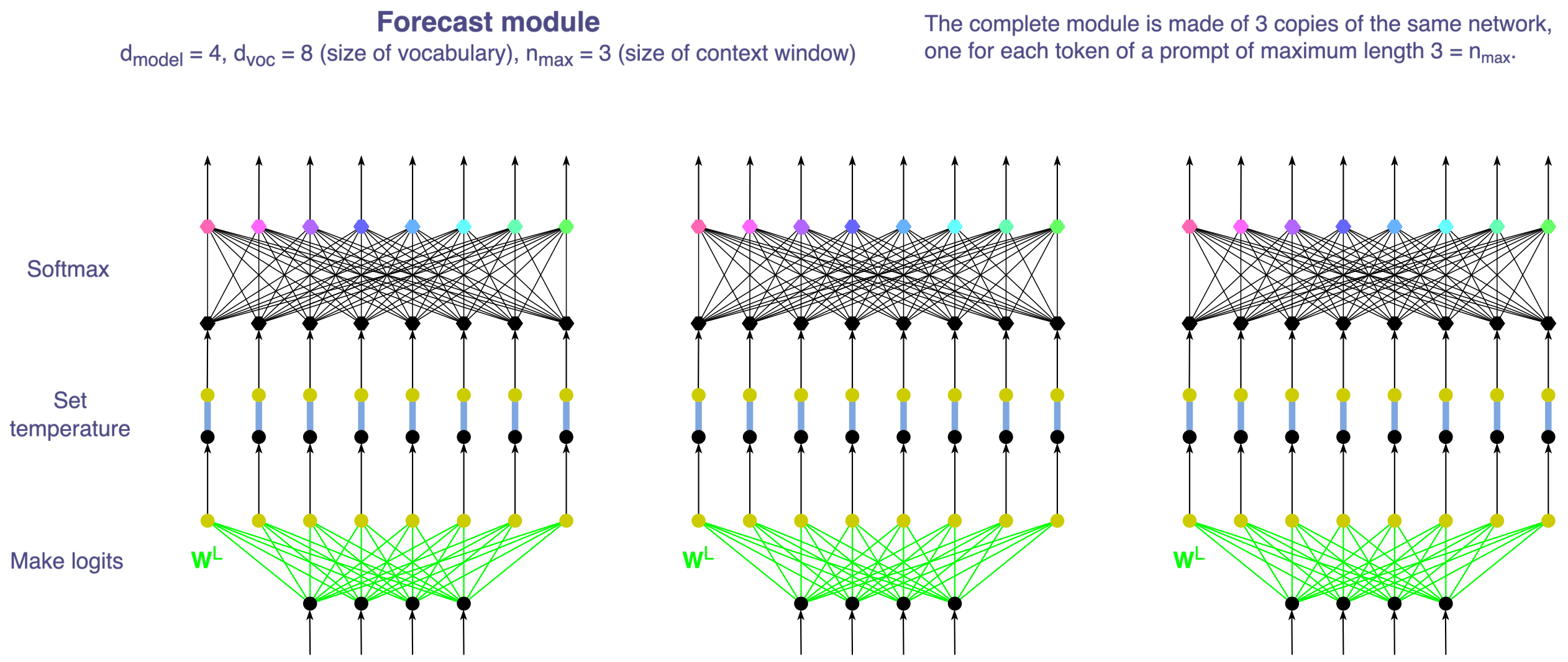

*Figure S15a. The same scheme with the networks activated by an input of n = 1 token and the corresponding output.*

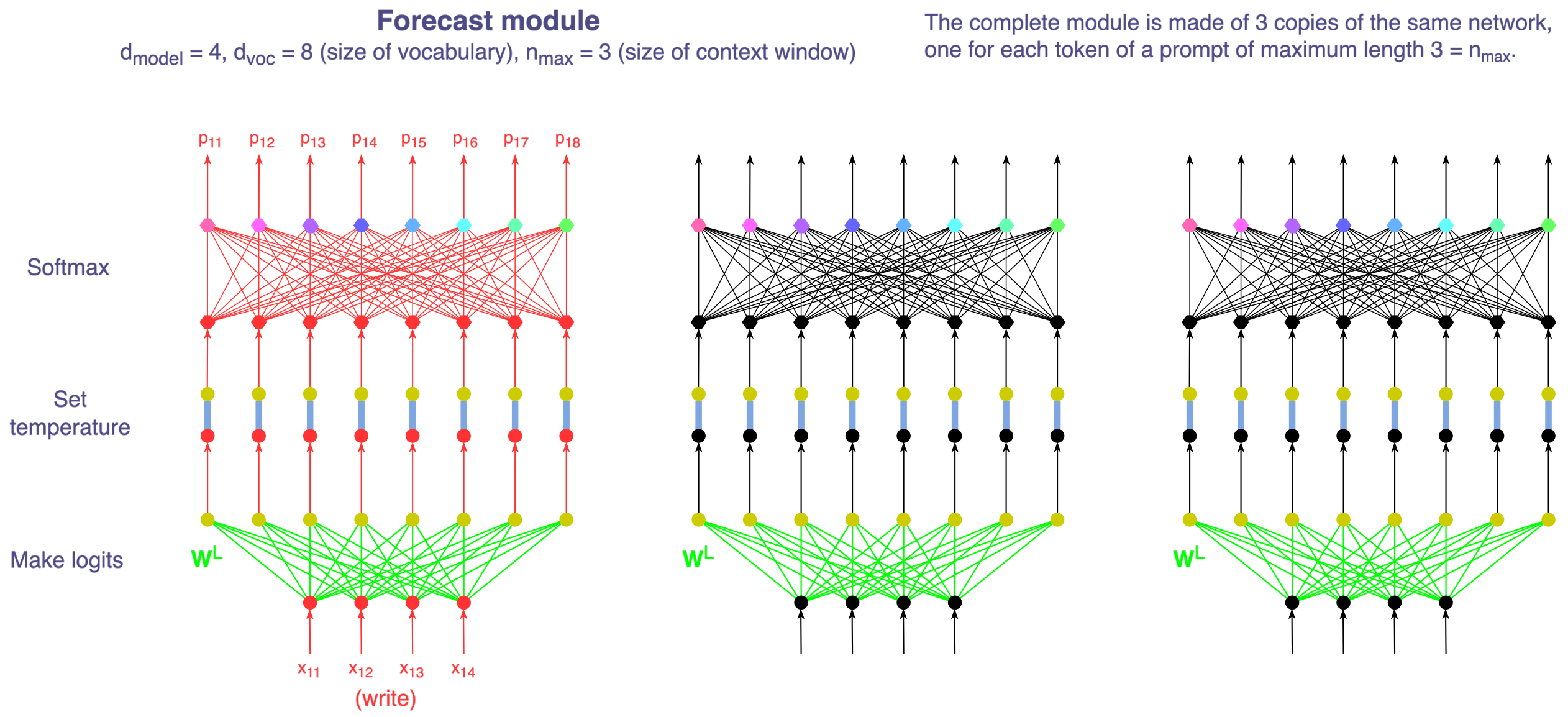

*Figure S15b. The same scheme with the networks activated by an input of n = 2 tokens and the corresponding output.*

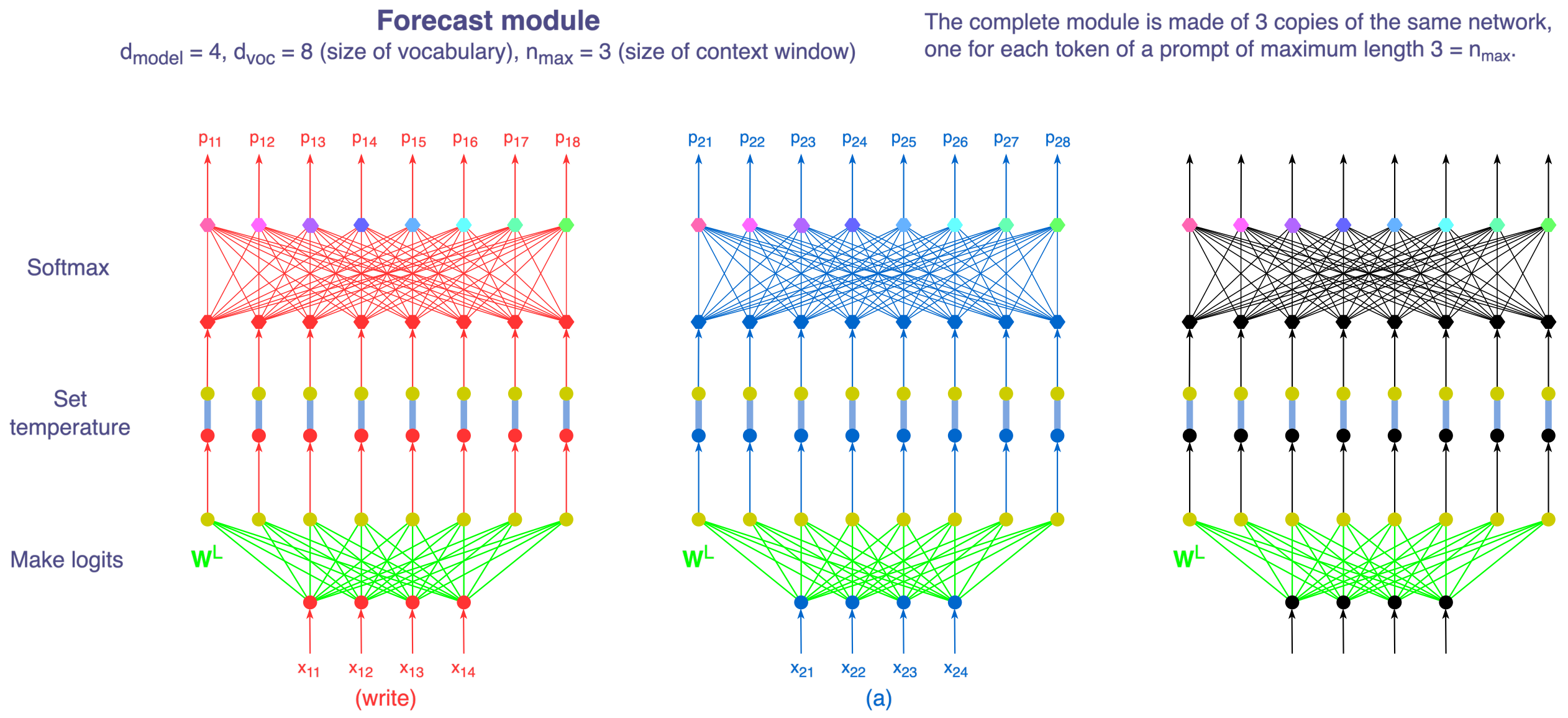

*Figure S15c. The same scheme with the networks activated by an input of n = 3 tokens and the corresponding output.*

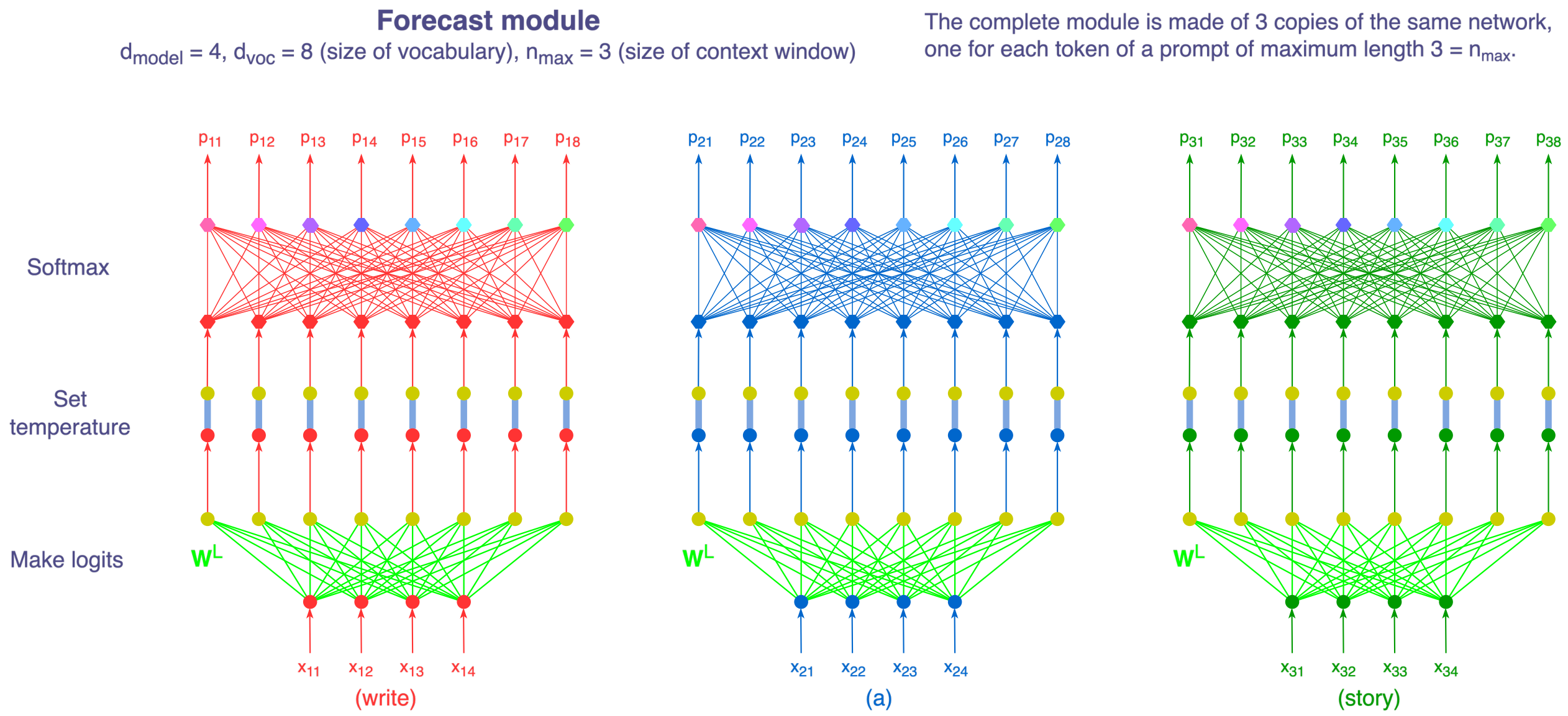

## SUPPLEMENTARY FIGURES. Part C: Two representations of simple neural networks: traditional and neuromorphic

*Figure S16. A dense linear network with dynamic weights determined by output–weight interconnections originating from output units of other networks.*

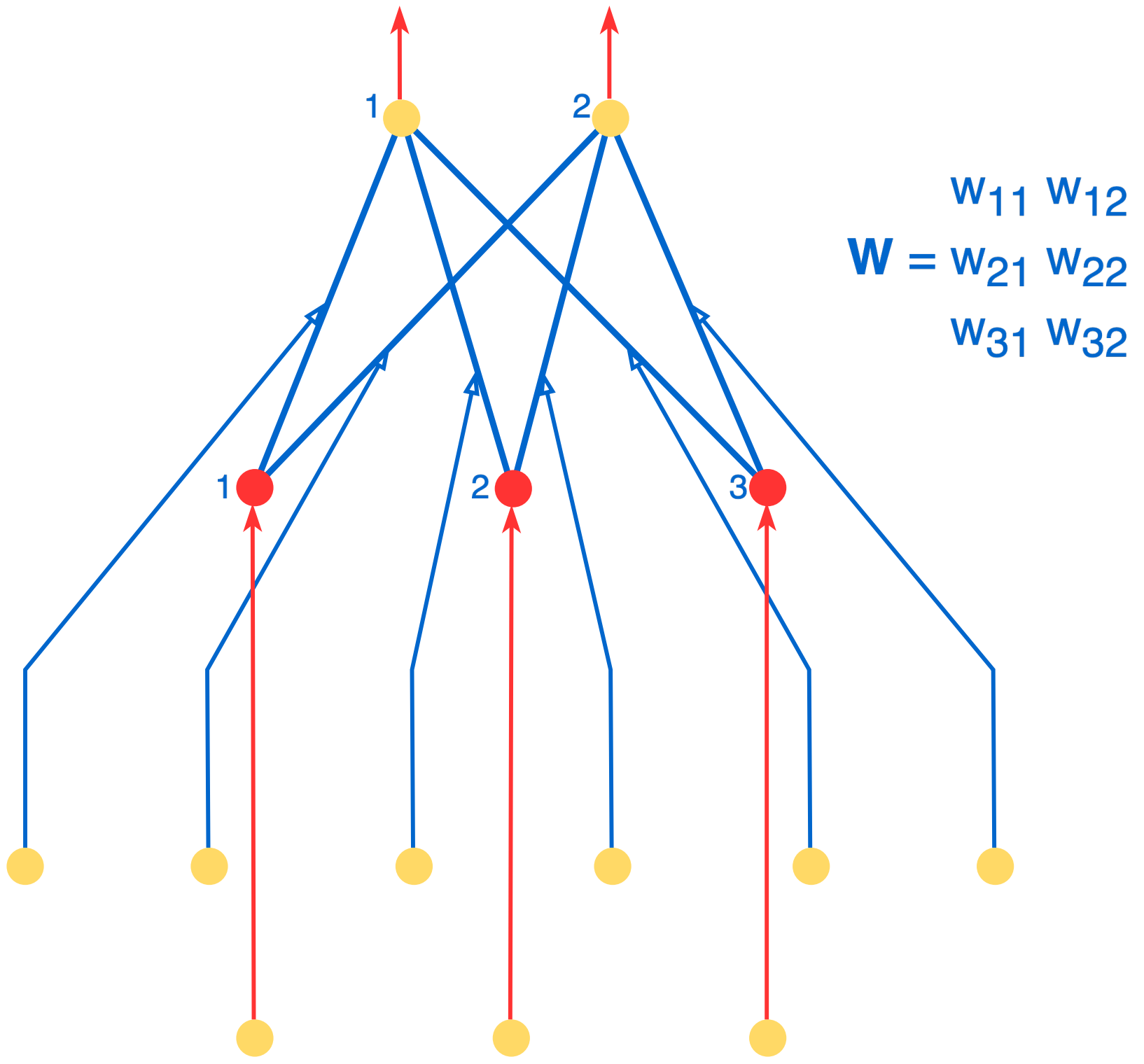

*Figure S17. Traditional representation of a dense linear network with static weights. The output units are the only functional units; they compute the weighted sum of the signals received from each input unit. The input units do not perform any transformation on the incoming signals, but merely route them to the output units.*

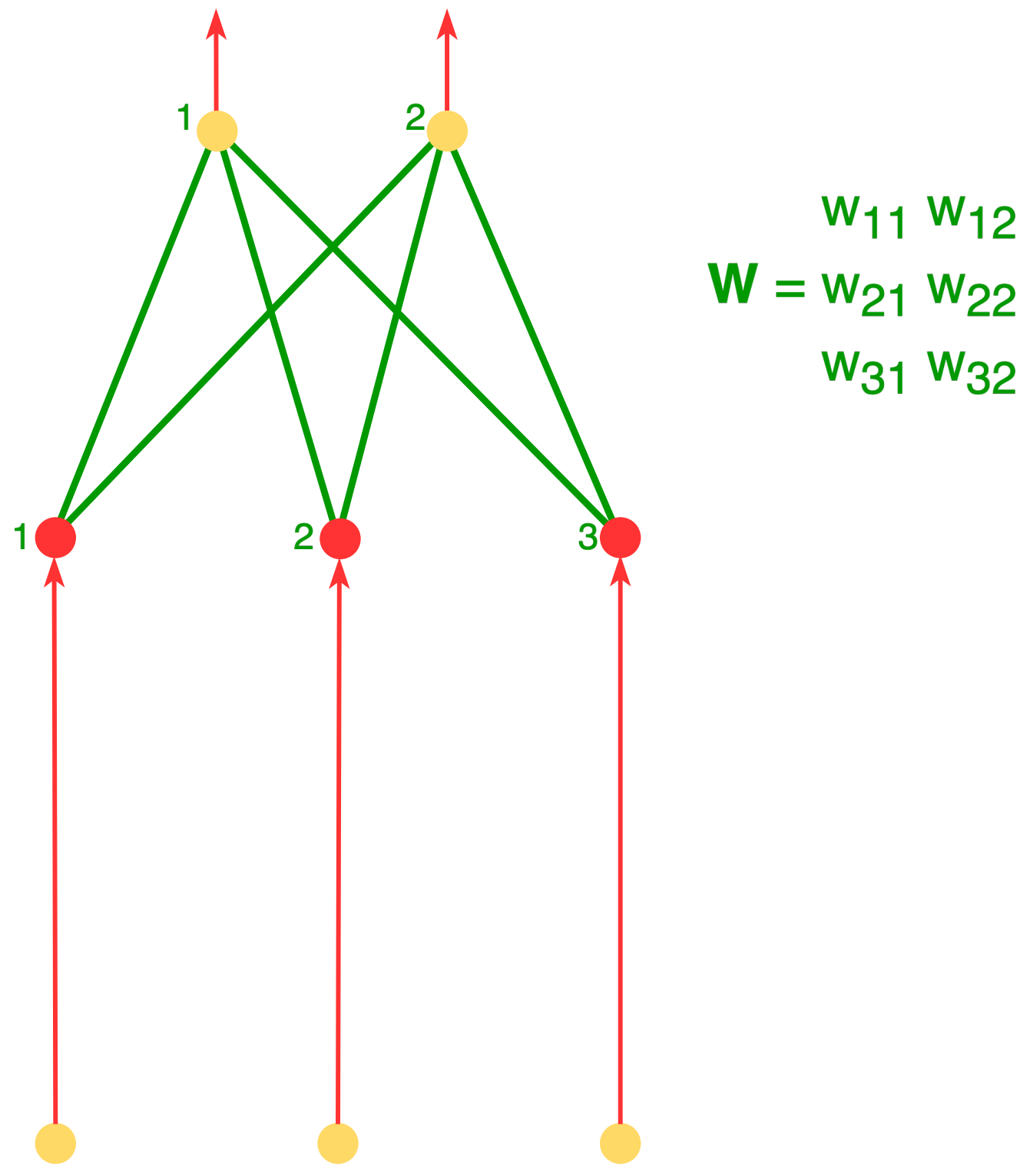

*Figure S18. Neuromorphic representation of the dense linear network with static weights shown in Fig. S17. The output units compute only the sum of the already weighted signals received from the multiplicative units located on the connections and represented by green circles.*

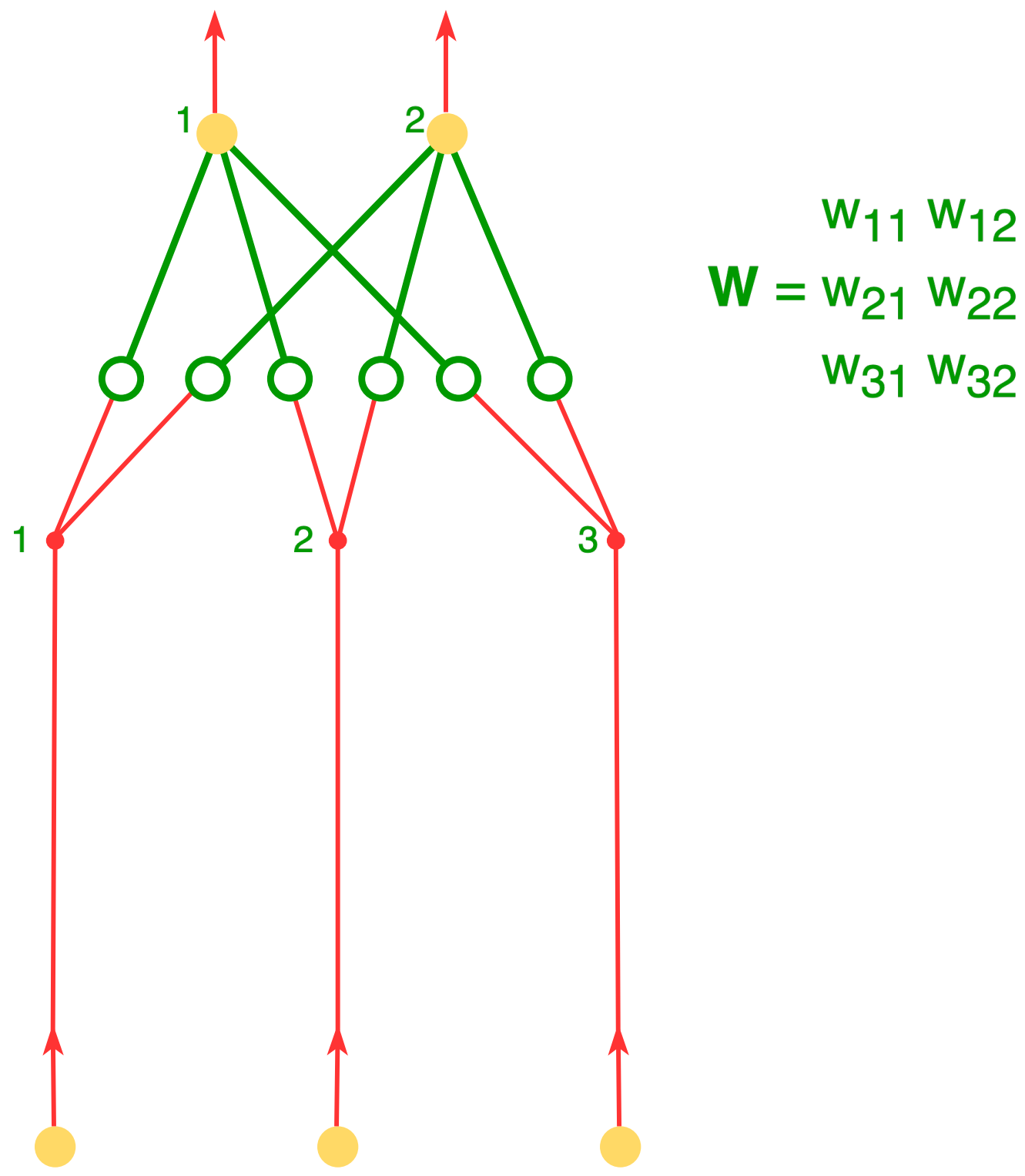

*Figure S19. Neuromorphic representation of the dense linear network with dynamic weights shown in Fig. S16. The multiplicative units located on the connections have two inputs: the second inputs (in blue) provide the connection weights, which come from output units of other networks.*

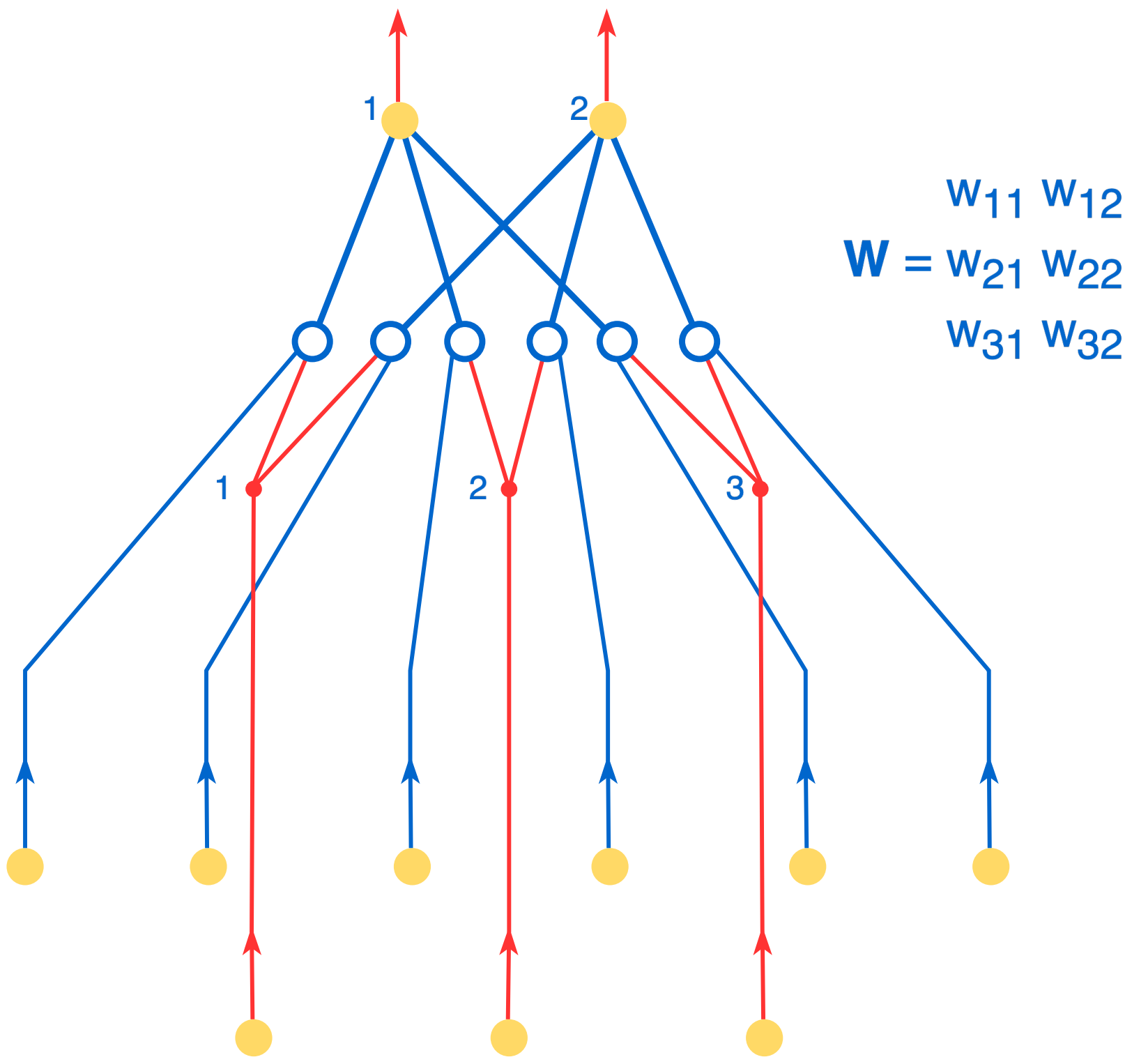

*Figure S20. Schematic representation of an axo-axonic synapse. A second axon converges on the presynaptic bouton of a first axon, modulating its neurotransmitter release.*

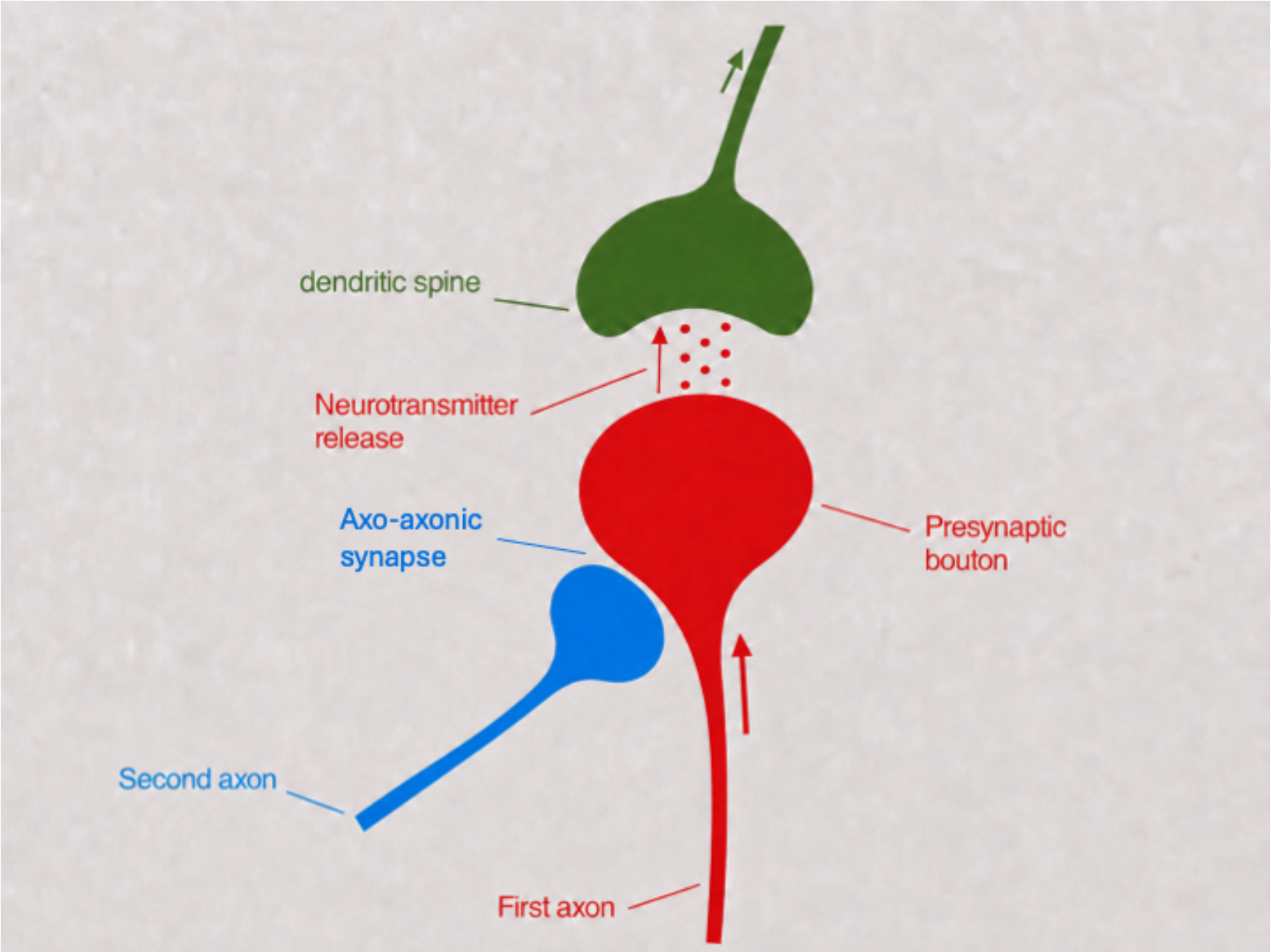